\documentclass{ieeeaccess}
\usepackage{cite}
\usepackage{multirow}
\usepackage{amsmath,amssymb,amsfonts}
\usepackage{algorithmic}
\usepackage{graphicx}
\usepackage{subcaption}
\usepackage{caption}
\usepackage{textcomp}
\usepackage{booktabs} 
\usepackage{array}
\usepackage{stfloats}                  % lets table* appear at bottom with [!b] if needed
\usepackage{placeins}                  % optional: \FloatBarrier to flush earlier floats

\usepackage[figuresright]{rotating} % was \usepackage{rotating}
\usepackage{graphicx}        % for \resizebox 
\usepackage{booktabs}        % for \toprule, \midrule, \cmidrule, \bottomrule
\usepackage{multirow} % for \multirow
\usepackage[dvipsnames,table]{xcolor}

\usepackage{xcolor}
\usepackage{hyperref} 

\usepackage{enumitem}

\usepackage{url}

\usepackage{hyperref} 

\PassOptionsToPackage{hyphens}{url}\usepackage{hyperref}

\def\BibTeX{{\rm B\kern-.05em{\sc i\kern-.025em b}\kern-.08em
    T\kern-.1667em\lower.7ex\hbox{E}\kern-.125emX}}
\begin{document}
\history{Date of publication xxxx 00, 0000, date of current version xxxx 00, 0000.}
\doi{10.1109/ACCESS.2017.DOI}

\title{A Large-scale Benchmark on Geological Fault Delineation Models: Domain Shift, Training Dynamics, Generalizability, Evaluation and Inferential Behavior}
\author{\uppercase{Jorge Quesada}\authorrefmark{1}, \uppercase{Chen Zhou}\authorrefmark{1}, \uppercase{Prithwijit Chowdhury}\authorrefmark{1}, \uppercase{Mohammad Alotaibi}\authorrefmark{1},  \uppercase{Ahmad Mustafa}\authorrefmark{2}, 
\uppercase{Yusufjon Kumakov}\authorrefmark{3}, \uppercase{Mohit Prabhushankar (MEMBER, IEEE)}\authorrefmark{1}, and \uppercase{Ghassan AlRegib (FELLOW, IEEE)}\authorrefmark{1},}
\address[1]{OLIVES at the Georgia Institute of Technology}
\address[2]{
Information Technology University}
\address[3]{
Tashkent State Technical University}

\tfootnote{This work is supported by the ML4Seismic Industry Partners at Georgia Tech}

\markboth
{Author \headeretal: Preparation of Papers for IEEE TRANSACTIONS and JOURNALS}
{Author \headeretal: Preparation of Papers for IEEE TRANSACTIONS and JOURNALS}

\corresp{Corresponding author: Ghassan AlRegib (e-mail: alregib@gatech.edu).}

%------------------------------------------------------

\onecolumn % make sure you keep this coverpage as one column. In this template, we force the coverpage to be one column with this command and then switch to double column for the remaining of the paper with the \doublecolumn command. 

\begin{description}[labelindent=0cm,leftmargin=3cm,style=multiline]

\item[\textbf{Citation}]{J. Quesada, C. Zhou, P. Chowdhury, M. Alotaibi, A. Mustafa, Y. Kumakov, M. Prabhushankar, and G. AlRegib, "A Large-scale Benchmark on Geological Fault Delineation Models: Domain Shift, Training Dynamics, Generalizability, Evaluation and Inferential Behavior", accepted at  IEEE Access}

\item[\textbf{Review}]{First submission: 12 May 2025 (Major revision 1) \\
                        Second submission: 1 Sept 2025  (Major revision 2) \\
                        Third submission: 30 oct 2025 (Accepted, minor revision)}

\item[\textbf{Video abstract}]{\url{https://youtu.be/SIjY0qywEpw?si=qEvIbC_aWTT66hUv}}

\item[\textbf{Code}]{\url{https://github.com/olivesgatech/large-bench-geo}}

\item[\textbf{Dataset}]{https://alregib.ece.gatech.edu/software-and-datasets/cracks-crowdsourcing-resources-for-analysis-and- \\categorization-of-key-subsurface-faults/}

%\item[\textbf{Code}]{\url{https://github.com/olivesgatech/PointPrompt}}% If you do not have data related to this paper, you can remove the data keyword.

%\item[\textbf{Dataset}]{\url{https://zenodo.org/records/11580815}}

\item[\textbf{Bib}]{@ARTICLE\{Quesada2025Largescale,\\ 
author=\{J. Quesada and C. Zhou and P.Chowdhury and M. Alotaibi and A. Mustafa and Y. Kumakov and M. Prabhushankar and G. AlRegib\},\\ 
journal=\{IEEE Access\},\\ 
title=\{A Large-scale Benchmark on Geological Fault Delineation Models: Domain Shift, Training Dynamics, Generalizability, Evaluation and Inferential Behavior\}, \\ 
year=\{2025\},\}\\ 
}

% Preprint sharing policy can vary depending on the publisher. Before posting a paper to arXiv, please specifically check the transaction/convference you are targeting. In some transactions, papers are usually added to arxiv after acceptance. Pubslishers usually allow the authors to share accepted version of their papers but not the final formatted version that is provided by the pubisher. In case of sharing preprints, publishers usually ask to add DOI and citation to the paper along with a copyright notice.

\item[\textbf{Copyright}]{\textcopyright Creative Commons Attribution CCBY 4.0 }

\item[\textbf{Contact}]{\{jpacora3, alregib\}@gatech.edu  \\ \url{https://alregib.ece.gatech.edu/} }

\item[\textbf{Corresponding author}]{alregib@gatech.edu }

\end{description}

%Following command sequence was used to start the paper content from the following page and avoid numbering cover page.
\thispagestyle{empty}
\newpage
\clearpage
\setcounter{page}{1}

%Cover page was 1 column. \twocolumn changes the page format back to double column.
\twocolumn

%------------------------------------------------------

\begin{abstract}
Machine learning has taken a critical role in seismic interpretation workflows, especially in fault delineation tasks. However, despite the recent proliferation of pretrained models and synthetic datasets, the field still lacks a systematic understanding of the generalizability limits of these models across seismic data representing diverse geologic, acquisition, and processing settings. In practice, distributional shifts between surveys, limitations in fine-tuning strategies, and inconsistent evaluation protocols remain major obstacles to deploying reliable models in real exploration settings. In this paper, we present the first large-scale benchmarking study explicitly designed to evaluate domain shift strategies for seismic fault delineation. Our benchmark spans over 200 experimental setups combining eight architectures, three datasets (\texttt{FaultSeg3D}, \texttt{CRACKS}, \texttt{Thebe}), and multiple training regimes (individual training, fine-tuning, and joint training). We evaluate performance using three complementary metrics (Dice, Hausdorff Distance, and Bidirectional Chamfer Distance) and analyze both segmentation accuracy and structural fidelity. Our results show that fine-tuning across dissimilar domains can reduce source-domain Dice by up to 75\%, demonstrating severe catastrophic forgetting, whereas larger models such as Segformer tend to be more robust to adaptation than smaller architectures. We also find that domain adaptation methods outperform fine-tuning under large distributional gaps but underperform when domains are closely aligned. Finally, we complement conventional metrics with an analysis of fault-network descriptors (length, curvature, sinuosity, segmentation density), revealing the nuances in the interplay between architectural choices and data properties. Overall, this benchmark provides a reproducible foundation for evaluating transferability in seismic fault delineation and offers actionable insights for effective deployment of fault delineation models within seismic interpretation workflows.

\end{abstract}


\titlepgskip=-15pt

\maketitle

\section{Introduction}
\label{sec:introduction}
\PARstart{R}{ecent} advances in deep learning (DL) have brought about a tectonic shift in the seismic interpretation workflow \cite{isah_review_2025, wenxue_overview_2025}, mirroring broader trends across other domains like geoscience \cite{yaqoob_geocrack_2024, qaiser_vision_2025}, sustainable energy systems \cite{yaqoob_advancing_2025, yaqoob_microcrystalnet_2025, yaqoob_fluidnet-lite_2025} and biomedical applications \cite{logan2022multi, logan2022patient, kokilepersaud2023clinically, quesada2022mtneuro, kokilepersaud2022gradient, iffa2025development, ansari_survey_2025}. DL-assisted approaches are leveraged in different parts of the pipeline, specifically in automated fault detection. Faults are geologically critical features that control fluid flow in Earth's subsurface, influence reservoir compartmentalization, and pose drilling hazards in hydrocarbon exploration \cite{wu2019faultSeg, chowdhury2025unified}. Beyond the energy sector, faults play a central role in earthquake nucleation and propagation, as well as geohazard and risk assessment in tectonically active regions \cite{benzion2008collective,KANAMORI20031205}. Accurate and scalable fault interpretation is therefore a high-impact task across several geoscience domains. Early work in this area often repurposed semantic segmentation models from computer vision (such as those developed for natural images) to detect discontinuities in seismic sections. However, these generic frameworks struggled with the unique characteristics of seismic data, particularly the thin, curvilinear geometry of faults and the presence of acquisition artifacts and structural noise. As a result, the field has increasingly shifted toward fault delineation, a task-specific variant of segmentation that emphasizes the extraction of coherent fault structures rather than general pixel-wise classification. This shift has prompted the development of domain-specific neural networks, feature encoders, adapted loss functions, and structural priors that better capture the morphological signatures of faults \cite{rs16050922,faultsam}. %Despite the rich literature of classical faults interpretation methods~\cite{alregib2018subsurface}, 
In general, the growing availability of DL methods has accelerated the adoption of data-driven interpretations that have provided new insights into fault structures within large seismic volumes  \cite{rs16050922,faultsam}.

% \textcolor{red}{Early work in seismic analysis often repurposed pure computer vision (CV) detection and segmentation models—originally developed for natural images—to identify discontinuities in seismic sections. However, these generic CV frameworks frequently struggled with the unique characteristics of fault patterns, such as thin, curvilinear geometries and noisy subsurface data. As a result, the field has transitioned toward specialized fault delineation approaches, incorporating domain-specific feature encoders, adapted loss functions, and structural priors that better capture the morphological signatures of faults in seismic volumes.}  

\begin{figure*}
    \centering
    \includegraphics[width=\linewidth]{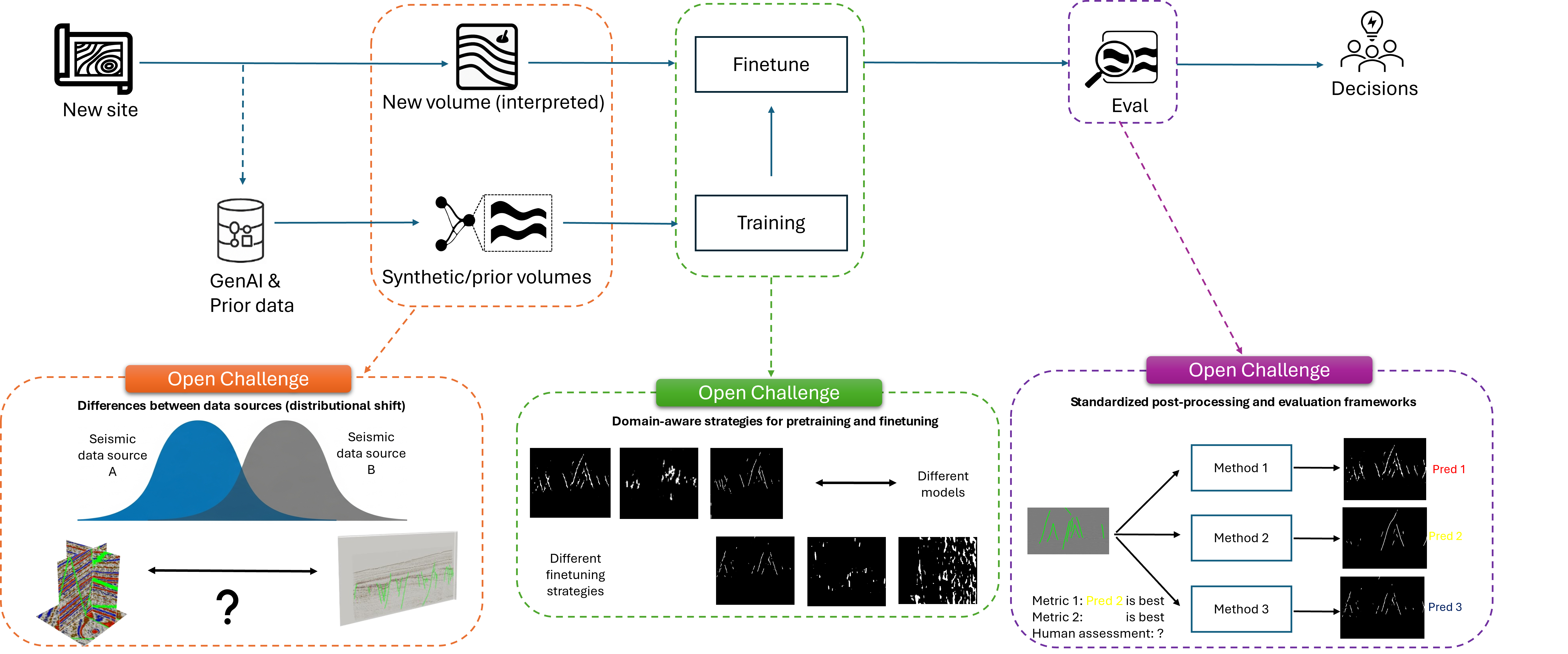}
    \caption{Typical DL-assisted seismic interpretation workflow}
    \label{fig:motivation}
\end{figure*}

A DL-assisted seismic interpretation pipeline is illustrated in Fig.~\ref{fig:motivation} (top diagram). In a standard workflow, interpreters at a new site begin by leveraging existing labeled datasets or synthetically generated volumes to train machine learning models. These models capture broad geophysical patterns and serve as a foundational prior, which can then be adapted to the new site through fine-tuning on a smaller set of locally interpreted seismic data. After fine-tuning, the models are evaluated within a consistent framework to ensure that their outputs align with geological expectations and interpretation standards. The validated predictions subsequently guide subsurface decision-making, and the new volume may in turn contribute to the pool of training data for future sites.  However, while conceptually straightforward, this workflow embeds multiple assumptions and design choices that present significant challenges in practice, which we highlight at the bottom of Figure \ref{fig:motivation}.

% \Figure[t!](topskip=0pt, botskip=0pt, midskip=0pt){Figures/motivation.png} {Typical seismic interpretation workflow.\label{fig:motivation}}

% \begin{figure}
%     \centering
%     \includegraphics[width=\linewidth]{Figures/expertvsnovice_1.png}
%     \caption{Expert Vs. Novice labels, when used for pre-training models. The results indicate that experts labels can be replaced with lower cost labels.}
%     \label{fig:expertvsnovice}
% \end{figure}

%\input{Tables/2vs3fintune}

% \begin{figure}
%     \centering
%     \begin{subfigure}{\linewidth}
%         \centering
%         \includegraphics[width=0.9\linewidth]{Figures/problemsa.png}
%         \caption{Distributional Shift}
%         \label{fig:problemsa}
%     \end{subfigure}

%     \vspace{1em}

%     \begin{subfigure}{\linewidth}
%         \centering
%         \includegraphics[width=0.9\linewidth]{Figures/problemsb.png}
%         \caption{Domain-aware Fine-tuning}
%         \label{fig:problemsb}
%     \end{subfigure}

%     \vspace{1em}

%     \begin{subfigure}{\linewidth}
%         \centering
%         \includegraphics[width=0.9\linewidth]{Figures/problemsc.png}
%         \caption{Standardized Evaluation}
%         \label{fig:problemsc}
%     \end{subfigure}
%     \caption{Existing challenges in the current interpretation workflow}
%     \label{fig:probs}
% \end{figure}

%We further illustrate these challenges in Fig. ~\ref{fig:probs}: 
Two central and intertwined challenges in the interpretation pipeline arise from the diversity of data sources (Fig. ~\ref{fig:motivation}, orange box) and the strategies used to adapt models across them (Fig.~\ref{fig:motivation}, green box).
%Two of the central challenges in the above pipeline arise from (i) the diversity of data sources and (ii) the strategies used to adapt models across them. 
Seismic datasets differ widely in their geological characteristics, acquisition parameters, and resolution, among other factors. Synthetic data, while providing a fully specified ground truth, often fails to capture the complexity and variability of field data. Prior datasets from other regions, though more realistic, may still differ significantly from new target volumes. These domain shifts, whether from \emph{synthetic-to-real} or across real-world basins, can cause pretrained models to generalize poorly to new sites \cite{nasim2020seismic}, and eventually provide weak performance for the interpretation task.   

\begin{figure}
    \centering
    \includegraphics[width=.85\linewidth]{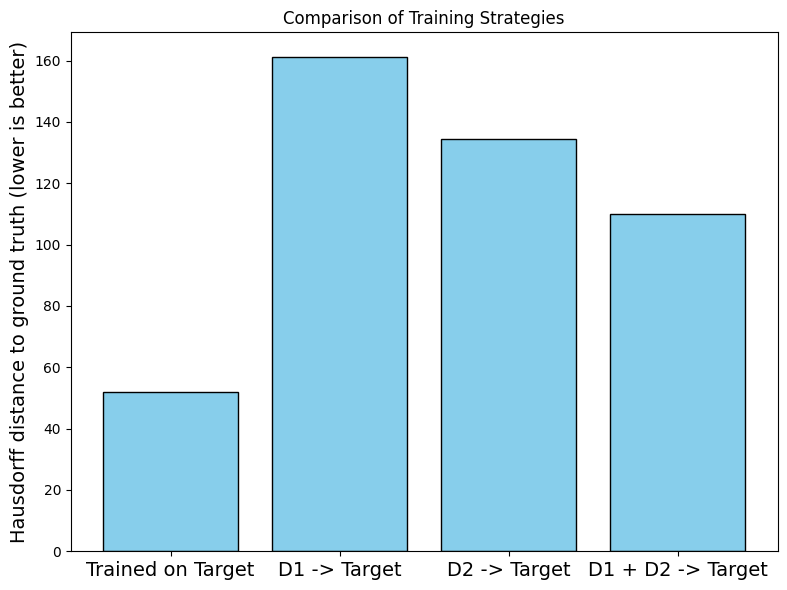}
    \caption{Finetuning example from a real dataset (D1) and a synthetic dataset (D2) to a target real dataset}
    \label{fig:hook}
\end{figure}

To address the challenge of domain adaptation, researchers have proposed a variety of methods that address the effects of domain differences along the pipeline. Several studies introduce enhancements to model architectures \cite{LI2023105412}, dataset construction \cite{alaudah2018structure,alaudah2019machine,wu2019faultSeg,AN2021107219}, uncertainty estimation~\cite{benkert2022reliable} or explore training strategies aimed at improving data efficiency and transferability \cite{benkert2023samples,mustafa2021man}. A common approach involves pretraining models on synthetic data and subsequently fine-tuning them on real seismic volumes, as in~\cite{wu2019faultSeg}. More recent methods have explored using self-supervised learning to learn robust features without large amounts of labeled data \cite{wang2016interactive, cohen2006detection, roberts2001curvature, di2017seismic, shafiq2018towards, kokilepersaud2022volumetric}, or adapting large vision foundation models trained on natural images to the seismic domain~\cite{faultsam,Quesada_2024_CVPR, quesada2024benchmarking}. While these approaches offer promising results, their effectiveness is often dependent on the compatibility between source and target domains, and they remain vulnerable to issues such as catastrophic forgetting~\cite{kirkpatrick2017overcoming} during adaptation. Moreover, the success of these approaches is influenced by several factors such as architecture and training design choices, as well as dataset-specific properties such as seismic section resolution, heterogeneity, and sample size.

Fig.~\ref{fig:motivation} (green box) showcases a simple example of the aforementioned issues. The figure contains two rows, each displaying three seismic sections. The first row presents the output of three segmentation models commonly used in the seismic community, trained under the same settings with the same training data. We can observe that the three predictions differ significantly from each other.  In the second row of the box, three segmentation outputs are shown after applying three different fine-tuning strategies. The first output in the second row is obtained by a model trained from scratch on the target data, while the other two are produced by models pretrained on other (larger) datasets and then finetuned on the target data. We further conduct objective evaluation of these outputs and summarize these metrics in  Fig.~\ref{fig:hook}. The y-axis in this figure is the average Hausdorff distance between the output of the model and the ground truth, where smaller values signify better predictions. There are four strategies shown. The first one is where the model is trained on the target volume. The second strategy is to use a model that was trained on a real dataset and fine-tune it on the target dataset. The third strategy is to use a model that was trained in a synthetic dataset and then fine-tuned on the target dataset. These three strategies match the three depicted in Fig.~\ref{fig:motivation}(green box). A fourth strategy is to use a model that was jointly trained on both synthetic and real data, then fine-tuned on the target data. Although we will discuss these experiments in more detail later in the paper, the figure here shows the large difference in performance across these four strategies. For now, this shows that despite this growing body of work, the field still lacks a systematic understanding of \emph{when}, \emph{how}, and \emph{why} a given strategy outperforms others under domain shift. This paper is the first attempt, to our knowledge, within the community to provide answers and guidelines for domain shift strategies in seismic interpretation.

%where we attempt to transfer knowledge from models trained either on real or synthetic data (or a combination of both) by finetuning on a target real dataset, and observe that this holds no improvement over just training the model directly on the target dataset from scratch. 

The third critical complication in the pipeline arises in evaluation, as depicted in Fig.~\ref{fig:motivation} (purple box). Fault labels are derived from expert interpretation and are often subjective~\cite{alcalde2017impact, guillon2020ground}, particularly in complex or ambiguous structural settings. This subjectivity leads to inconsistencies in ground truth across datasets or among annotators, and complicates objective model evaluation. Furthermore, a given metric may penalize correct but unannotated predictions, while failing to reflect geological plausibility or structural consistency \cite{sarajarvi2020robust, rs16050922}. To address this, our study provides an in-depth analysis of commonly used evaluation metrics (such as Dice coefficient, Chamfer and Hausdorff distances) in the context of fault interpretation. This analysis highlights the need for a more holistic evaluation approach that integrates quantitative measures with geological interpretability.

This paper presents a large-scale benchmarking study regarding the performance variability of finetuning strategies for fault delineation models across training regimes, data characteristics, and architectural choices. Rather than introducing a new model, this work establishes a systematic experimental framework: a methodological foundation for reproducible and cross-domain evaluation of deep learning architectures in seismic fault interpretation. Our benchmarked fault delineation models include a combination of eight deep learning architectures, trained and fine-tuned on three distinct datasets, including both synthetic and real volumes. Our experiments cover a broad spectrum of training strategies, including pretraining on a single fault dataset, pre-training on large-scale natural image dataset such as ImageNet~\cite{deng2009imagenet}, using randomly initialized models and jointly training on multiple fault datasets. Beyond benchmarking, we also use targeted case studies (through domain adaptation and continual learning setups) to diagnose model behavior under catastrophic forgetting and domain shift, demonstrating how the benchmark can serve as a tool for both evaluation and insight. 

While prior works have examined transfer learning and domain adaptation in geoscience~\cite{sun2016deep, tsai2018learning, trinidad2023seismic}, these efforts have typically been limited in scope: focusing on single architectures or isolated dataset pairs. By contrast, our study provides the first large-scale, systematic benchmark that unifies evaluation across multiple model families, datasets, and training paradigms. This effort parallels the role of domain-shift benchmarks in fields like medical imaging~\cite{guan_domain_2022}, where standardized comparisons have proven essential for understanding generalization and reproducibility across domains.

Our contributions are as follows:

\begin{itemize}
    \item We present the first large-scale benchmarking study focused on fault delineation under domain shift, spanning over 200 exhaustive combinations of experimental setups across 8 DL architectures, 5 pretraining-finetuning datasets and 3 evaluation ones, as well as several training configurations.
    \item  We systematically evaluate pretraining, fine-tuning, and joint training strategies, exposing failure cases such as catastrophic forgetting and dataset-dependent brittleness.
    \item We analyze the effect of architectural scale and design on transferability, highlighting that larger models often adapt more effectively to finetuning, and smaller models benefit from domain adaptation strategies under large domain shift.
    \item  We compare evaluation metrics and highlight their limitations, advocating for evaluation practices that incorporate structural and geological plausibility. We also introduce a novel analysis based on fault characteristic metrics, which sheds new light on the way data properties and model behavior are entangled, and opens up new avenues from which to understand finetuning and domain transfer strategies.
    \item We open-source our code and models to provide a foundational reference to scientists and practitioners.
\end{itemize}

The remainder of this paper is organized as follows. Section~\ref{sec:related_work} reviews prior work on machine learning for fault delineation in the context of seismic interpretation workflows. Section~\ref{sec:datasets} describes the three datasets used in our benchmarking study, highlighting their geological and statistical characteristics. Section~\ref{sec:experimental_setup} details our experimental setup, including data preparation, model architectures, and evaluation protocols. Section~\ref{sec:results} presents the results of over 200 training configurations, analyzed across generalization and transferability, training dynamics, and metric evaluation. Section~\ref{sec:fault_metric_results} introduces an additional analysis on geometric and topological metrics to characterize dataset–model interactions. Finally, Section~\ref{sec:conclusion} concludes the paper by summarizing key findings, outlining practical guidelines, and identifying open challenges for future research. The full cosde and data used for this paper can be found at \url{https://github.com/olivesgatech/large-bench-geo}.

\section{Related Work}
\label{sec:related_work}
The structure of this section mirrors the stages outlined in Fig.~\ref{fig:motivation}. Existing literature on machine learning for fault delineation follows a similar modular approach: beginning with data preprocessing and input normalization, proceeding through model training, adaptation strategies and inferential behavior, and culminating in evaluation protocols. To provide both breadth and depth, each subsection first summarizes the general deep learning principles that motivate a given stage, and then narrows the discussion to prior work in seismic interpretation and fault segmentation specifically. In this way, we highlight how established DL techniques intersect with geophysical challenges, and how each component contributes to generalization across seismic domains.

\subsection{Data Preparation and Preprocessing}

In computer vision segmentation pipelines, raw images from natural scenes \cite{long2015fully,chen2017deeplab,kokilepersaud2023focal} and medical scans~\cite{ronneberger2015u,isensee2021nnu, prabhushankar2022olives} are first standardized to ensure consistent spatial dimensions and intensity distributions \cite{ioffe2015batch,ulyanov2016instance}. Typical preprocessing steps include resizing or cropping to a fixed height and width, data augmentation, whitening~\cite{prabhushankar2017generating}, and contrast adjustments to mitigate variability in lighting or sensor settings \cite{shorten2019survey,wang2017effectiveness}. These operations guarantee that each input conforms to the network’s architectural requirements and that learned features are not biased by extraneous intensity fluctuations.

Seismic segmentation extends these practices to volumetric data acquired in formats such as SEG-Y (\texttt{.sgy}), raw binary (\texttt{.dat}), or serialized NumPy arrays (\texttt{.npy}). A common workflow begins by reading trace headers to assemble a 3D volume of dimensions (inline \(\times\) xline \(\times\) time/depth), and either processing the volume at that level \cite{wu2019faultSeg, rs16050922} or extracting 2D inline or crossline sections for further processing \cite{alaudah2018structure, di2017seismic}. To enlarge the training set and fit GPU memory constraints, each section is tiled with an overlapping sliding window of size \(H \times W\) pixels (the effects of different tiling standards are studied and results presented in Section \ref{sec:window_size}.).

Since raw seismic amplitudes can span orders of magnitude and contain acquisition artifacts, it is a standard practice to normalize each volume either via min–max scaling or z-score transformation. %For min–max, each amplitude \(v_{ij}\) becomes
% \begin{equation}
% v'_{ij} = 2\,\frac{v_{ij} - \lambda_{\min}}{\lambda_{\max} - \lambda_{\min}} - 1,
% \label{eq:minmax}
% \end{equation}
% where \(\lambda_{\min}\) and \(\lambda_{\max}\) are the global minimum and maximum amplitudes in the volume. Alternatively, z-score normalization
% \begin{equation}
% v''_{ij} = \frac{v_{ij} - \mu}{\sigma}
% \label{eq:zscore}
% \end{equation}
%centers the data to zero mean and unit variance, with \(\mu\) and \(\sigma\) computed over all voxels. 
Fault masks (either interpreter‐drawn or synthetically generated) are saved as binary images (\texttt{.png} or \texttt{.npy}) and cropped identically to their corresponding seismic windows. This systematic preprocessing pipeline (from raw SEG-Y/DAT ingestion, through standardized normalization and windowed slicing, to precisely aligned input–mask pairs) provides a consistent basis for benchmarking and comparing fault delineation models across diverse seismic domains.

\subsection{Training Dynamics and Setups}

Popular network choices designed for natural image segmentation are sensitive to training and design choices like model architecture, input window size, and loss functions. Networks such as \textsf{U-Net} \cite{ronneberger2015u}, \textsf{DeepLab} \cite{chen2017deeplab}, and \textsf{SegFormer} \cite{xie2021segformer} typically accept fixed‐size $n \times n$ patches (usually $n \in \{128, 256, 512\}$), balancing the need for sufficient context against GPU memory limits. During each training iteration, these patches are sampled, sometimes at multiple scales to expose the network to varied object sizes and to regularize against overfitting. Segmentation models optimize losses designed to reconcile pixel‐wise accuracy with region‐level coherence. The binary cross-entropy (BCE) loss \cite{cox1958regression} focuses on classifying each pixel correctly, while the Dice loss \cite{milletari2016v} measures the overlap between predicted and ground-truth fault regions, making it well-suited for imbalanced datasets. In practice, many works combine BCE and Dice to take advantage of both pixel fidelity and shape alignment \cite{wazir2022histoseg, zu2024resaceunet}.

% For binary masks, a commonly used loss is the pixel‐wise binary cross‐entropy (BCE) \cite{cox1958regression}, given as follows:   
% \begin{equation}
% \mathcal{L}_{\mathrm{BCE}} = -\frac{1}{N}\sum_{i=1}^{N}\bigl[y_i\log(p_i) + (1-y_i)\log(1-p_i)\bigr],
% \label{eq:bce}
% \end{equation}
% where \(p_i\) is the predicted probability at pixel \(i\), \(y_i\in\{0,1\}\) is the ground‐truth label, and \(N\) is the number of pixels in the patch. An alternative that tends to better adapt to settings with class imbalance and emphasize overlap is the Dice loss \cite{milletari2016v} given as follows:
% \begin{equation}
% \mathcal{L}_{\mathrm{Dice}} = 1 - \frac{2\sum_{i=1}^{N}p_i\,y_i + \epsilon}{\sum_{i=1}^{N}p_i + \sum_{i=1}^{N}y_i + \epsilon},
% \label{eq:dice}
% \end{equation}
% where \(\epsilon\) is a small constant to provide numerical stability. The BCE loss emphasizes the pixel fidelity while the Dice loss takes the overlap into consideration. Therefore, hybrid objectives combining BCE and Dice, e.g. \(\mathcal{L} = \alpha\mathcal{L}_{\mathrm{BCE}} + (1-\alpha)\mathcal{L}_{\mathrm{Dice}}\), are commonplace to leverage both pixel fidelity and region overlap \cite{wazir2022histoseg, zu2024resaceunet}. 

In seismic fault delineation, similar principles apply but with additional considerations. Input patches are typically larger to capture fault continuity across sections, and training often uses overlapping windows with a stride smaller than the patch size to ensure boundary faults are adequately sampled~\cite{wu2019faultSeg, di2019developing, di2019improving, di2019semi}. The aforementioned loss functions, coupled with learning rate schedules and different regularization functions, form the backbone of seismic training setups \cite{dou2022md, benkert2024effective, mustafa2023active}. By carefully tuning window sizes and loss compositions, researchers mitigate class imbalance and preserve structural continuity in fault delineation \cite{mustafa2024visual}.

\subsection{Generalization and Transferability}
\label{subsec:GeneralTransf}

Generalization refers to a model’s ability to maintain performance when presented with new, unseen data that (ideally) follows the same distribution as the training set \cite{goodfellow2016deep, benkert2022example, mustafa2021joint}. In segmentation, this means accurately delineating objects or regions under variations in lighting, scale, or background texture. However, a common obstacle to generalization is domain shift, which occurs when the statistical properties of training and test data differ (such as changes in camera sensors or scene composition) often leading to degraded performance \cite{torralba2011unbiased, Temel2017_NIPS, temel2018cure, temel2019traffic, prabhushankar2022introspective}. Domain adaptation encompasses strategies to reduce this gap when the target data is limited, for instance by aligning feature distributions between source and target domains \cite{ganin2016domain, chen2019temporal}. Another common technique to adapt to target domains is fine-tuning: a network pretrained on a large, generic dataset (the source) is adapted to a more specialized task (the target) by retraining some or all layers, thereby leveraging learned representations while adjusting to new data characteristics \cite{panigrahi2020survey,yosinski2014transferable}.

In seismic segmentation, generalization to new domains is particularly difficult due to variability in acquisition parameters, stratigraphy, frequency content, and noise levels across surveys. This is especially true for fault delineation, where subtle and discontinuous features are easily obscured by processing artifacts or geologic heterogeneity \cite{ercoli2023evidencing, dou2022md}. Models trained on one domain (such as synthetic datasets like \texttt{FaultSeg3D} \cite{wu2019faultSeg}) often fail to transfer effectively to real data from different basins. Recent surveys \cite{mousavi_deep-learning_2022, mousavi_applications_2024} provide comprehensive overviews over the way deep learning approaches in exploration seismology struggle with such domain shifts and emphasize fault segmentation as a particularly brittle task.

To mitigate these challenges, researchers have explored a variety of domain adaptation strategies, including feature-space  \cite{sun2016deep} and frequency-space \cite{trinidad2023seismic} alignment, adversarial learning \cite{tsai2018learning}, and seismic-style transfer \cite{du2022disentangling}. Fault-specific adaptations include synthetic fault injection to enrich training distributions \cite{wang2022structural}, contrastive learning schemes designed for curvilinear structures \cite{kokilepersaud2022volumetric}, and self-supervised pretraining frameworks that improve transferability across surveys with limited labels \cite{rs16050922, faultsam}. Transfer learning pipelines typically rely on pretrained encoders (either from natural images \cite{he2016deep, simonyan2014very} or large-scale seismic simulations with domain-specific decoders \cite{dou2022md, wu2019faultSeg}) followed by fine-tuning on limited labeled sections. These strategies must navigate a tradeoff between plasticity and stability: aggressive weight updates are prone to inducing catastrophic forgetting \cite{kirkpatrick2017overcoming}, while conservative tuning may fail to capture domain-specific features. As a result, generalization in seismic fault delineation requires careful calibration of both model architecture and adaptation strategy to handle the subtle, spatially sparse structures across diverse geological, acquisition, and imaging settings.

\subsection{Inferential Behavior}

In semantic segmentation, inference entails mapping learned feature representations to discrete pixel‐level predictions \cite{long2015fully}. The fidelity of this mapping depends on the architecture’s ability to aggregate context and fuse features: models with narrow receptive fields may accurately localize edges but miss global structure \cite{ronneberger2015u}, while those with extensive context capture coherent regions at the expense of sharp boundaries \cite{chen2017deeplab}. During inference, the bias–variance trade‐off emerges as a tension between boundary precision and region consistency, sometimes mitigated by post‐processing, e.g., Conditional Random Fields \cite{krahenbuhl2011efficient} or edge‐aware refinement modules \cite{pohlen2017full}.

In seismic fault delineation, these inferential tendencies are amplified by the thin, curvilinear nature of faults and the high noise intrinsic to seismic volumes \cite{ercoli2023evidencing}. \textsf{U-Net}’s symmetric encoder–decoder and skip connections excel at preserving local detail, producing crisp fault traces when the signal‐to‐noise ratio is high \cite{ronneberger2015u}, but its reliance on local convolutions and fixed strides can fragment continuous faults under heteroskedastic noise \cite{dou2022md}. \textsf{DeepLab}’s \cite{chen2017deeplab} atrous convolutions and Atrous Spatial Pyramid Pooling (ASPP) module gather multi‐scale context, yielding smoother, globally coherent fault masks; yet the dilation patterns can inadvertently merge adjacent non‐fault discontinuities, introducing false positives along stratigraphic horizons. \textsf{SegFormer}’s transformer‐based encoder captures long‐range dependencies and adapts to complex fault geometries enhancing continuity across sections, though its patch‐based attention can produce coarser boundaries if patch size is not carefully chosen \cite{xie2021segformer}. 

Furthermore, recent works in both seismic \cite{mousavi_deep-learning_2022, mousavi_applications_2024} and general DL \cite{recht2019imagenet, geirhos2020shortcut} domains emphasize that inferential biases are not merely architectural, but also dataset-driven:  synthetic datasets may thus encourage smoother, well-connected predictions, while field datasets can induce models to replicate jagged, discontinuous, or crossover-heavy structures (we analyze these trends in depth using the fault characteristic metrics described in Section \ref{sec:flt_metrics}). New approaches attempt to counteract these biases by introducing geological priors, like fault continuity preservation \cite{faultsam} or curvature-based losses \cite{kokilepersaud2022volumetric}.  These studies highlight that inferential behavior in seismic fault delineation is shaped jointly by architectural design and the structural biases present in training data, underscoring the importance of evaluation frameworks that go beyond pixel overlap to account for geological plausibility.

%These model‐specific inferential biases%—detailed boundary retention in U‐Net, global smoothing in DeepLab, and context‐driven consistency in SegFormer—
%drive distinct behaviors in fault delineation and inform the choice of architectures and post‐processing strategies for robust seismic interpretation.

\subsection{Performance Evaluation}
\label{sec:rw_metrics}

% Segmentation evaluation seeks to quantify how closely a predicted mask \(P\) approximates the ground-truth mask \(G\) as subsets of the image domain. Evaluation metrics are categorized in the following two ways: (i) region based measures, which treat masks as sets and quantify volumetric overlap, and (ii) distance-based measures, which treat masks as shapes and quantify geometric deviations along their contours.

% Region-based metrics include the Jaccard index (also known as intersection-over-union):
% \begin{equation}
% \mathrm{Jaccard}(P,G)
% = \frac{\lvert P \cap G\rvert}{\lvert P \cup G\rvert},
% \label{eq:jaccard}
% \end{equation}
% and the Dice coefficient:

Evaluation metrics for fault interpretation can be broadly grouped into two categories: region-based and distance-based metrics. Region-based metrics quantify the overlap between predicted and reference fault regions, while distance-based metrics measure the geometric discrepancy between their boundaries.

\textbf{Region-based metrics.} The Dice coefficient $D$ measures the spatial overlap between prediction $P$ and ground truth $G$
\begin{equation}
D
\;=\;
\frac{2\,\lvert P \cap G\rvert}{\lvert P\rvert + \lvert G\rvert}
\label{eq:dice}
\end{equation}
% which captures the proportion of shared pixels but may understate errors on thin or tortuous structures.

\textbf{Distance-based metrics.} The modified Hausdorff distance $HD$ captures the largest average boundary deviation:
\begin{equation}
    HD = \max\Biggl(
    \frac{1}{N_p} \sum_{p \in P} \min_{g \in G} d(p,g), \frac{1}{N_g} \sum_{g \in G} \min_{p \in P} d(p,g)
\Biggr) 
\label{eq:hd}
\end{equation}
where $d(\cdot, \cdot)$ is the Euclidean distance between two points, and \({N_p} \) and \({N_g} \) are the number of elements in the sets $P$ and $G$, respectively. Additionally, the Bidirectional Chamfer Distance $BCD$ computes the average shortest distance between boundary points in both directions:
\begin{equation}
    BCD 
= \frac{1}{N_p} \sum_{p \in G} \min_{g \in G} d(p,g) + \frac{1}{N_g}\sum_{g \in G} \min_{p \in P} d(p,g)
\label{eq:bcd}
\end{equation}
% We illustrate the complimentary nature of these metrics through a toy example in Fig. \ref{fig:metrics_toy_example}. 
This categorization provides a basis for later discussion in the Results section, where we compare the behaviors of overlap- and distance-focused metrics.

\subsection{Fault Characteristics Metrics}
\label{sec:flt_metrics}

To characterize fault networks within each dataset, we measure a set of geometric and topological descriptors. These include \textbf{Length}, \textbf{Curvature}, \textbf{Sinuosity}, \textbf{Segments}, and \textbf{Stepover Density}, which correspond to the statistics reported in Table~\ref{tab:dataset_stats}.

\textbf{Length ($L$)} \cite{kim2005relationship} is the total length of all fault traces, computed as
\begin{equation}
L = \sum_{i=1}^{N_{\text{faults}}} \ell_i ,
\end{equation}
where $\ell_i$ denotes the length of the $i$-th fault and $N_{\text{faults}}$ is the number of fault traces.

\textbf{Curvature ($\kappa$)} \cite{rutter2018geometry} quantifies the local bending of fault traces. At each arc-length position $s$, curvature is defined as
\begin{equation}
\kappa(s) = \left|\frac{d\theta}{ds}\right|,
\end{equation}
where $\theta$ is the tangent orientation along the fault.

\textbf{Sinuosity ($S$)} \cite{hanson2004style} measures how tortuous a fault is, defined as the ratio of its length to the straight-line distance between its endpoints:
\begin{equation}
S = \frac{L_{\text{trace}}}{D_{\text{endpoints}}}.
\end{equation}

\textbf{Segments ($N_{\text{seg}}$)} \cite{schwartz1989fault} is the total number of discrete fault segments observed in the network.

\textbf{Stepover Density ($D_{\text{stepover}}$)} \cite{manighetti2021fault} measures the relative occurrence of stepovers, normalized by fault length:
\begin{equation}
D_{\text{stepover}} = \frac{N_{\text{stepover}}}{L},
\end{equation}
where $N_{\text{stepover}}$ is the total number of stepovers. These descriptors enable dataset-level comparisons of fault geometry and topology.

% \subsubsection{Comparative Metrics}

% While individual metrics quantify dataset characteristics in isolation, comparative metrics evaluate the similarity between predicted and ground-truth faults. The metrics in Table~\ref{tab:comparative-metrics} extend the individual descriptors into differences, ratios, and statistical comparisons.

\begin{table*}[ht]
%\centering
\caption{Comparison of Netherlands F3 and Thebe Seismic Datasets}
\label{tab:F3_Thebe_comparison}
\resizebox{\textwidth}{!}{\begin{tabular}{p{4.2cm} p{6.2cm} p{6.2cm}}
\toprule
\textbf{Parameter} & \textbf{Netherlands F3 (North Sea) \cite{terranubis_f3_demo_2023}} & \textbf{Thebe (Exmouth Plateau, Australia) \cite{bailey2013geological}} \\
\midrule
Location & Offshore Netherlands, Block F3 & NW Shelf of Australia, Carnarvon Basin \\
Year of Acquisition & 1987 & 2007 \\
Area Coverage & $\sim$387 km\textsuperscript{2} & $\sim$1200 km\textsuperscript{2} \\
Inline × Crossline & 651 × 951 & $\sim$3174 × $\sim$1803 \\
Bin Size & 25 m × 25 m & $\sim$25 m × $\sim$25 m (assumed) \\
Record Length / Sampling & 0–1.848 s TWT, 4 ms & $\sim$0–4.5 s TWT, 2–4 ms (typical) \\
Original Purpose & Jurassic/Cretaceous hydrocarbon exploration & Triassic gas exploration \\
Public Wells / Logs & 4 wells with logs (sonic/gamma; 2 with density) & 2 wells (Thebe-1, Thebe-2) with gas discovery \\
Data Format & 3D post-stack time migrated (SEG-Y, OpenDtect) & 3D post-stack time migrated (SEG-Y, fault-labeled) \\
Structural Setting & Shallow deltaic shelf, minor faults, salt dome & Rifted margin, rotated fault blocks \\
Main Stratigraphy & Miocene–Pliocene clinoforms; deeper Jurassic/Cretaceous & Triassic Mungaroo Formation; Jurassic–Cretaceous seal \\
Reservoir Target & Jurassic sandstones; shallow biogenic gas pockets & Triassic fluvial sands (Mungaroo Formation) \\
Data Availability & Fully open (TNO/dGB/OpendTect) & Public fault-annotated subset; full survey open-file \\
Key Features & Clinoforms, shallow gas, polygonal faults, salt dome & Fault block trap, flat spots, complex fault network \\
Hydrocarbons & Shallow biogenic gas (non-commercial) & Confirmed dry gas field ($\sim$2–3 Tcf) \\
Source / Seal & Biogenic gas; intraformational shale; Zechstein salt & Mungaroo source/reservoir; Muderong Shale seal \\
\bottomrule
\end{tabular}}
\end{table*}

\section{Fault Delineation Datasets}
\label{sec:datasets}

%In this section, we discuss existing datasets for fault delineation and the challenges they pose to the development of machine learning models.

\begin{figure*}[t]
    \centering
    % Subfigure (a)
    \begin{subfigure}[b]{0.18\textwidth}
        \centering
        \includegraphics[width=\linewidth]{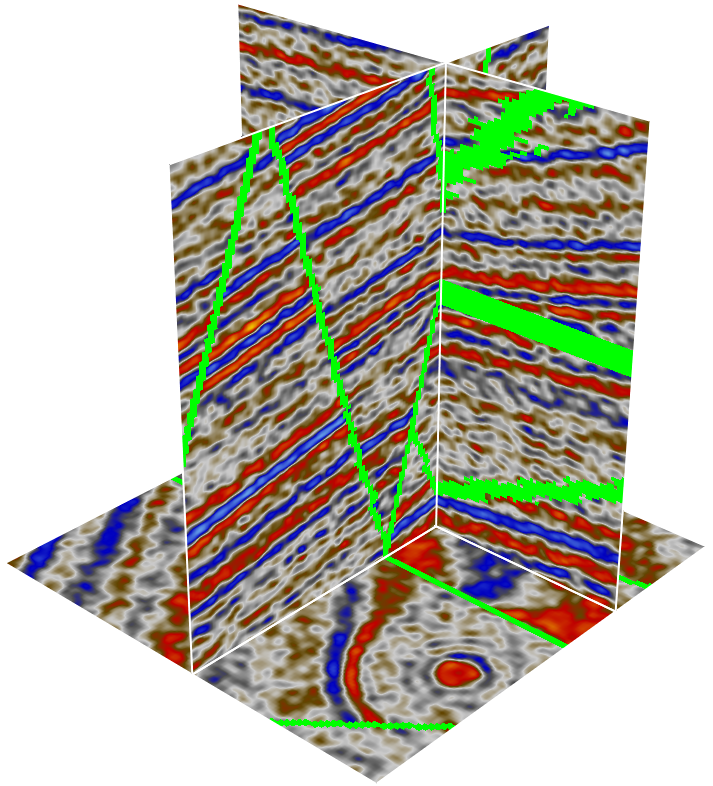}
        \caption{\texttt{FaultSeg3D}}
        \label{fig:datase-sub1}
    \end{subfigure}
    \hfill
    % Subfigure (b)
    \begin{subfigure}[b]{0.35\textwidth}
        \centering
        \includegraphics[width=\linewidth]{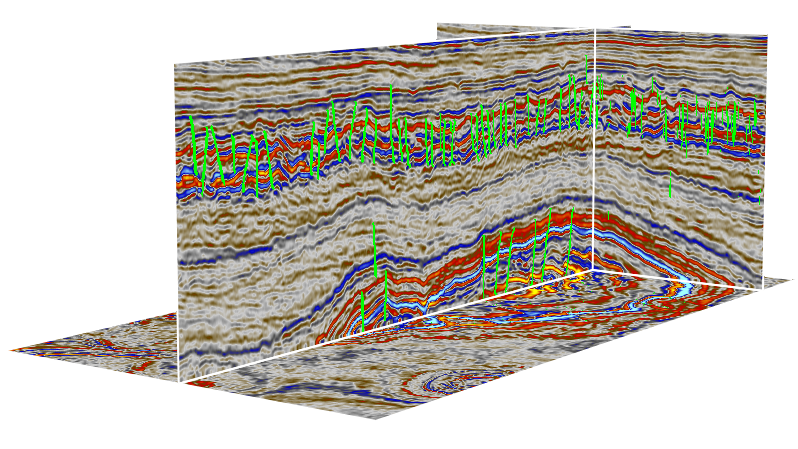}
        \caption{\texttt{Cracks}}
        \label{fig:datase-sub2}
    \end{subfigure}
    \hfill
    % Subfigure (c)
    \begin{subfigure}[b]{0.35\textwidth}
        \centering
        \includegraphics[width=\linewidth]{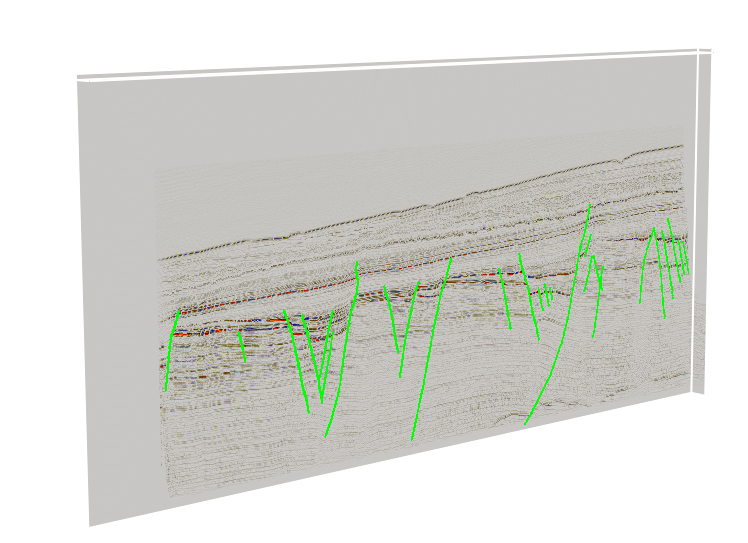}
        \caption{\texttt{Thebe}}
        \label{fig:datase-sub3}
    \end{subfigure}

    \caption{%
    Visuals from three datasets:~(\subref{fig:datase-sub1}) synthesized faults in \texttt{FaultSeg3D}, (\subref{fig:datase-sub2}) expert labels in \texttt{CRACKS}, and (\subref{fig:datase-sub3}) expert labels in \texttt{Thebe}.}
    \label{fig:datasets}
\end{figure*}

\begin{figure}
    %\centering
    \begin{subfigure}[]{.48\linewidth}
        %\centering
        \includegraphics[width=\textwidth]{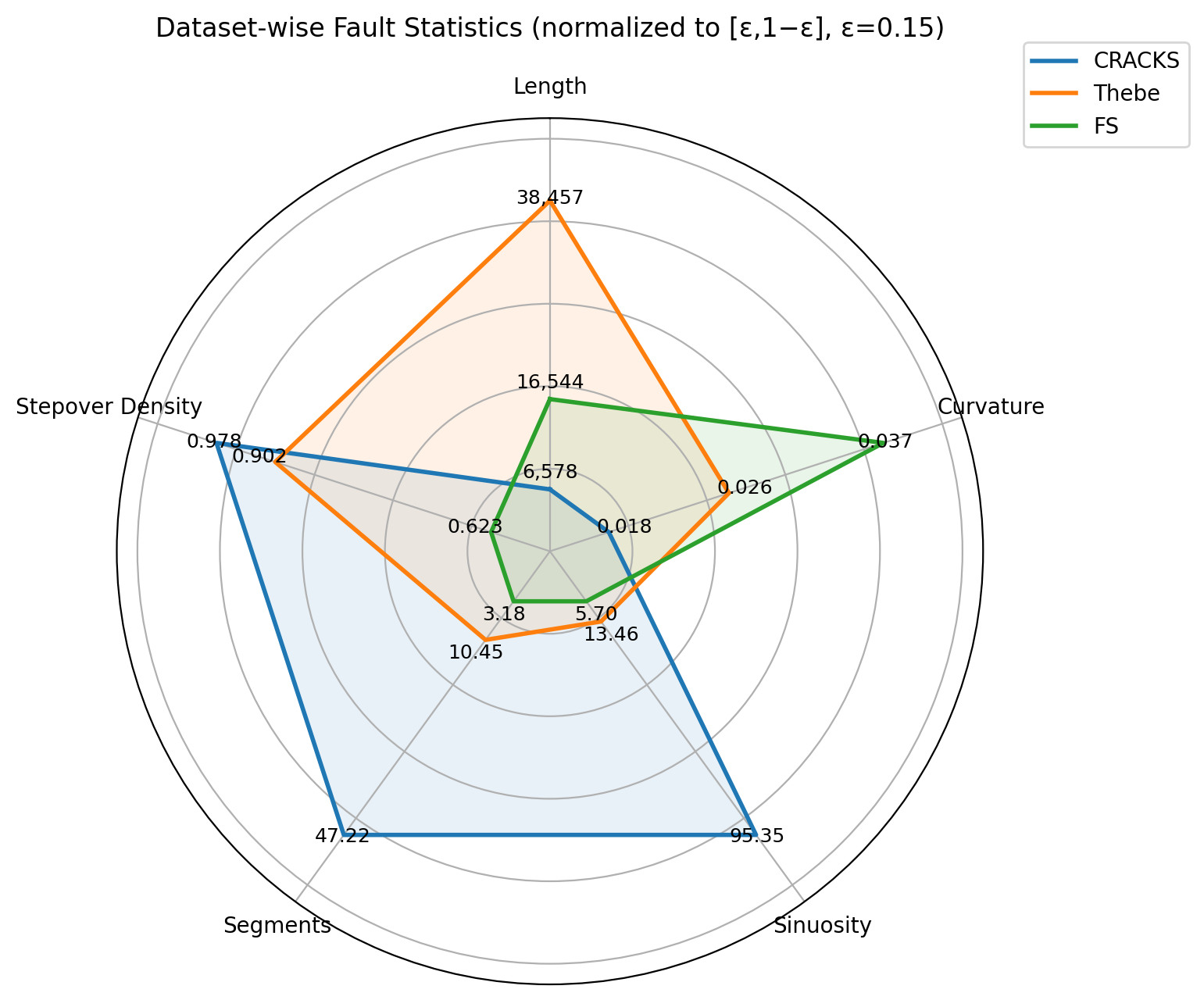} 
        \caption{Fault-oriented metrics.}
        \label{fig:radar-data}
    \end{subfigure}
    %\vspace{5mm} 
    \hfill
    \begin{subfigure}[]{.48\linewidth}
        %\centering
        \includegraphics[width=\textwidth]{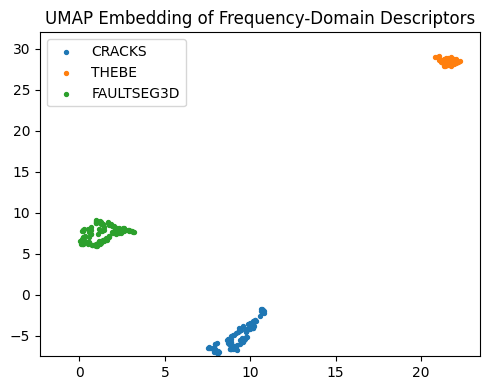}  
        \caption{Spectral descriptors.}
        \label{fig:char-freq}
    \end{subfigure}

\caption{Charaterization of the three considered datasets using (a) fault-oriented metrics  and  (b) spectral descriptors.}
\end{figure}

Among all $74$ datasets used for fault delineation \cite{an2023current}, only $4$ field datasets (\texttt{LANDMASS} \cite{alaudah2015curvelet, alaudah2018structure, alfarraj2018multiresolution, long2018comparative, long2015characterization, alaudah2018learning, aribido2020self, aribido2021self}, \texttt{GSB} \cite{di2019improving, di2018seismic, an2021deep}, \texttt{Thebe} \cite{an2021deep, an2020overlap}, \texttt{CRACKS} \cite{prabhushankar2024cracks})  and $4$ synthetic datasets (\texttt{FaultSeg3D} \cite{wu2019faultSeg}, \texttt{Bi’s 3D synthetic} \cite{bi2021deep}, \texttt{Wu’s 2D SR} \cite{lin2022automatic}, \texttt{Pochet’s 2D synthetic} \cite{pochet2018seismic}) open-sourced both seismic data and labels. The low ratio of open-source labeled field data hinders the creation of benchmarks for training and evaluation of models. 

Besides the lack of public availability, the different characteristics of the datasets pose challenges to the development of generalizable DL models. Specifically, \texttt{LANDMASS} contains image-level fault labels that cannot be used to numerically evaluate the delineation of pixel-wise faults. \texttt{GSB} contains pixel-wise fault labels annotated on only $5$ crossline sections, which limits the evaluation scalability on large test data. Additionally, it is challenging to achieve generalization using only a small number of labels for finetuning \cite{mustafa2024visual}. In contrast, \texttt{Thebe} provides a large amount of pixel-level geophysicist labels across $1803$ crossline sections. \texttt{CRACKS} provides fault labels of varying quality collected from a group of interpreters, including a geophysicist expert, across $400$ inline sections. 

Among the four publicly available datasets, we select \texttt{Thebe} and \texttt{CRACKS} considering the model development and evaluation at the pixel level. All the seismic sections in \texttt{Pochet’s 2D synthetic} contain only one straight fault crossing the entire section, presenting less diversity in angle and density compared to the faults in \texttt{FaultSeg3D}. Both \texttt{Bi’s 3D synthetic} and \texttt{Wu’s 2D SR} \cite{lin2022automatic} originate from \texttt{FaultSeg3D} using the same synthesizing workflow. Thus, we use \texttt{FaultSeg3D} as a reference synthetic dataset with a diverse set of faults. We showcase the acquisition and geological properties of \texttt{CRACKS} and \texttt{Thebe} in Table \ref{tab:F3_Thebe_comparison}.
%Below we describe and compare the three datasets used for our benchmarking study in detail.

Given that proprietary seismic data is, by definition, not publicly accessible, our study intentionally focuses on open-source datasets to ensure reproducibility and transparency. This choice aligns with established benchmarking practices in machine learning, where the goal is to provide reproducible and extensible frameworks rather than case-specific proprietary analyses. Despite being open-source, the selected datasets span markedly different geological, statistical, and labeling characteristics, including synthetic versus real data, varying signal-to-noise ratios, and both expert and crowdsourced interpretations. These variations provide meaningful heterogeneity for assessing domain shift and model generalization. To illustrate this diversity, Figure~\ref{fig:radar-data} presents a comparative visualization of the three datasets using the geometric and topological fault metrics introduced in Section~\ref{sec:flt_metrics}. These metrics quantitatively capture differences in fault density, continuity, and complexity, reinforcing that the chosen datasets represent heterogeneous fault conditions suitable for benchmarking generalization. We also showcase a Uniform Manifold Approximation and Projection (UMAP) embedding of frequency descriptors for all three dataset in Figure~\ref{fig:char-freq}, which reveals three well-separated clusters, confirming systematic acquisition and processing-driven distributional shifts. Below we describe and compare the three datasets used for our benchmarking study in detail.

\texttt{FaultSeg3D} is a 3D synthetic dataset with $220$ volumes each with dimensions of $128 \times 128 \times 128$~\cite{wu2019faultSeg} . In order to better approximate realistic conditions, the authors added background noise estimated from real seismic volumes. The sampling rate and the frequency of the synthetic data vary across the volume to improve the diversity of the data. An example volume is shown in Fig.~\ref{fig:datase-sub1}.

\texttt{CRACKS} is an open-source dataset with diverse faults delineated across $400$ inline sections of the Netherlands F3 Block~\cite{prabhushankar2024cracks, dgb1987netherlands}. The authors in \cite{alaudah2019machine} open-sourced a fully-annotated 3D geological volume of the Netherlands F3 Block for training different models and comparing the performance with objective metrics. Thus, this volume is one of the most extensively studied geographical zones for developing DL-assisted seismic interpretation frameworks \cite{chowdhury2025unified, benkert2024effective, mustafa2024visual, quesada2024benchmarking, mustafa2024explainable, mustafa2023active, kokilepersaud2022volumetric, benkert2022example, benkert2022reliable, benkert2023samples, zhou2023perceptual, chowdhury2023counterfactual, shafiq2022novel, iqbal2023blind, mustafa2021joint, mustafa2021comparative, mustafa2021man, benkert2021explaining, benkert2021explainable, aribido2021self, aribido2020self, mustafa2020spatiotemporal, mustafa2020joint, soliman2020s, alaudah2019machine, di2019developing, di2019improving, alfarraj2019semisupervised, di2019semi, di2019reflector, alaudah2019facies, mustafa2019estimation, alfarraj2019semi}.  The diverse fault features in the F3 block, including major versus minor faults and varying orientations, make it an excellent seismic dataset to train and evaluate fault delineation models \cite{ishak2018application, safari2023structural}.   However, the annotations in \cite{alaudah2019machine} do not provide pixel-wise fault labels. Thus, \texttt{CRACKS} open-sources fully hand-labeled fault annotations by a group of $32$ interpreters with varying degrees of expertise and a domain expert geophysicist. This dataset not only establishes a standardized benchmark for objective comparison but also can be used to investigate the impact of multiple sets of labels with varying quality. Three sets of fault annotations are used to investigate the impact of training labels with varying quality, including expert labels and two more sets of lower quality labels from two other annotators. Fig.~\ref{fig:datase-sub2} shows an example inline section with expert fault labels from \texttt{CRACKS}. \texttt{CRACKS} and \texttt{FaultSeg3D} share geological similarity in the seismic sections in addition to the similar label density.

\texttt{Thebe} is taken from a seismic survey called Thebe Gas Field in the Exmouth Plateau of the Carnarvan Basin on the NW shelf of Australia~\cite{AN2021107219, an2021deep}. The dataset contains $1803$ labeled crossline sections of size $1737 \times 3174$, making it the largest publicly-available field dataset. The seismic intensity of this dataset exhibits a low variation/standard deviation, which can be observed from the low contrast in Fig.~\ref{fig:datase-sub3}. %The faults are delineated by experts using the professional Petrel labeling software.

\section{Experimental Setup}
\label{sec:experimental_setup}
\begin{figure*}
    \centering
    \includegraphics[width=0.9\linewidth]{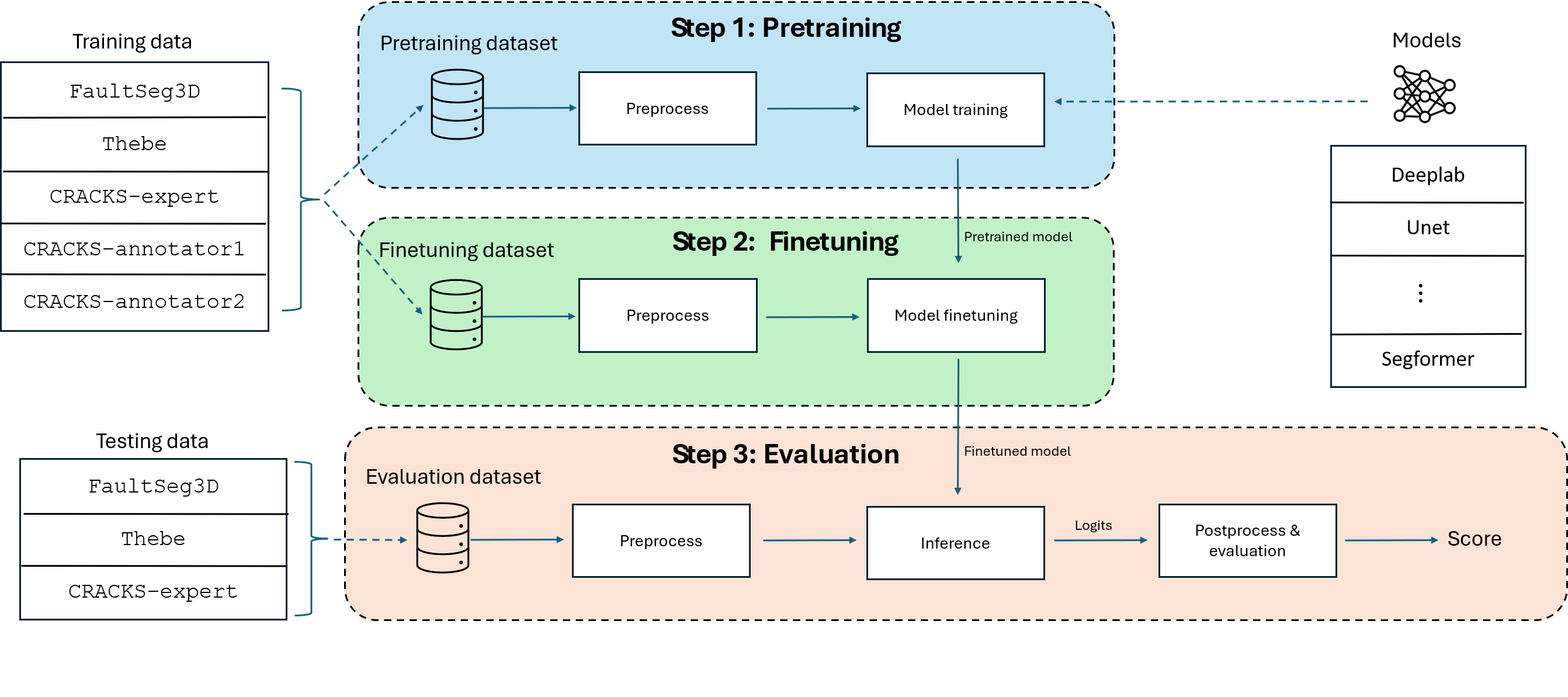}
    \caption{The block diagram of our experimental setup.}
    \label{fig:block_diagram}
\end{figure*}

The workflow of DL-assisted fault delineation involves multiple choices including ($i$) the decision to fine-tune or not, ($ii$) the selection of the datasets for pre-training and fine-tuning, ($iii$) the selection of different models, ($iv$) the development of pre-processing and post-processing strategies, and ($v$) the standardization of the evaluation protocols. We summarize these steps in Fig.~\ref{fig:block_diagram}, where the pipeline is organized chronologically in a top-to-bottom fashion to reflect the order in which decisions are made in practice. This systematic layout enables us to holistically compare the effect of each component and their combinations. We provide details on each stage of the pipeline in the remainder of this section.

\subsection{Data Preparation}
\subsubsection{Standardizing Fault Annotations using Morphological Operations}
\label{sec:exp_std_morph}

% There exist significant differences between the thickness of fault annotations in \texttt{CRACKS}, \texttt{Thebe},  and \texttt{FaultSeg3D}. We show three labeled patches of the same size from \texttt{Thebe}, \texttt{CRACKS}, and \texttt{FaultSeg3D} in Fig.~\ref{fig:all_patch_fault}. The faults manually delineated by interpreters in \texttt{Thebe} and \texttt{CRACKS} are thicker than the synthesized faults in  \texttt{FaultSeg3D}. The inconsistent thickness of fault annotations often influences model optimization, which typically uses pixel-based loss functions. However, existing workflows neglect this fault annotation inconsistency \cite{an2023understanding, lin2022automatic, an2021deep, an2020overlap}. The synthesized faults in \texttt{FaultSeg3D} precisely delineate the thin fractures in the seismic volumes. We therefore standardize the fault thickness across datasets by processing the manually delineated faults in \texttt{Thebe} and \texttt{CRACKS} to have the same thickness of the faults in \texttt{FaultSeg3D}. Specifically, we apply a simple sequence of morphological operations: we first skeletonize the raw fault annotations and then apply dilation with a rank-3 structural element to uniformize fault thickness and close any small gaps, as shown in Fig.~\ref{fig:all_patch_faultMorphed}. This process is denoted by the Preprocess block in Fig.~\ref{fig:block_diagram}. We use standard Python libraries \verb|scikit-image| and \verb|scipy| for these morphological operations.

\begin{figure}
    \centering
    \begin{subfigure}[t]{\linewidth}
        \centering
        \includegraphics[width=\textwidth]{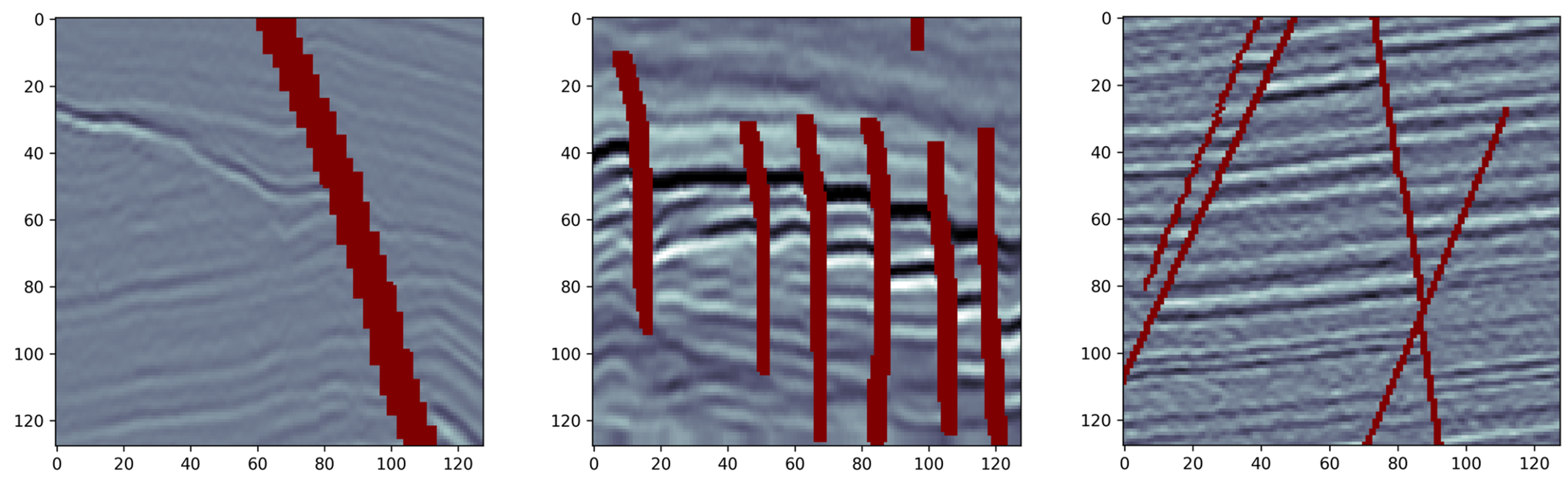} 
        \caption{Fault labels before morphological operations}
        \label{fig:all_patch_fault}
    \end{subfigure}
    \vspace{5mm} 
    \begin{subfigure}[t]{\linewidth}
        \centering
        \includegraphics[width=\textwidth]{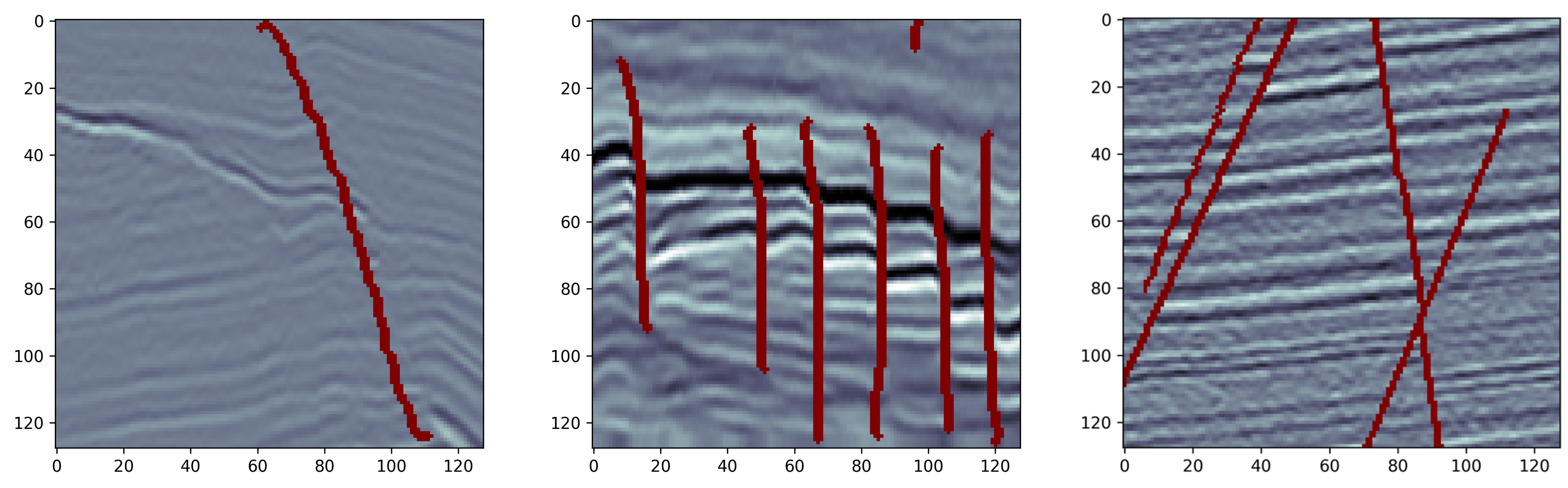}  
        \caption{Fault labels after morphological operations}
        \label{fig:all_patch_faultMorphed}
    \end{subfigure}

\caption{Example $128\times 128$ patches of fault annotations with varying thickness in the three datasets. Left: \texttt{Thebe}, middle: \texttt{CRACKS}, right: \texttt{FaultSeg3D}}
%     \label{fig:}
\end{figure}

There exist significant differences between the thickness of fault annotations in \texttt{CRACKS}, \texttt{Thebe}, and \texttt{FaultSeg3D}. As shown in Fig.\ref{fig:all_patch_fault}, the manually delineated faults in \texttt{Thebe} and \texttt{CRACKS} are considerably thicker than the synthesized faults in \texttt{FaultSeg3D}. Such inconsistency can systematically bias model training and evaluation, especially under pixel-wise loss functions such as the Dice loss. In particular, Dice loss computes similarity between predicted and reference masks by balancing overlap against the number of positive pixels in each mask. Thicker annotations artificially increase the proportion of positive pixels, making the loss less sensitive to small spatial deviations and allowing models to achieve high scores without precisely localizing the fault centerline. Conversely, thinner annotations lead to a stronger penalization of misalignments, requiring sharper localization for similar Dice scores.

Without standardization, these annotation thickness differences could confound cross-dataset comparisons in our benchmark: a model evaluated on thicker annotations may appear more accurate than one evaluated on thinner ones, even if their true localization capability differs. To mitigate this bias, we standardize fault thickness across datasets by processing the manually delineated faults in \texttt{Thebe} and \texttt{CRACKS} to match the thinner style of \texttt{FaultSeg3D}. Specifically, we skeletonize the raw annotations and then apply dilation with a rank-3 structural element to produce uniform thickness and close small gaps, as illustrated in Fig.\ref{fig:all_patch_faultMorphed}. This preprocessing, shown in the Preprocess block in Fig.\ref{fig:block_diagram}, is implemented with \verb|scikit-image| and \verb|scipy|.

\subsubsection{Training and Test Splits}
To ensure meaningful evaluations, we adopt a consistent splitting strategy across the considered datasets. Specifically, we maximize diversity in the test set while ensuring no overlap with the training data. For each dataset, we select spatially distinct subsets to prevent redundancies and simulate deployment conditions on previously unseen segments. In \texttt{CRACKS}, which consists of 400 contiguous inlines, we designate the first 30 and last 30 sections as the test set, totaling 60 sections. The remaining 340 central sections are used for training. This split captures geological variability across the volume while maintaining spatial separation between training and test data. For \texttt{Thebe}, we use 400 sections for training and reserve 100 sections from other parts of the volume for testing, following a similar diversity-maximizing approach. For \texttt{FaultSeg3D}, we follow the established setup from \cite{wu2019faultSeg}, using 200 synthetic volumes for training and 20 distinct volumes for testing.

%While there are different approaches to split the contiguous $400$ sections in \texttt{CRACKS} for training and testing, we aim to maximize data diversity in the test set while creating non-overlapping splits. Thus, our test set consists of the first $30$ and the last $30$ sections in the volume, and the remaining $340$ sections in the middle of the volume is the training set. Consequently, the test data is not biased towards either part of the entire volume. In order to maintain a similar amount of training and test data,  we use $400$ sections as the training set and $100$ sections as the test set for \texttt{Thebe}. For \texttt{FaultSeg3D}, we follow the common protocol \cite{wu2019faultSeg} that uses $200$ cubes as the training set and the other $20$ cubes as the test set. 

\subsection{Model setup}
 
We consider eight segmentation architectures that are among the most widely used in the seismic interpretation literature \cite{an2021deep, lin2022automatic, rs16050922, 10042020}. This selection was designed to cover the set of architectures commonly adopted for fault delineation, spanning both classical convolutional models and more recent transformer-based variants. Specifically, we include:
\begin{itemize}
    \item \textsf{deeplab}: Deeplab \cite{chen2018encoder} architecture with a Resnet50 \cite{he2016deep} backbone
    \item \textsf{deeplab-m}:Deeplab architecture with a Mobilenet \cite{sandler2018mobilenetv2} backbone
    \item \textsf{hed}: Hollistically-nested edge detection model \cite{xie2015holistically}
    \item \textsf{rcf}: Richer convolutional features \cite{liu2017richer}
    \item  \textsf{unet}: Unet \cite{ronneberger2015u} architecture with a Resnet50 backbone
    \item \textsf{unet++}: Unet++  \cite{zhou2018unet++} architecture with a Resnet50 backbone
    \item \textsf{unet-b}: Original Unet architecture as presented in \cite{ronneberger2015u} 
    \item \textsf{segformer}: Transformer-based model introduced in \cite{xie2021segformer}
\end{itemize}

For each of these models, we perform all pairwise combinations of pretraining-finetuning settings between the $5$ sets of labeled training data in the top left of Fig.~\ref{fig:block_diagram}. For example, a model is first pretrained on \texttt{CRACKS}-expert and then finetuned on all $5$ datasets, resulting in $25$ sets of weights for each considered model. Each of these model weights is then evaluated on \texttt{CRACKS}, \texttt{Thebe}, and \texttt{FaultSeg3D}. To complete our baselines, we also use four ImageNet-pretrained models and finetune them on our seismic datasets.

\subsection{Evaluation Setup}
\subsubsection{Post-processing for Model Predictions}

Common practices of evaluating deep fault delineation networks involve two steps \cite{mustafa2024visual, LI2023105412, an2020overlap, an2021deep}: $(i)$ thresholding the network predictions to obtain binary outputs, and $(ii)$ comparing the binary outcomes with the fault test labels using a metric. The test labels are processed with the same morphological operations as the training faults in order to achieve consistent fault thickness at the input and output of a model. Consequently, the thresholding in step $(i)$ needs to be adaptive to the model and the data accordingly, followed by the same morphological operations for numerical evaluation. We compute the optimal threshold for the dataset using the Optimal Dataset Scale (ODS) metric \cite{an2023understanding}. For a model, its optimal threshold is computed using the training set, which is then applied to binarize the predictions on the test data, followed by the same morphological operation for evaluation.

\subsubsection{Performance Metrics}
As mentioned in Section~\ref{sec:rw_metrics}, pixel-based and distance-based metrics capture different aspects of prediction quality, and each can fail to fully reflect the structural accuracy of the predicted faults, making the evaluation of fault delineation methods an inherently challenging task.
%evaluating the faults predicted by a model is a challenging task due to their thin continuous structure, which introduces class-imbalance and sensitivity to small errors. Both pixel-based and distance-based metrics can fail to fully reflect the structural accuracy of the predicted faults.
Our benchmark study thus adopts a holistic approach to assess prediction quality by considering a combination of multiple metrics alongside subjective inspection. In this study we choose to report one pixel-based metric, Dice Coefficient (defined in Eq.~(\ref{eq:dice})) and two distance-based metrics: BCD and the modified Hausdorff (defined in Eq. \ref{eq:bcd}, and Eq. \ref{eq:hd}, respectively). These metrics have not only been extensively used in the seismic literature for model performance assessment \cite{wu2019faultSeg, guillon2020ground, 576361}, but allow for two different evaluation axes: pixel overlap-based and structure-based. 

\subsubsection{Fault Characteristic Comparisons}
While the individual metrics described in Section \ref{sec:flt_metrics} quantify dataset characteristics in isolation, comparative metrics evaluate the similarity between predicted and ground-truth faults. We describe below the metrics we use in Table \ref{tab:training_stats} to extend the individual descriptors into differences, ratios, and statistical comparisons.

\textbf{Strike Similarity ($\text{StrikeSim}$)} measures the similarity of orientation distributions between predicted ($P(\theta)$) and ground truth ($G(\theta)$) faults using cosine similarity:
\begin{equation}
\text{StrikeSim} = \frac{\sum_{\theta} P(\theta) G(\theta)}{\sqrt{\sum_{\theta} P(\theta)^2} \, \sqrt{\sum_{\theta} G(\theta)^2}}.
\end{equation}

\textbf{Curvature Metrics} compare curvature distributions across predictions and ground truth. These include:
\begin{align}
\Delta \kappa &= \big| \overline{\kappa}_{\text{pred}} - \overline{\kappa}_{\text{gt}} \big| , \\
\text{RMSE}_\kappa &= \sqrt{ \frac{1}{N} \sum_{i=1}^N \left(\kappa_i^{\text{pred}} - \kappa_i^{\text{gt}}\right)^2 } , \\
\text{Corr}_\kappa &= \frac{\text{Cov}\!\left(\kappa^{\text{pred}}, \kappa^{\text{gt}}\right)}{\sigma_{\kappa^{\text{pred}}}\sigma_{\kappa^{\text{gt}}}} .
\end{align}

\textbf{Sinuosity Metrics} evaluate differences in tortuosity:
\begin{align}
\Delta S &= \overline{S}_{\text{pred}} - \overline{S}_{\text{gt}} , \\
R_S &= \frac{\overline{S}_{\text{pred}}}{\overline{S}_{\text{gt}}} .
\end{align}

\textbf{Length Metrics} evaluate differences in fault trace length:
\begin{align}
\Delta L &= L_{\text{pred}} - L_{\text{gt}} , \\
R_L &= \frac{L_{\text{pred}}}{L_{\text{gt}}} .
\end{align}

\textbf{Segment Metrics} compare the number of interpreted fault segments:
\begin{align}
\Delta N_{\text{seg}} &= N_{\text{seg}}^{\text{pred}} - N_{\text{seg}}^{\text{gt}} , \\
R_{N_{\text{seg}}} &= \frac{N_{\text{seg}}^{\text{pred}}}{N_{\text{seg}}^{\text{gt}}} .
\end{align}

\textbf{Stepover Metrics} compare the relative frequency of stepovers:
\begin{align}
\Delta D_{\text{stepover}} &= D_{\text{stepover}}^{\text{pred}} - D_{\text{stepover}}^{\text{gt}} , \\
R_{D_{\text{stepover}}} &= \frac{D_{\text{stepover}}^{\text{pred}}}{D_{\text{stepover}}^{\text{gt}}} .
\end{align}

By combining individual metrics with their comparative counterparts, we assess not only whether faults are detected, but also whether their structural and geometric properties are faithfully reproduced.

\section{Results}
\label{sec:results}
The results of our benchmarking experiments are analyzed across three different thematic axes: (1) generalizability and transferability, (2) training dynamics, and (3) metric evaluation. %A summarized account of our key findings and observations is presented in Table \ref{tab:key_observations}. %; and a metrics-based axis, in which we aim to understand the benefits and shortcomings of our benchmarking setup by looking into what each metric optimizes for. 

\begin{figure}
    \centering
    \includegraphics[width=\linewidth]{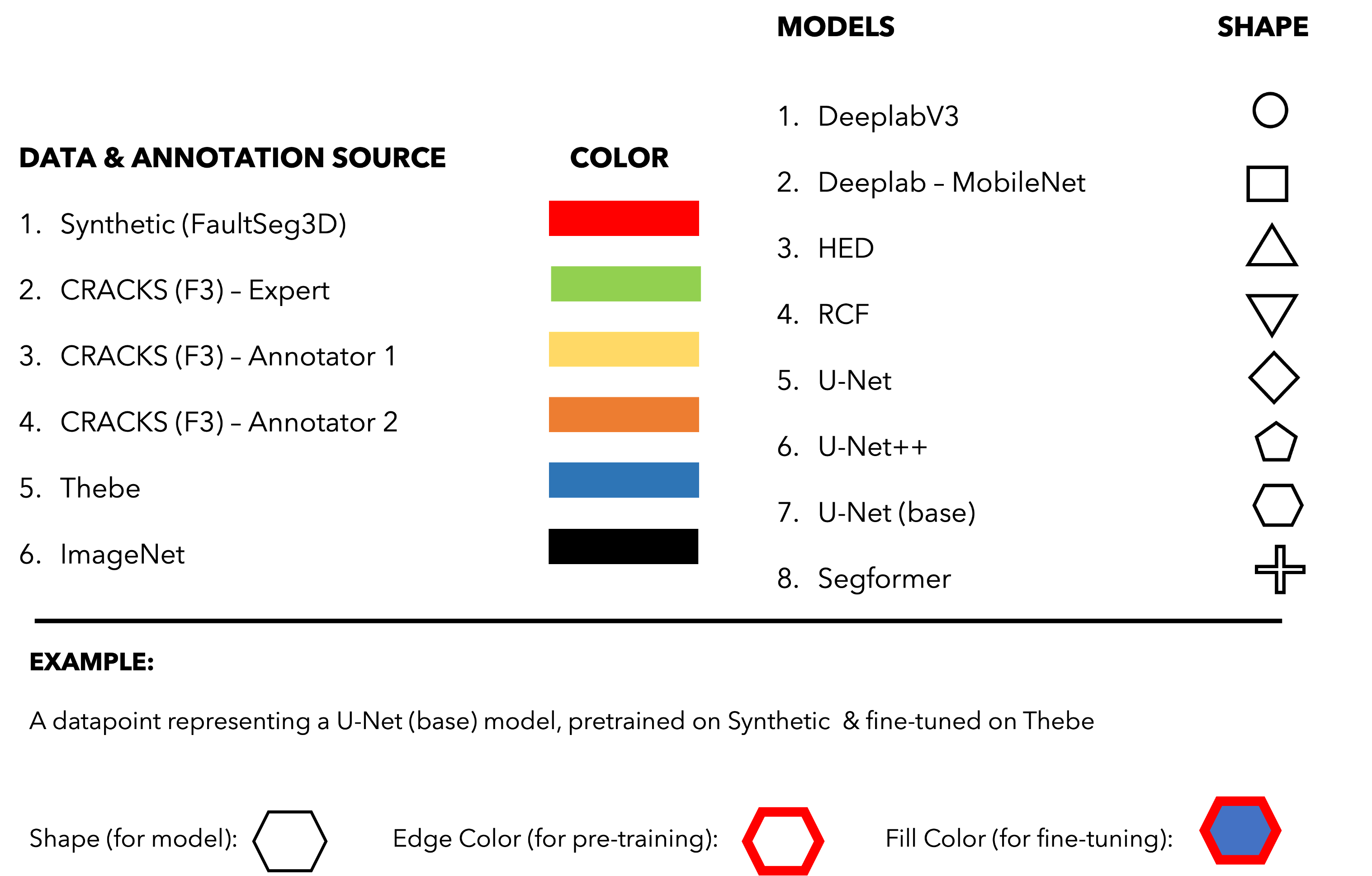}
    \caption{Visual encoding to represent the different models pretrained and finetuned on different datasets. Border color signifies the pretraining data source and the fill color signifies the fine-tuning dataset. Models are represented using various shapes.}
    \label{fig:legend}
\end{figure}

\subsection{Generalization and Transferability}

As mentioned in Section~\ref{sec:introduction}, models are often used to process data from new surveys that can differ from their original training sources. As such, generalization and transferability are critical for reliable deployment and to reduce the labeling overhead. Understanding how various training regimes perform across domains is key to developing scalable workflows.

In this subsection, the generalization of models trained under different pretraining-finetuning regimes is analyzed across \texttt{CRACKS}, \texttt{FaultSeg3D}, and \texttt{Thebe}. Unless stated otherwise, we structure our scatter plots using the convention depicted in Fig.~\ref{fig:legend}. The shape of a given point encodes the model used, the border color corresponds to the dataset used for pretraining the model, and the fill color corresponds to the datasets used for finetuning the model. All of our reported figures and plots correspond to models being evaluated in a held-out test partition of one of these 3 datasets.

\subsubsection{Dataset Alignment and Transfer Trends}
\begin{figure*}
    \centering
    \begin{subfigure}{0.3\linewidth}
        \includegraphics[width=\linewidth]{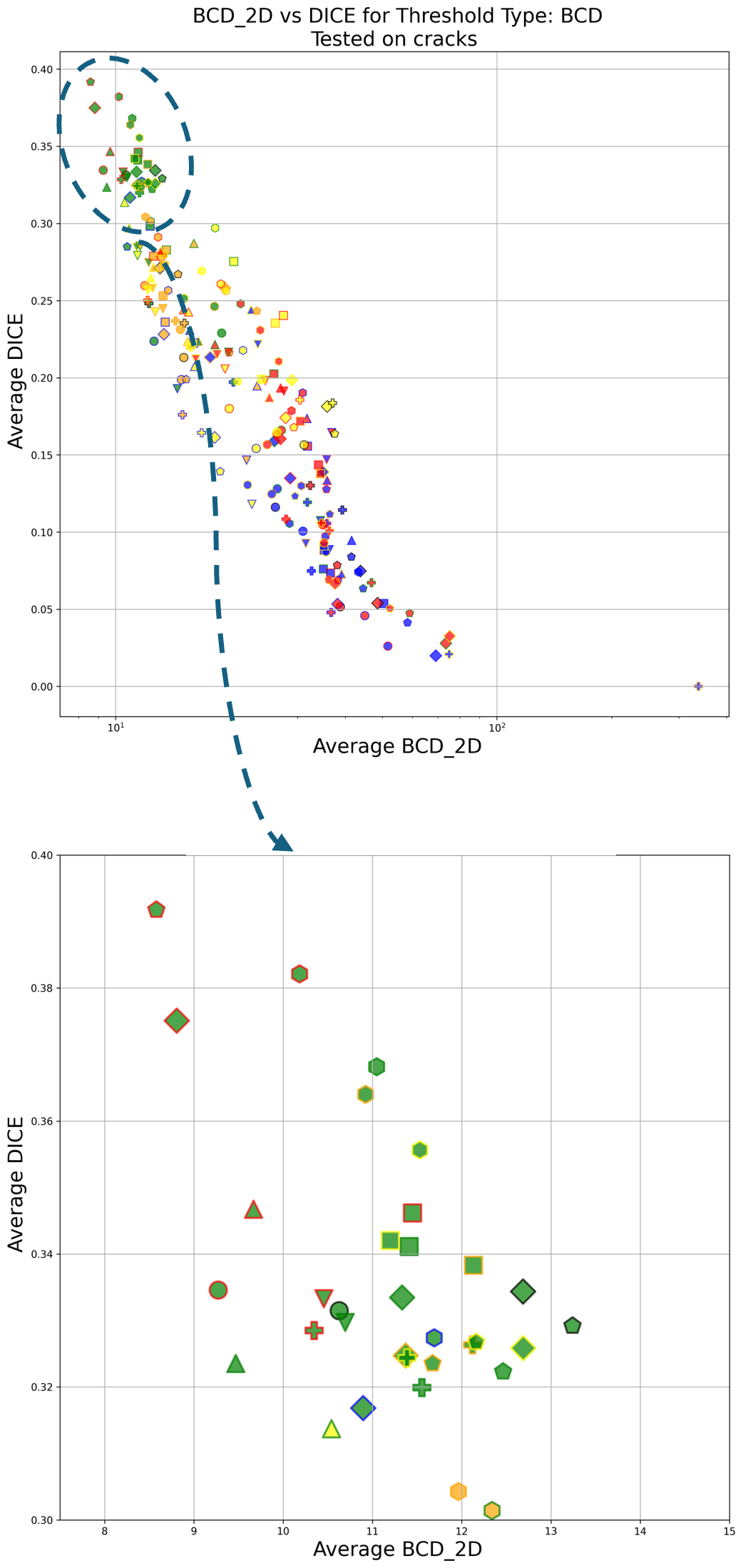}
        \caption{Test on \texttt{CRACKS} dataset}
        \label{fig:CRACKS}
    \end{subfigure}
    \hfill
    \begin{subfigure}{0.3\linewidth}
        \includegraphics[width=\linewidth]{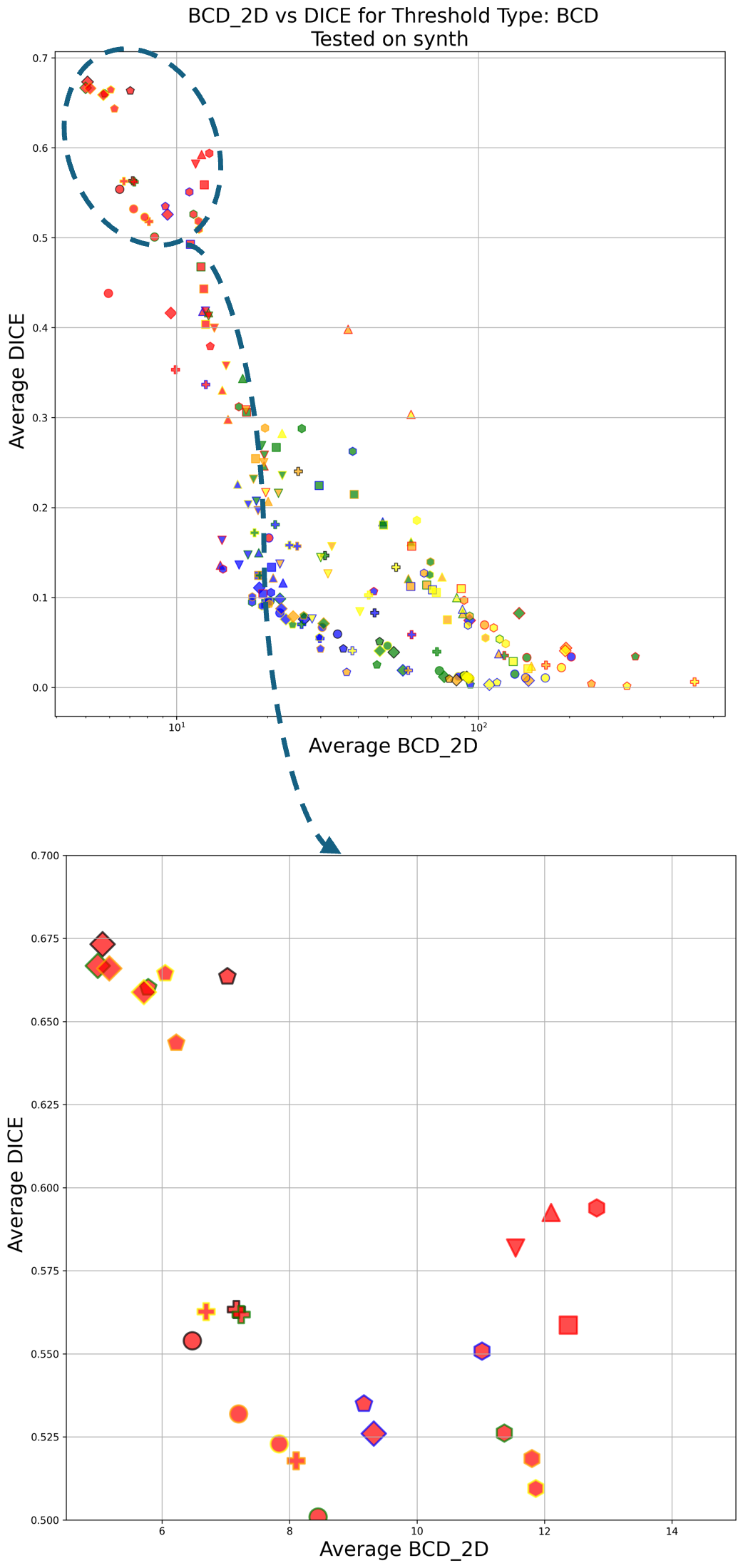}
        \caption{Test on \texttt{FaultSeg3D} dataset}
        \label{fig:synth}
    \end{subfigure}
    \hfill
    \begin{subfigure}{0.3\linewidth}
        \includegraphics[width=\linewidth]{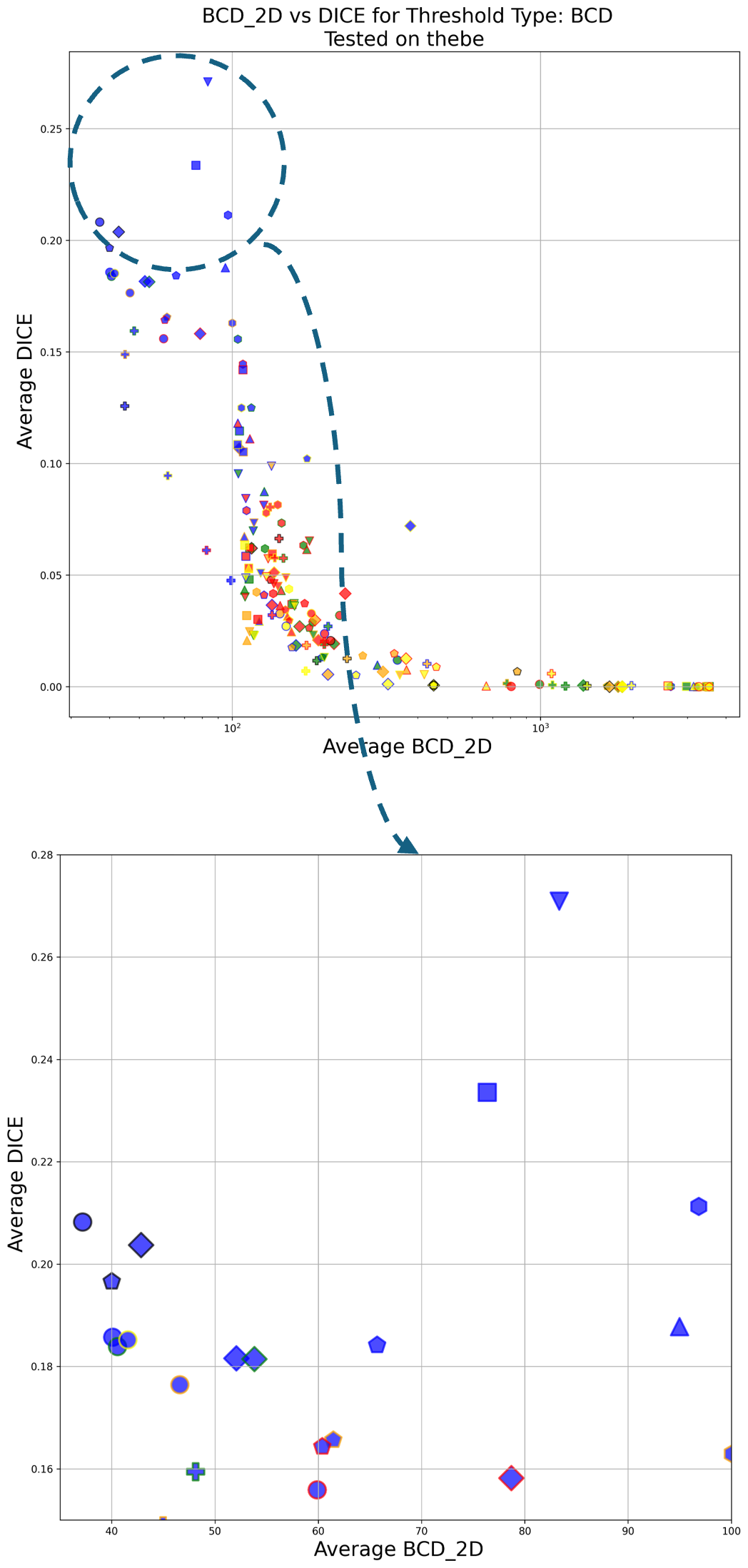}
        \caption{Test on \texttt{Thebe} dataset}
        \label{fig:thebe}
    \end{subfigure}
    \caption{Top: All of out models across our different pretraining and finetuning setups. Bottom: Best performing models for each dataset.}
    \label{fig:macro_zoom}
\end{figure*}

Fig.~\ref{fig:macro_zoom} provides a macroscopic view of the model behaviors across the different training setups. On \texttt{CRACKS} (Fig. \ref{fig:CRACKS}), the top-performing configurations are those pretrained on \texttt{FaultSeg3D} data and fine-tuned on \texttt{CRACKS}. The observation indicates strong geophysical commonalities between \texttt{FaultSeg3D} and \texttt{CRACKS} data, and supports the utility of \texttt{FaultSeg3D} data as a viable pretraining source when real annotations are limited. The fill-color distribution in Fig.~\ref{fig:CRACKS} also establishes a hierarchy of effective fine-tuning datasets for \texttt{CRACKS}: \texttt{CRACKS} > \texttt{\texttt{FaultSeg3D}} > \texttt{Thebe}. The poor finetuning performance on \texttt{Thebe} indicates that it is more distributionally distant from \texttt{CRACKS} than \texttt{FaultSeg3D} data.

Furthermore, when tested on \texttt{FaultSeg3D} data (Fig.~\ref{fig:synth}), models pretrained on \texttt{CRACKS} again outperform those trained from scratch, indicating that the aforementioned alignment is reciprocal. However, for \texttt{Thebe} (Fig.~\ref{fig:thebe}), top-performing models are those trained from scratch on \texttt{Thebe} itself. Transferring from either \texttt{\texttt{FaultSeg3D}} or \texttt{\texttt{CRACKS}} results in performance degradation, suggesting that \texttt{Thebe} resides in a distinct feature space.

% \begin{figure}[htbp]
%     \centering
%     \includegraphics[width=0.45\textwidth]{Figures/CRACKS.png}
%     \caption{Test on \texttt{CRACKS} dataset}
%     \label{fig:CRACKS}
% \end{figure}

% \begin{figure}[htbp]
%     \centering
%     \includegraphics[width=0.45\textwidth]{Figures/synth.png}
%     \caption{Test on \texttt{FaultSeg3D} dataset}
%     \label{fig:synth}
% \end{figure}

% \begin{figure}[htbp]
%     \centering
%     \includegraphics[width=0.45\textwidth]{Figures/thebe.png}
%     \caption{Test on \texttt{Thebe} dataset}
%     \label{fig:thebe}
% \end{figure}

\subsubsection{Domain Shift and Joint Training}

Domain shift is a pressing challenge in seismic DL, particularly when deploying models across surveys with different geological properties. When overlooked, this can lead to brittle models that perform well on training data but fail on new volumes, or models that catastrophically forget features from their original data after finetuning. An illustration of this catastrophic forgetting occurs when models pretrained on \texttt{CRACKS} are fine-tuned on \texttt{Thebe}: despite reasonable performance on \texttt{Thebe}, these models experience dramatic performance degradation when re-evaluated on their original domain. In the case of \textsf{unet++}, for instance, the Dice score on \texttt{CRACKS} drops from 0.34 to 0.12, while the BCD increases twofold, indicating a complete erasure of useful representations, and clear case of catastrophic forgetting. A statistical analysis shows that \texttt{Thebe} has a significantly lower standard deviation \textbf{(0.124)} compared to \texttt{CRACKS} \textbf{(1.149)} and \texttt{FaultSeg3D} \textbf{(1.052)}. The contrast and intensity variations in \texttt{Thebe} are lower, while \texttt{CRACKS} and \texttt{FaultSeg3D} have diverse intensity distributions, as shown in Fig. \ref{fig:datasets}. An explanation for this catastrophic forgetting phenomenon is that models trained on \texttt{Thebe} learn from a constrained input range and struggle when tested on datasets with richer intensity distributions. These results suggest that other normalization techniques before training could help reduce these distributional mismatches.

%An explanation for the degree to which finetuning on \texttt{Thebe} degrades a model’s performance on the same testing setup has to do with the statistical properties of the dataset. A statistical analysis shows that \texttt{Thebe} has a significantly lower standard deviation \textbf{(0.124)} compared to \texttt{CRACKS} \textbf{(1.149)} and \texttt{FaultSeg3D} \textbf{(1.052)}. The contrast and intensity variations in \texttt{Thebe} are lower, while \texttt{CRACKS} and \texttt{FaultSeg3D} have diverse intensity distributions, as shown in Figure \ref{fig:datasets}. As a result, models trained on \texttt{Thebe} learn from a restricted input range and struggle when tested on datasets with richer intensity distributions. A \texttt{CRACKS}-pretrained model which is then fine-tuned on \texttt{Thebe} might forget features useful for \texttt{CRACKS}, a case of catastrophic forgetting confirmed by sharp DICE drops and BCD increases. Thus, domain shift in seismic segmentation is caused by data distribution differences rather than feature learning alone.

 %Techniques like global min-max normalization or adaptive histogram equalization may better align dataset statistics. 

To further explore mitigation strategies for catastrophic forgetting, we conducted a focused case study using Elastic Weight Consolidation (EWC) with the \textsf{deeplabv3} (ResNet-50 backbone) architecture, when transferring from the synthetic \texttt{FaultSeg3D} dataset to the real \texttt{CRACKS} and \texttt{Thebe} volumes. The EWC regularizer constrains the update of parameters that are important for the source task, aiming to preserve previously acquired knowledge during fine-tuning. As shown in Table~\ref{tab:ewc}, the expected behavior holds for the \texttt{FaultSeg3D} to \texttt{CRACKS} setting: performance on the target domain decreases slightly relative to standard fine-tuning, but performance on the source domain improves, indicating partial retention of source information. However, this pattern does not hold for the \texttt{FaultSeg3D} to \texttt{Thebe} case, where performance drops on both domains. This outcome is consistent with our broader observations that \texttt{Thebe} represents a strong outlier whose distribution is sufficiently distinct that attempts to retain source-domain information hinder learning of its features altogether. These results highlight that a driving factor in catastrophic forgetting is the distributional structure of the source-target pair.

The empirical pattern we observe (severe performance loss on the source domain after fine-tuning on Thebe, and the mixed effect of EWC across transfer pairs) can be explained by the interaction of three seismic-specific factors. First, faults are sparse, line-like signals: a network’s predictive capacity depends on a small number of localized, edge-sensitive features. Second, amplitude and contrast statistics differ markedly between datasets (e.g., Thebe’s standard deviation = 0.124 vs. CRACKS = 1.149 and FaultSeg3D = 1.052; see Fig.~\ref{fig:datasets}); such compression of dynamic range alters internal activation distributions (batchnorm and convolutional responses), rendering source-trained low-level filters less discriminative. Third, regularization schemes like EWC constrain parameters important for the source via a Fisher-based penalty: this helps when the target is moderately different (it preserves source knowledge while allowing limited adaptation, as in \texttt{FaultSeg3D} to \texttt{CRACKS}), but it can prevent learning altogether when the target is an extreme outlier (as in \texttt{FaultSeg3D} to \texttt{Thebe}), producing simultaneous degradation on both domains. Together these effects make seismic catastrophic forgetting both more likely and more abrupt than in many dense, object-centric vision tasks. Practical mitigations that follow from this analysis potentially include input-level alignment (histogram matching or style transfer), adaptive normalization layers, selective (layer-wise) fine-tuning, rehearsal or pseudo-replay, and capacity-aware choices (larger models or explicit adaptation for small models).
 
 As an additional baseline, experiments on joint training on all possible combinations of datasets in a single training round are conducted using the \textsf{unet} model with a ResNet50 backbone. The results for all \textsf{unet} experiments, including these joint training configurations, are shown in Table \ref{tab:full_performance}. The results further showcase that features learned from \texttt{CRACKS} and \texttt{FaultSeg3D} data align together, but these learned features are not easily transferable to \texttt{Thebe}. On the other hand, \texttt{Thebe} may act as a regularizer when paired with a large dataset like \texttt{FaultSeg3D}, hurting performance on seen datasets (compared to training them individually) but providing modest generalization to unseen ones: performance on the unseen dataset for this setting (i.e. \texttt{CRACKS}) does not drop as badly as that of the unseen dataset for other joint settings (e.g. \texttt{FaultSeg3D + CRACKS}).
 
As an additional step to assess the robustness and reliability of our results, we also perform our \textsf{unet} experiments on \texttt{FaultSeg3D} using 5 fold cross validation, which we report on Table \ref{tab:cross_valid}. We can observe that the results in these experiments deviate minimally from the ones we originally reported in Table \ref{tab:full_performance}, showcasing that our experimental framework is robust against stochasticity.

\begin{table*}[ht]
\centering
\caption{Performance Metrics (DICE, BCD, Hausdorff) for different training schemes on Unet. All values are rounded to three decimals. Schemes are ranked as best, second best or worst based on satisfying at least two of the three metric criteria, to account for cases where metrics disagree. (\textcolor{SkyBlue!100}{blue} for highest, \textcolor{pink!100}{pink} for second, and \textcolor{gray!100}{gray} for the worst.)}

\label{tab:full_performance}
\resizebox{\textwidth}{!}{%
\begin{tabular}{lccc|ccc|ccc}
\toprule
\multirow{2}{*}{\textbf{Training Configuration}} 
 & \multicolumn{3}{c|}{\textbf{Test on} \texttt{FaultSeg3D}}
 & \multicolumn{3}{c|}{\textbf{Test on} \texttt{Cracks}} 
 & \multicolumn{3}{c}{\textbf{Test on} \texttt{Thebe}} \\
\cmidrule(lr){2-4} \cmidrule(lr){5-7} \cmidrule(lr){8-10}
 & DICE & BCD & Hausdorff 
 & DICE & BCD & Hausdorff 
 & DICE & BCD & Hausdorff \\
\midrule
\multicolumn{10}{c}{\textbf{Individual Training}} \\

\texttt{Thebe} 
 & 0.111 (+0.0347 , -0.0380)
 & 18.767 (+33.9692 , -6.4945)
 & 16.095 (+33.6256 , -6.3892)
 & \cellcolor{gray!30}0.020 (+0.0082 , -0.0077)
 & \cellcolor{gray!30}69.043 (+17.8047 , -13.3343)
 & \cellcolor{gray!30}49.012 (+19.7797 , -12.1662)
 & \cellcolor{SkyBlue!30}0.182 (+0.0230 , -0.0158)
 & \cellcolor{SkyBlue!30}52.054 (+7.3546 , -8.1523)
 & \cellcolor{SkyBlue!30}31.615 (+8.2126 , -4.7123)
\\[2mm]

\texttt{FaultSeg3D} 
 & 0.4165 (+0.0473, -0.0717) 
 & 9.5632 (+14.6540, -5.2962) %0.337 (+0.2988 , -0.2512) 
 & 7.1908 (+12.4008, -4.1851)
 & 0.1604 (+0.0479, -0.0445) 
 & 27.0161 (+8.7621, -6.1338)
 & 20.6379 (+8.1430, -5.3172)
 & 0.0417 (+0.0085, -0.0081) 
 & 232.7213 (+29.4637, -28.2944) 
 & 189.9385 (+27.9787, -26.0703)
\\[2mm]

\texttt{Cracks} 
 & 0.012(+0.057 , -0.008)
 & 76.904 (+60.57 , -30.058)
 & 57.798 (+60.57 , -30.058)
 & \cellcolor{pink!30}0.333 (+0.0574 , -0.0819)
 & \cellcolor{pink!30}11.330 (+10.8538 , -3.1426)
 & \cellcolor{pink!30}7.341 (+7.1827 , -2.4529)
 & \cellcolor{gray!30}0.001 (+0.0013 , -0.0007)
 & \cellcolor{gray!30}1377.945 (+550.5007 , -578.69)
 & \cellcolor{gray!30}1022.510 (+222.6350 , -321.1643)
\\
\midrule
\multicolumn{10}{c}{\textbf{Combined Training}} \\

\texttt{FaultSeg3D + Cracks}
 & 0.662 (+0.115 , -0.082)
 & 6.143 (+18.342 , -2.673)
 & 5.282 (+15.365 , -1.890)
 & 0.333 (+0.026 , -0.017)
 & 15.436 (+4.649 , -1.583)
 & 11.597 (+5.264 , -2.119)
 & 0.001 (+0.0012 , -0.0007)
 & 659.708 (+37.582 , -49.906)
 & 430.887 (+31.481, -39.851)
\\[2mm]

\texttt{FaultSeg3D + Thebe}
 & 0.060 (+0.025 , -0.018)
 & 29.834 (+18.619 , -7.824)
 & 20.151 (+14.253 , -5.342)
 & 0.167 (+0.018 , -0.011)
 & 29.717 (+7.782 , -3.947)
 & 26.136 (+6.492 , -2.371)
 & 0.211 (+0.012 , -0.008)
 & 52.952 (+8.459 , -4.671)
 & 43.348 (+7.935 , -3.649)
\\[2mm]

\texttt{Cracks + Thebe}
 & 0.038 (+0.0475 , -0.0236)
 & 34.183 (+17.2146 , -9.1272)
 & 22.751 (+15.6639 , -7.0688)
 & 0.195 (+0.039 , -0.0389) 
 & 17.8 (+6.04 , -3.61) 
 & 13.13 (+6.54 , -3.4)
 & 0.112 (+0.021 , -0.0178)
 & 144.67 (+35.17 , -24.26)
 & 98.25 (+21.13 , -18.07)
\\[2mm]

\texttt{All }
 & 0.337 (+0.2988 , -0.2512) 
 & 28.569 (+55.1384 , -20.0588)
 & 23.674 (+35.9425 , -16.5400)
 & 0.1899 (+0.032 , -0.04)
 & 25.62 (+10.37 , -7.3)
 & 17.88 (+10.7 , -7.3)
 & 0.144 (+0.028 , -0.033)
 & 89.69 (+55.91 , -25.90)
 & 75.85 (+58.49 , -27.09)
\\
\midrule
\multicolumn{10}{c}{\textbf{Fine-Tuning}} \\

\texttt{Cracks $\rightarrow$ FaultSeg3D} 
 & \cellcolor{SkyBlue!30}0.667 (+0.152 , -0.015)
 & \cellcolor{SkyBlue!30}4.994 (+20.04 , -2.603) 
 & \cellcolor{SkyBlue!30}3.821 (+15.34 , -2.923) 
 & 0.028 (+0.0270, -0.0328) 
 & 73.433 (+19.3709, -15.0939) 
 & 45.378 (+16.5661, -12.8940) 
 & 0.027 (+0.0026, -0.0026)
 & 165.315 (+14.13, -10.47)
 & 99.996 (+10.03, -5.91)
\\[2mm]

\texttt{Cracks $\rightarrow$ Thebe} 
 & 0.099 (+0.02, -0.034) 
 & 21.959 (+34.82, -7.034) 
 & 15.771 (+10.93, -4.034)
 & 0.159 (+0.0270, -0.0328) 
 & 26.070 (+8.2453, 6.2880) 
 & 20.929 (+7.4229, -5.2930) 
 & \cellcolor{pink!30}0.181 (+0.0172, -0.0132) 
 & \cellcolor{pink!30}53.808 (+11.815, -9.61) 
 & \cellcolor{pink!30}42.169 (+13.56, -12.09)
\\[2mm]

\texttt{FaultSeg3D $\rightarrow$ Cracks}
 & \cellcolor{gray!30}0.0825 (+0.1060, -0.0587)
 & \cellcolor{gray!30}136.2934 (+256.0289, -72.6493) 
 & \cellcolor{gray!30}134.0948 (+255.0063, -71.8545)
 & \cellcolor{SkyBlue!30}0.3751 (+0.0539, -0.0650) 
 & \cellcolor{SkyBlue!30}8.8064 (+5.0173, -2.6473)   
 & \cellcolor{SkyBlue!30}5.5773 (+3.4140, -2.0301)
 & 0.0193 (+0.0055, -0.0045) 
 & 213.6949 (+21.8752, -18.4942) 
 & 147.2127 (+14.6880, -11.1500)
\\[2mm]

\texttt{FaultSeg3D $\rightarrow$ Thebe}
 & 0.0745 (+0.1185, -0.0550) 
 & 93.4884 (+91.8717, -43.4160) 
 & 84.4202 (+84.7160, -38.6755)
 & 0.1349 (+0.0246, -0.0233) 
 & 28.7014 (+7.7890, -6.2796) 
 & 23.3679 (+8.1318, -6.1587)
 & 0.1582 (+0.0149, -0.0148) 
 & 78.6831 (+48.4932, -25.2388) 
 & 51.6490 (+47.2406, -19.6911)
\\[2mm]

\texttt{Thebe $\rightarrow$ Cracks}
 & 0.019 (+0.0082 , -0.0077)
 & 56.145 (+55.3901 , -18.078)
 & 42.899 (+59.5264 , -15.8497)
 & 0.317 (+0.0454 , -0.0514)
 & 10.896 (+3.7560 , -2.1167)
 & 6.656 (+3.2688 , -1.8525)
 & 0.019 (+0.0052 , -0.0050)
 & 161.131 (+14.8345 , -14.0856)
 & 88.344 (+9.4150 , -9.1093)
\\[2mm]

\texttt{Thebe $\rightarrow$ FaultSeg3D}
 & \cellcolor{pink!30}0.526 (+0.1585 , -0.2580)
 & \cellcolor{pink!30}9.319 (+42.6775 , -6.5765)
 & \cellcolor{pink!30}8.182 (+40.3980 , -5.9642)
 & 0.053 (+0.0116 , -0.0128)
 & 38.141 (+9.1915 , -5.8445)
 & 34.319 (+9.1316 , -5.7956)
 & 0.037 (+0.0029 , -0.0027)
 & 134.522 (+8.3162 , -6.5321)
 & 122.936 (+8.0421 , -6.2061)
\\
\bottomrule
\end{tabular}%
}
\end{table*}

\begin{table}[]
\centering
\caption{U-Net performance on FaultSeg3D dataset using 5-fold cross-validation}
\label{tab:cross_valid}
{%
\begin{tabular}{cc}
\hline
Metric & Result \\ \hline
Dice & $0.4493 \pm 0.0285$ \\
Hausdorff & $6.1464 \pm 0.5806$ \\ 
BCD & $8.6323 \pm 0.7638$ \\ \hline
\end{tabular}
}
\end{table}

% \usepackage{multirow}
\begin{table}[]
\centering
\caption{Results on EWC case study}
\label{tab:ewc}
\resizebox{\linewidth}{!}{%
\begin{tabular}{cccc|ccc}

\multirow{2}{*}{Training configuration} & \multicolumn{3}{c|}{Source} & \multicolumn{3}{c}{Target} \\ \cline{2-7} 
                                        & Dice      & Hausdorff & BCD    & Dice      & Hausdorff & BCD   \\ \hline
\texttt{FaultSeg3D $\rightarrow$ Cracks}         & 0.025     & 35.562 & 48.598      & 0.1998     & 7.5489  & 12.627     \\ \hline
\texttt{FaultSeg3D $\rightarrow$ Thebe}           & 0.012     & 148.567  & 179.181    & 0.061     & 77.232   & 112.493    \\ \hline
\end{tabular}
}
\end{table}

\subsubsection{Model Capacity and Transferability}
\begin{figure}
     \centering
     \includegraphics[width=\linewidth]{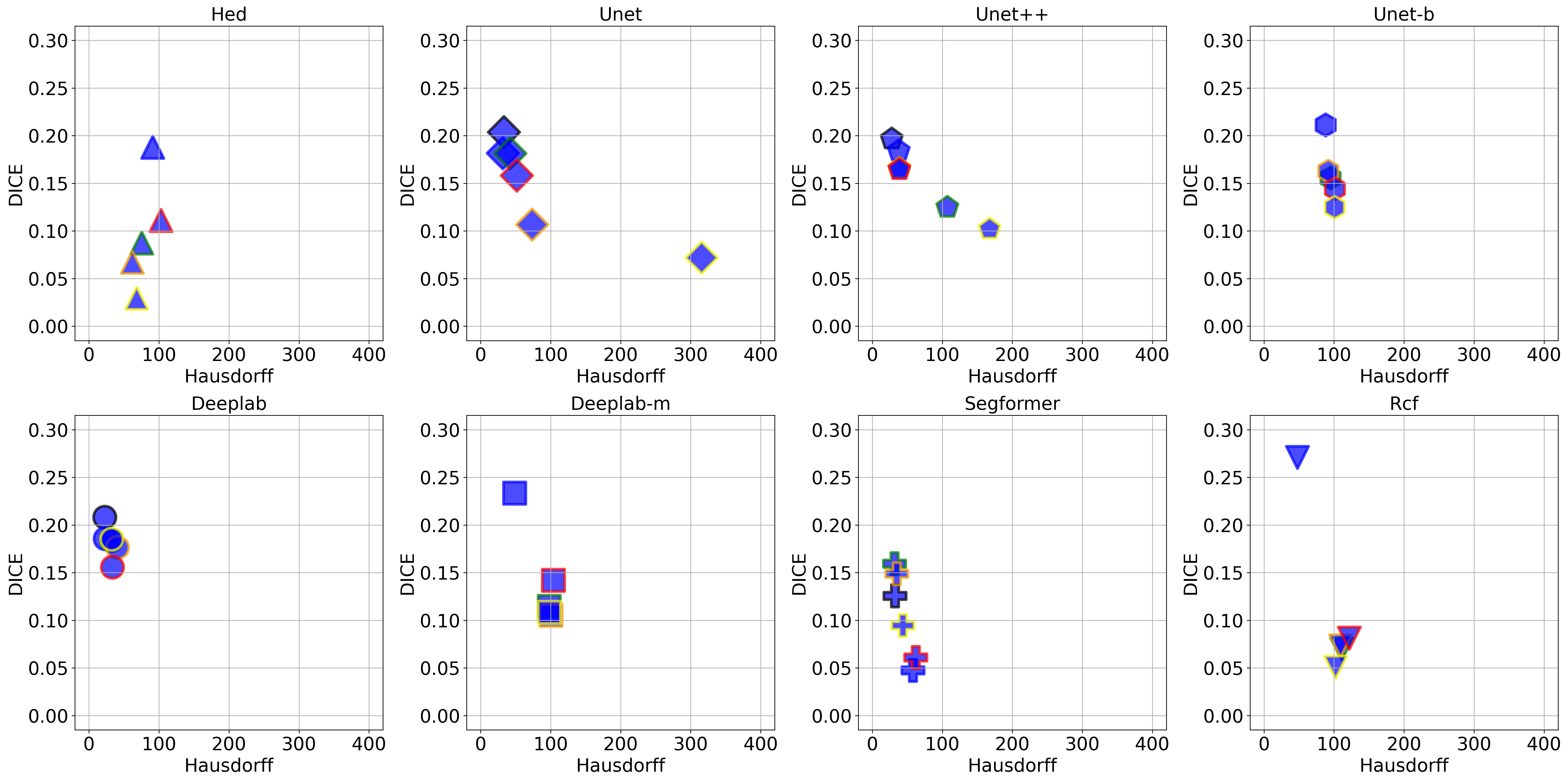}
     \caption{Individual Models tested on \texttt{Thebe}. Models with blue edges are pretrained on \texttt{Thebe} without finetuning. While other models are pretrained on different datasets. We show that pretraining on another faults dataset is not beneficial compared to using \texttt{Imagenet} weights or training from randomly initialized models.}
          \label{fig:thebe_models}
 \end{figure}

It is generally established in the literature that pretraining on a large dataset can boost the performance of a model even in self-supervised settings \cite{rs16050922}. %However, our results indicate that the benefits of pretraining depend critically on the distributional similarity between the pretraining and finetuning datasets, as detailed in the previous subsections. For instance, models pretrained on \texttt{FaultSeg3D} or \texttt{CRACKS} and fine-tuned on \texttt{Thebe} underperform compared to models trained from scratch or initialized from ImageNet weights, as we show in Figure~\ref{fig:thebe_models}. 
However, in seismic applications, the benefits of pretraining are critically sensitive to the distributional similarity between the source and target domains—a phenomenon that reflects the strong coupling between geologic variability and model transferability. As shown in our experiments, models pretrained on \texttt{FaultSeg3D} or \texttt{CRACKS} and fine-tuned on \texttt{Thebe} underperform compared to models trained from scratch or initialized from ImageNet weights (Fig.~\ref{fig:thebe_models}). In such distributionally mismatched cases, the standard pretaining-finetuning strategy may hinder rather than help performance.

A factor that also plays a role in modulating generalization is model capacity. Fig.~\ref{fig:thebe_models} shows that large models like \textsf{segformer} benefit from pretraining on \texttt{CRACKS}, while smaller models (e.g., \textsf{hed}, \textsf{deeplab-m}) generalize better when trained from scratch. The observation suggests that both data alignment and model capacity affect transfer effectiveness. %This interaction between model size and pretraining benefits suggests that a higher model capacity or parameter count is not always better, but rather amplifies both the advantages and disadvantages of transfer. Large-capacity architectures can better exploit well-aligned pretrained features but are also more prone to overfitting to domain-specific artifacts when source and target distributions diverge.
Larger architectures, particularly transformer-based models like \textsf{segformer}, possess greater representational capacity and self-attention mechanisms that capture long-range spatial dependencies, an important property for preserving fault continuity and contextual consistency across sections. These characteristics can potentially explain their stronger cross-domain generalization when pretrained features are well aligned. However, the same capacity also makes them more sensitive to distributional mismatches, leading to potential overfitting to domain-specific amplitude statistics or noise patterns when source and target differ substantially. Smaller-capacity models, by contrast, may act as an implicit regularizer, limiting over-specialization and enabling better zero-shot transfer in mismatched scenarios, as can be observed in Fig.~\ref{fig:thebe_models}.

\begin{figure}
    \centering
    \includegraphics[width=1\linewidth]{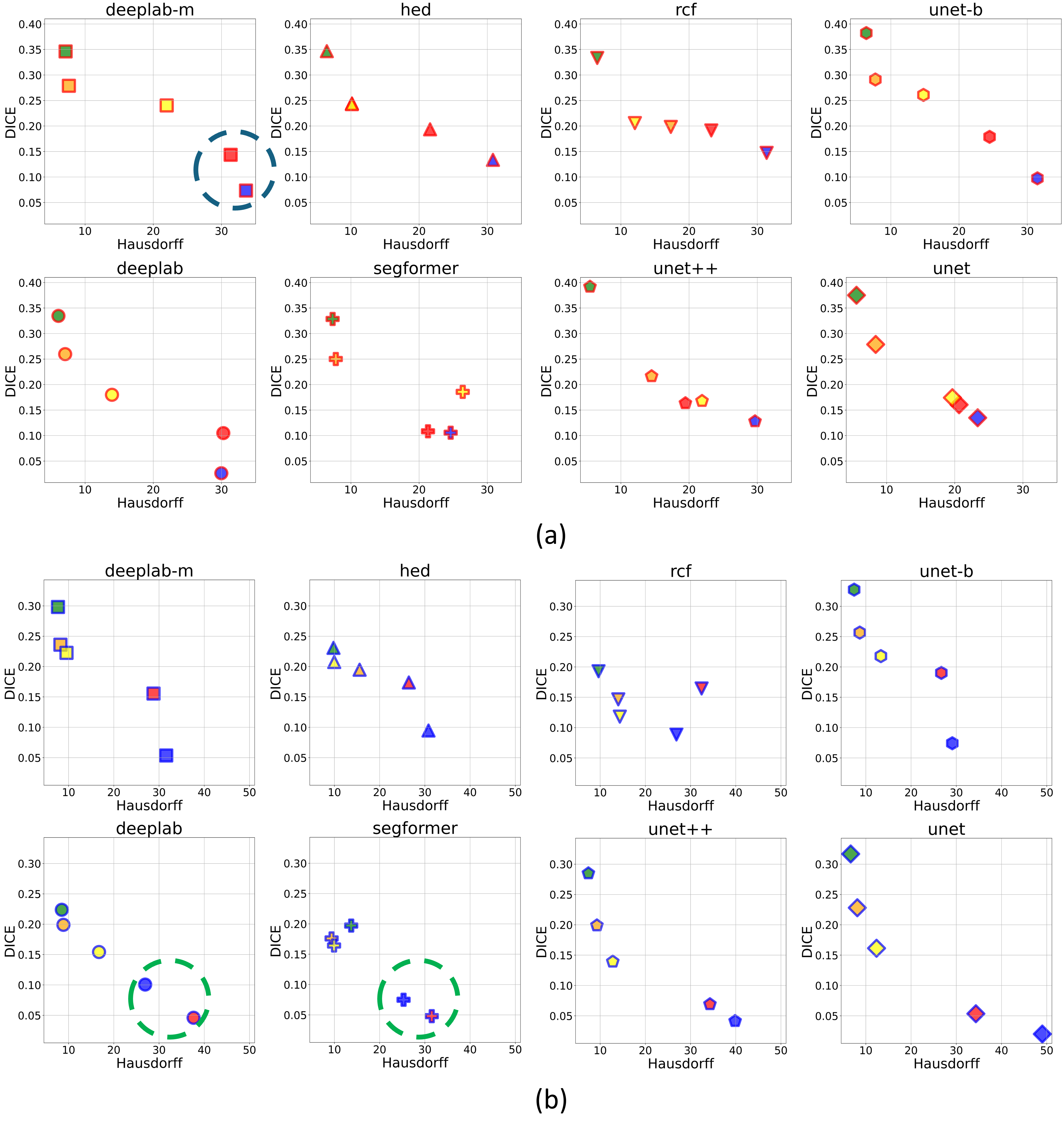}
    \caption{Individual models tested on  \texttt{CRACKS}. (a) Models pre-trained on the \texttt{FaultSeg3D} data, and fine-tuned on different data. (b) Models pre-trained on the Thebe data, and fine-tuned on different data.}
    \label{fig:models_synth_thebe-train_all-tune_CRACKS-test}
\end{figure}

Furthermore, models respond differently to fine-tuning. When pretraining on \texttt{FaultSeg3D} and fine-tuning on \texttt{Thebe}, models degrade in \texttt{CRACKS} performance (Fig.~\ref{fig:models_synth_thebe-train_all-tune_CRACKS-test}, blue circle) shows that \texttt{Thebe} induces domain shifts that are hard to unlearn. Conversely, even though the finetuning dataset is distributionally closer to the target, \texttt{Thebe}-pretrained \textsf{segformer} and \textsf{deeplab} also degrade on \texttt{CRACKS} after finetuning on \texttt{FaultSeg3D} as shown in Fig. ~\ref{fig:models_synth_thebe-train_all-tune_CRACKS-test} (green circles). The asymmetric behavior highlights the difficulty of finding universally robust pretraining strategies.

Given these limitations, we explored whether domain adaptation methods could help in such settings. We applied Domain-Adversarial Neural Networks (DANN) \cite{ganin2016domain} and Fourier Domain Adaptation (FDA) \cite{yang2020fda, trinidad2023seismic} to one of our smaller models, \textsf{unet-b}, under the same transfer setups used in our EWC experiments (\texttt{FaultSeg3D} to \texttt{CRACKS} and \texttt{FaultSeg3D} to \texttt{Thebe}). The results in Table~\ref{tab:da} reveal an intriguing pattern: adaptation improved performance for the more distributionally distant \texttt{FaultSeg3D}$\rightarrow$\texttt{Thebe} transfer, surpassing fine-tuning, yet underperformed for the closer \texttt{FaultSeg3D}$\rightarrow$\texttt{CRACKS} case. This reflects a known transfer learning phenomenon, often termed \emph{negative transfer} or \emph{over-adaptation}\cite{zhang2022survey}, where adapting already well-aligned domains can distort useful features. In contrast, for large domain shifts, adaptation effectively bridges representational gaps. Taken together, these findings suggest a simple yet practical guideline: when source and target domains are similar, fine-tuning is often sufficient, whereas substantial domain divergence may justify the additional complexity of adaptation methods.

% \usepackage{multirow}
\begin{table}[]
\centering
\caption{Results on domain adaptation case study}
\label{tab:da}
\resizebox{\linewidth}{!}{%
\begin{tabular}{ccc|cc|cc}

\multirow{2}{*}{Training configuration} & \multicolumn{2}{c|}{FDA} & \multicolumn{2}{c}{DANN} & \multicolumn{2}{c}{Finetuning} \\ \cline{2-7} 
                                        & Dice      & Hausdorff    & Dice      & Hausdorff & Dice      & Hausdorff    \\ \hline
\texttt{FaultSeg3D $\rightarrow$ Cracks}         & 0.139     & 48.221       & 0.156     & 41.058  & 0.375 & 5.577      \\ \hline
\texttt{FaultSeg3D $\rightarrow$ Thebe}           & 0.276     & 20.277       & 0.319     & 15.872  & 0.158  & 51.649     \\ \hline
\end{tabular}
}
\end{table}

%\subsubsection{Some characteristics in the data can make pretraining not a good choice} 
%\subsubsection{Sometimes it is better to start off with randomly initialized weights than pretraining on a different data} 

%\begin{figure}
%    \centering
%    \includegraphics[width=\linewidth]{Figures/pretrainedvsfintuned.png}
%    \caption{Different label sources across slices}
%    \label{fig:prevsfine}
%\end{figure}
% Finally, it is generally established in the literature that pretraining on a large dataset can boost the performance of a model even in self-supervised settings \cite{rs16050922}. However, our results indicate that this is conditioned to how much the pretraining and finetuning datasets are correlated and share the same distribution (section 4.1.1). In cases where these datasets are not correlated, training from scratch or using Imagenet weights is a better option, as we show in Figure~\ref{fig:thebe}. In this example, we compare different models pretrained on different datasets and fine-tuned on Thebe, against these same models trained from scratch on Thebe or from Imagenet. Since Thebe exhibits a domain shift from FaultSeg 3D and \texttt{CRACKS}, pretraining on these datasets does not yield performance gains. In fact, either training the model from scratch or using ImageNet weights to initialize the model is better than pertaining on Faultseg 3D or \texttt{CRACKS}. \\

\subsection{Training Dynamics and Inferential Behavior}
In this section we analyze the impact of different training dynamics and model architectures on performance and inferential behavior.

\subsubsection{Window Size and Loss Function}
\label{sec:window_size}
\begin{figure}
    \centering
    \includegraphics[width=\linewidth]{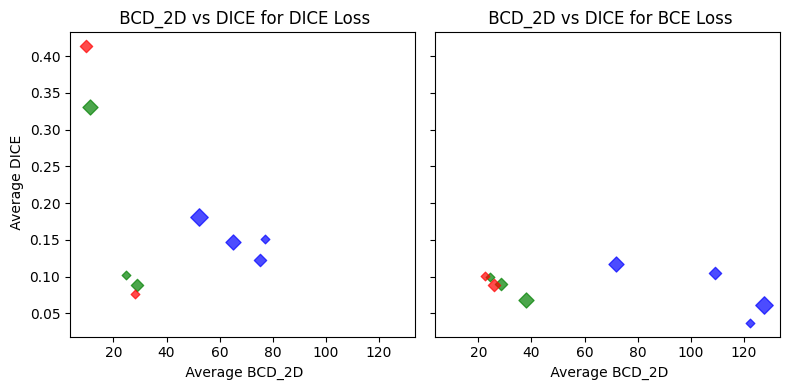}
    \caption{Behavior of DICE and BCE losses under different window sizes}
    \label{fig:losspatch}
\end{figure}

We evaluate the \textsf{unet} model with a ResNet50 backbone using Dice and Binary Cross-Entropy (BCE) losses, as well as across different window sizes: 96, 128, 256 and 512, or up to the size allowed by the original sections in each dataset. The results for these experiments are shown in Fig.~\ref{fig:losspatch}, where bigger markers correspond to bigger patch sizes. 

We can see that two trends emerge in these experiments. First, when using Dice as a loss, models benefit consistently from larger window sizes, with performance generally improving as the spatial context grows. This is likely because larger patches provide the network with a more complete view of fault structures, enabling better continuity modeling across sections. Second, when using BCE loss, this trend does not hold: performance remains flat or slightly declines with larger windows. This discrepancy stems from the class imbalance inherent to fault delineation \cite{sudre2017generalised}. BCE tends to work best when foreground and background classes are more balanced, whereas Dice loss is explicitly designed to handle imbalance. Using smaller patches effectively “zooms in” on the sparse fault regions, increasing the proportion of fault pixels and improving BCE performance.

From a practical perspective, these results suggest that in real-world seismic interpretation, the optimal windowing standard depends on the chosen loss function and the model’s need for contextual information. For losses that handle imbalance well (e.g., Dice), larger tiles are preferable because they capture longer fault segments and contextual cues that allow for improve spatial continuity. For imbalance-sensitive losses (e.g., BCE), smaller tiles may sometimes be beneficial, though at the cost of reduced global context. Given that Dice consistently benefits from larger windows, and that most modern segmentation pipelines for faults employ class-imbalance aware losses, our choice of using Dice with the largest possible window sizes in our benchmark experiments is both empirically supported and aligned with best practices in seismic fault delineation.

% We can see that when using Dice as a loss, models benefit from having access to larger window sizes, while the same does not hold when using the BCE loss. This discrepancy is due to the class imbalance nature of the fault delineation task: BCE tends to work best when foreground and background are balanced but fails otherwise, and Dice is generally designed for class-imbalanced scenarios \cite{sudre2017generalised}. In this context, using smaller patches is equivalent to 'zooming into' the small fault regions and therefore reducing the class imbalance, thus improving BCE performance. Overall, this further validates our choice of using the Dice loss on our benchmarking experiments.

\subsubsection{Model Nuances in Fault Delineation}
\begin{figure}
    \centering
    \includegraphics[width=1\linewidth]{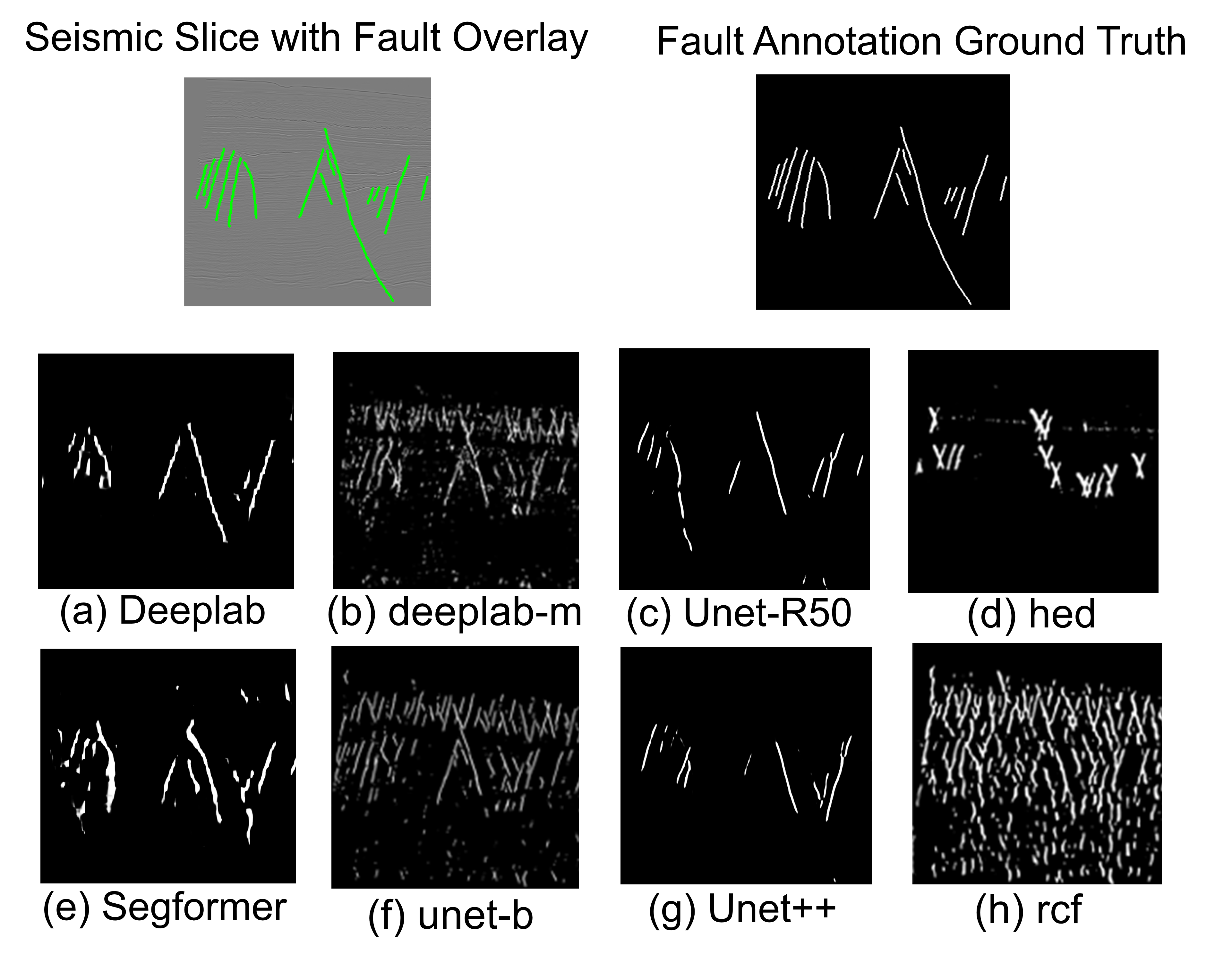}
    \caption{Predictions of the models pretrained on \texttt{FaultSeg3D} tested \texttt{CRACKS} }
    \label{fig:objective_results_synth_CRACKS}
\end{figure}

We also qualitatively observe that each of the evaluated models presents different nuances in the structure of their fault predictions, irrespective of the pretraining or finetuning strategy used, many of which can be observed in Fig.~\ref{fig:objective_results_synth_CRACKS}. For example, \textsf{deeplab} tends to produce irregular, stair-like faults, while \textsf{segformer} produces thicker, blob-like faults. \textsf{unet} architectures in general tend to produce thinner faults, with \textsf{unet++} generating more fragmented ones. These architectural signatures are consistent across training setups and highlight the influence model design choices inherently have in shaping the morphology of predicted faults, which is an important consideration when selecting models for downstream tasks or when interpreting evaluation results beyond numerical metrics.

\subsection{Metric Robustness and Observability}
Due to the high correlation among adjacent sections in a seismic volume, deep learning models tend to generate consistent patterns that vary minimally between neighboring sections. Evaluation metrics respond differently to these subtle variations; some metrics heavily penalize these deviations, while others are more tolerant. The effects of structural variations in fault predictions on the evaluation metrics are investigated.

\subsubsection{Sensitivity to Visual Structure}
\label{sec:sensitivity}
\begin{figure}[!t]
\centering

% --- Top Row ---
\begin{subfigure}[b]{0.47\linewidth}
    \centering
    \includegraphics[width=\linewidth]{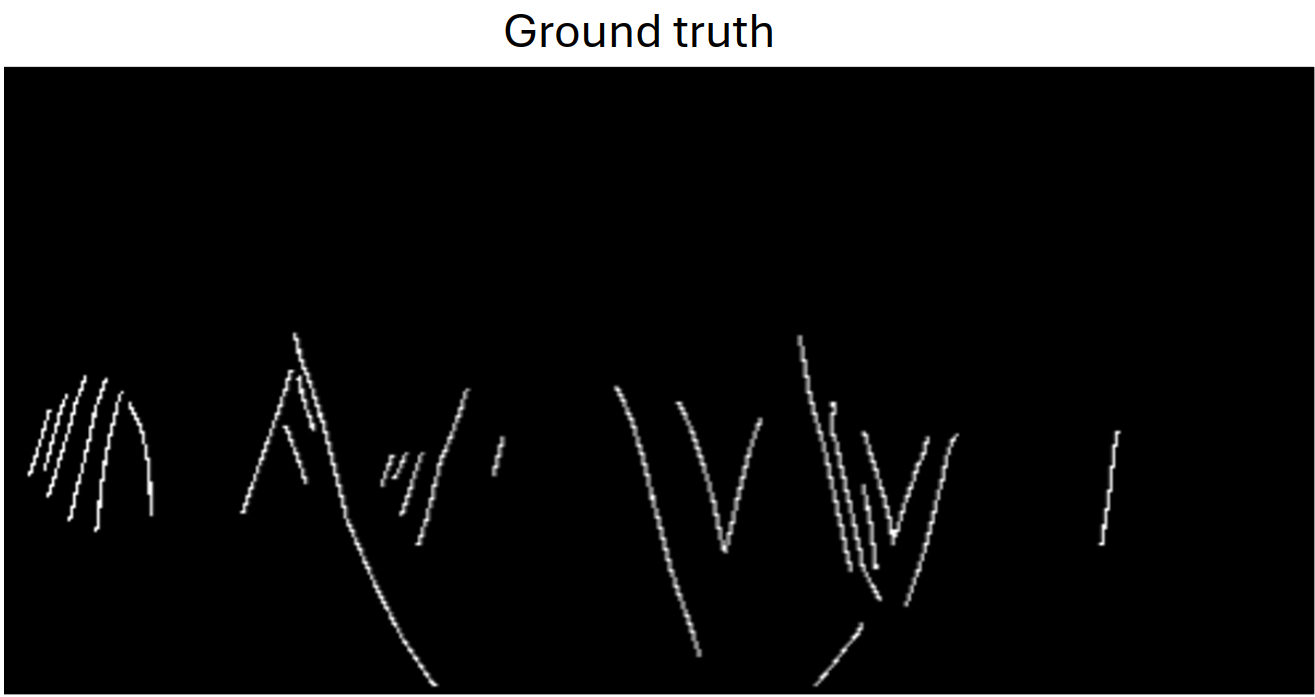}
    \caption{} % Automatically becomes (a)
    \label{fig:dicevsbcd_a}
\end{subfigure}
\hfill % Pushes the next subfigure to the right
\begin{subfigure}[b]{0.47\linewidth}
    \centering
    \includegraphics[width=\linewidth]{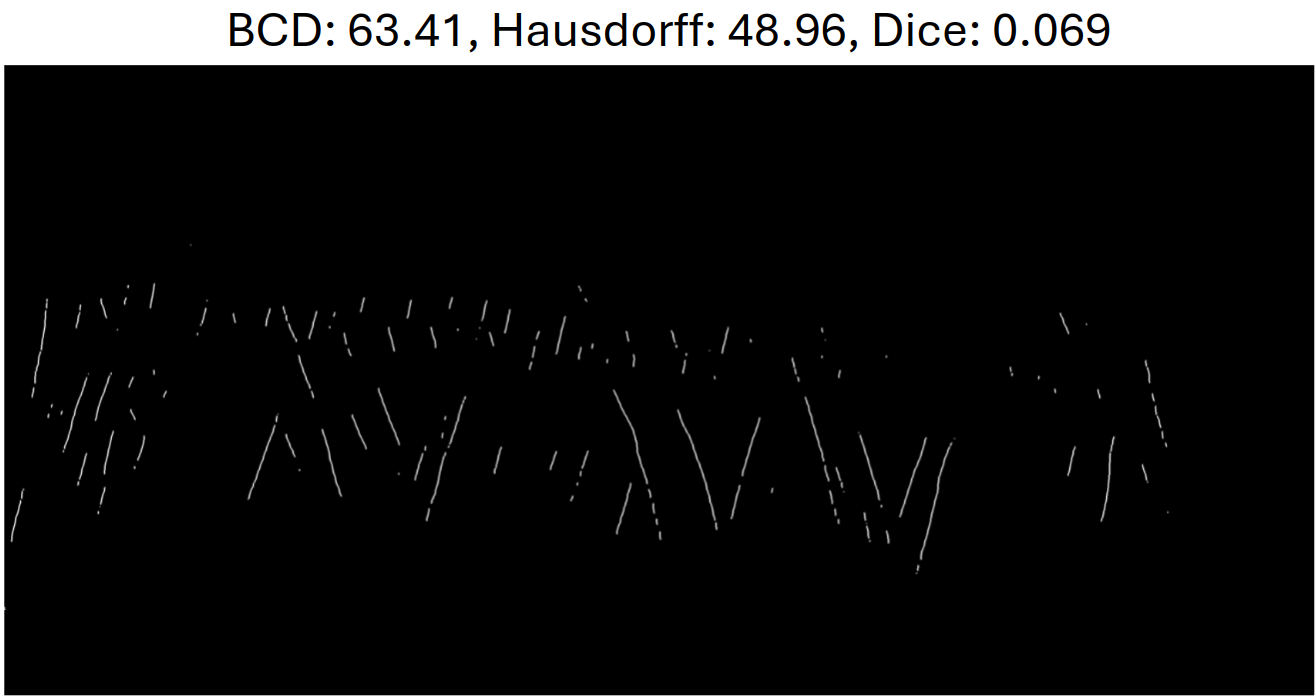}
    \caption{} % Automatically becomes (b)
    \label{fig:dicevsbcd_b}
\end{subfigure}

% --- This blank line creates the new row ---

% --- Bottom Row ---
\begin{subfigure}[b]{0.47\linewidth}
    \centering
    \includegraphics[width=\linewidth]{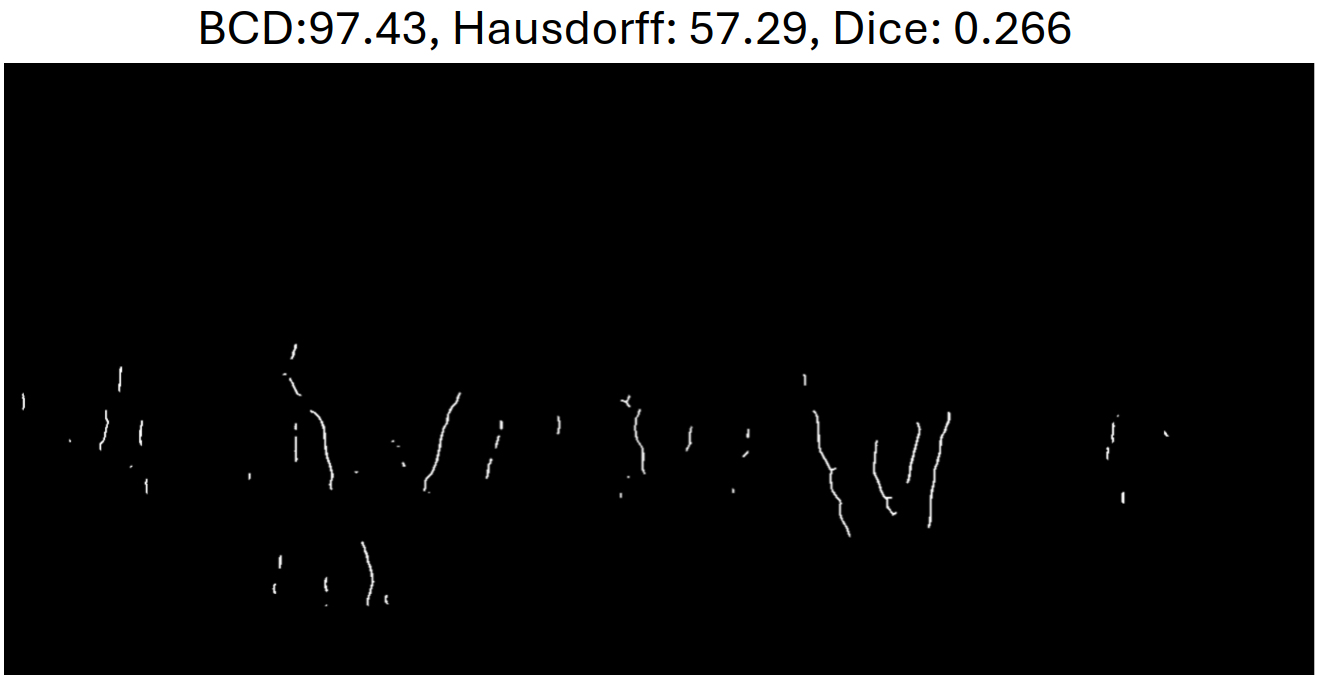}
    \caption{} % Automatically becomes (c)
    \label{fig:dicevsbcd_c}
\end{subfigure}

\caption{Example of the structure tolerance in distance-based metrics. (a) shows the ground truth fault annotations. (b) and (c) show predictions from two different models. While (b) appears structurally closer to the ground truth, it receives a significantly worse BCD and Hausdorff but a better Dice score.}
\label{fig:dicevsbcd}
\end{figure}

Distance-based metrics are generally more tolerant to the structure of the predicted fault. This behavior is illustrated in Fig. ~\ref{fig:dicevsbcd}. Where Fig. \ref{fig:dicevsbcd_a} represents the ground truth, while \ref{fig:dicevsbcd_b} and \ref{fig:dicevsbcd_c} show the predictions of two different models. Although the prediction in Fig. \ref{fig:dicevsbcd_b} appears structurally closer to the ground truth, it receives significantly worse BCD and Hausdorff scores compared to the prediction in Fig. \ref{fig:dicevsbcd_c}. Since distance-based metrics do not heavily weigh continuity, the inclusion of a few extra pixels around the fault can improve the BCD score even if those pixels lack proper structural alignment. Notably, these patterns are not anomalies, different models often generate such consistent outputs. Consequently, numerical metrics can be misleading and may display discrepancies between one another.
%
%Consequently,  as DL models generate these patterns consistently, numerical metrics can sometimes be misleading and may display discrepancies between one another.

% \begin{figure}
%     \centering
%     % \begin{subfigure}[]{0.45\linewidth}
%     %     \centering
%     %     \includegraphics[width=\linewidth]{Figures/fig18_a.png}
%     %     \caption{}%Ground Truth}
%     %     \label{fig:dicevsbcd_a}
%     % \end{subfigure}

%     % %\vspace{1em}

%     % \begin{subfigure}[]{0.45\linewidth}
%     %     \centering
%     %     \includegraphics[width=\linewidth]{Figures/fig18_b.png}
%     %     \caption{}%BCD: 63.41, Hausdorff: 48.96, Dice: 0.069}
%     %     \label{fig:dicevsbcd_b}
%     % \end{subfigure}

%     % %\vspace{1em}

%     % \begin{subfigure}[]{0.45\linewidth}
%     %     \centering
%     %     \includegraphics[width=\linewidth]{Figures/fig18_c.png}
%     %     \caption{}%BCD:97.43, Hausdorff: 57.29, Dice: 0.266}
%     %     \label{fig:dicevsbcd_c}
%     % \end{subfigure}

%     \caption{Example of the structure tolerance in distance-based metrics. (a) shows the ground truth fault annotations. (b) and (c) show predictions from two different models. While (b) appears structurally closer to the ground truth, it receives a significantly worse BCD and Hausdorff but a better Dice score.}
%     \label{fig:dicevsbcd}
% \end{figure}

\subsubsection{Fault Sparsity}
\begin{figure}[ht]
    \centering
    \begin{subfigure}{0.49\linewidth}
     \centering
        \includegraphics[width=1\linewidth]{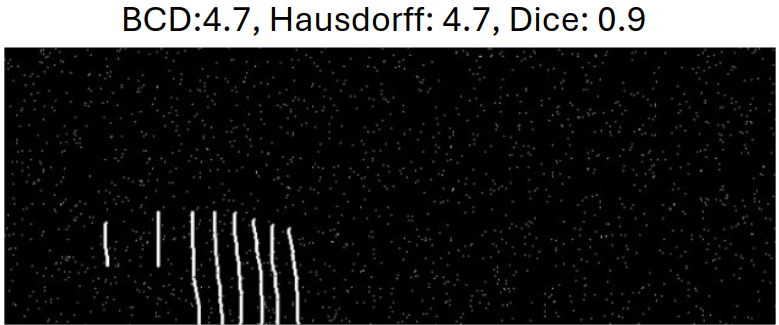}
        \caption{}
        \label{fig:fig17_a}
    \end{subfigure}
 \hfill
    \begin{subfigure}{0.49\linewidth}
     \centering
        \includegraphics[width=1\linewidth]{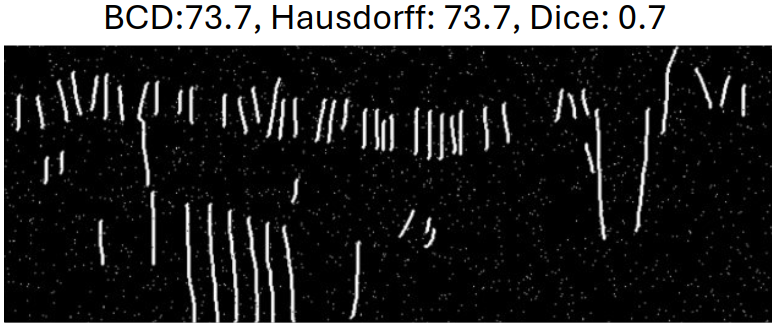}
        \caption{}
        \label{fig:fig17_b}
    \end{subfigure}
    \caption{Sensitivity of distance-based metrics to fault density: Sections with many (a) vs. few faults (b) are evaluated under identical noise. Despite equal noise levels, (b) is penalized more by distance-based metrics, while Dice remains stable.}
    \label{fig:noisy}
\end{figure}

Another issue with distance-based metrics is that they are designed to measure the quality of a single continuous object in the image, whereas faults can consist of multiple sparse objects. This makes distance-based metrics very sensitive to both the number of faults present in the ground truth and their sparsity. We showcase this issue in Fig. ~\ref{fig:noisy}, where we consider two cases: one with a few faults and another with many faults. Noise is added to both images, and each is compared against its original version. The case with fewer faults contains many outliers compared to the case with many faults, since the added noise often lies far from any existing fault. As discussed in Section \ref{sec:rw_metrics}, because BCD is bidirectional, it accounts for each added noise pixel by searching for its closest existing fault. Therefore, having fewer faults results in a much worse BCD score. DICE, while also penalizing noise, is more stable across different sparsity levels.

\subsubsection{Contradictory Scores and Human Judgement}
\begin{figure}
    \centering
    \begin{subfigure}[b]{\linewidth}
        \centering
        \includegraphics[width=0.8\linewidth]{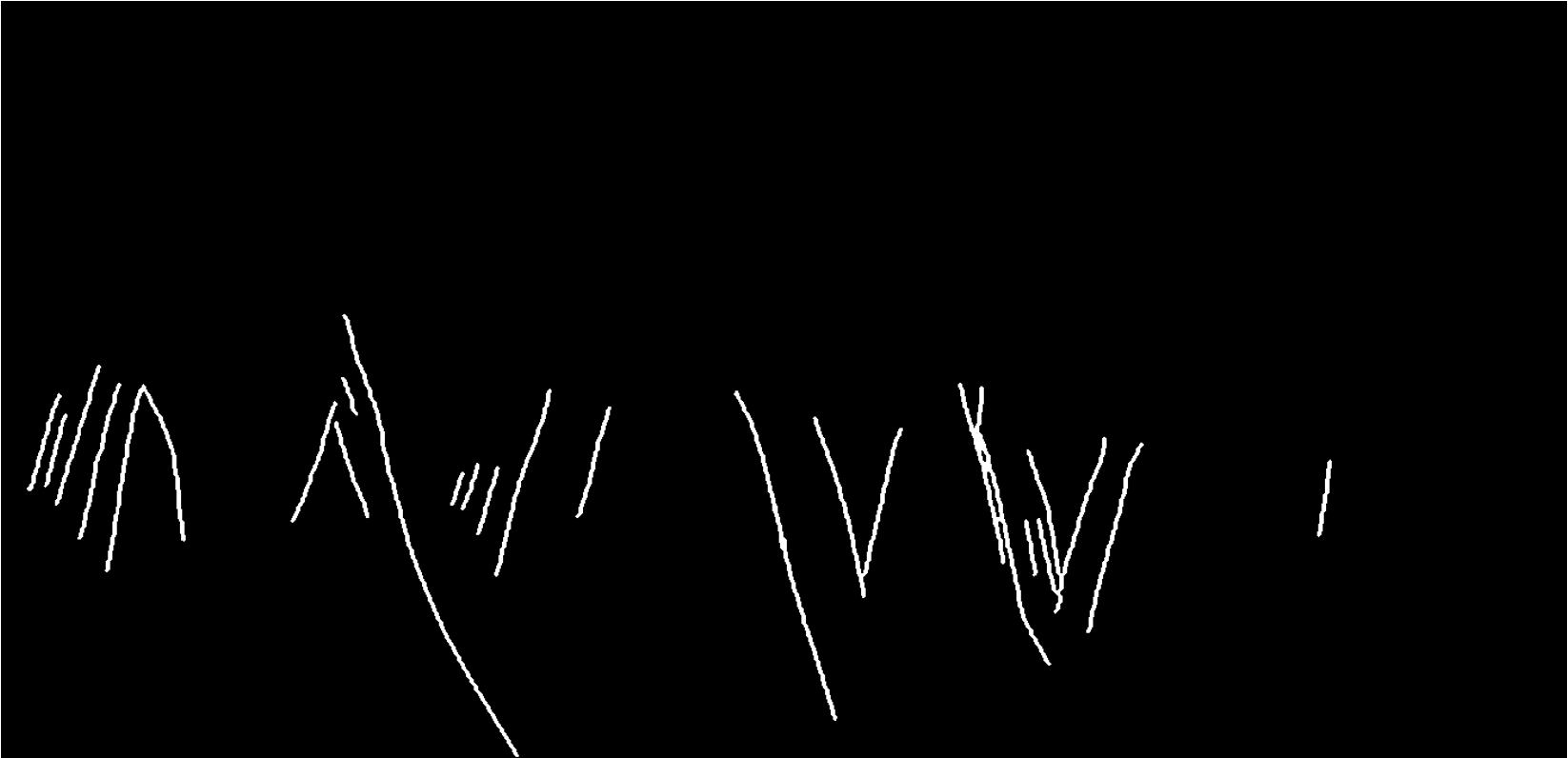}
        \caption{Ground Truth}
        \label{fig:bcdvsdice_a}
    \end{subfigure}

    \vspace{1em}

    \begin{subfigure}[b]{\linewidth}
        \centering
        \includegraphics[width=0.8\linewidth]{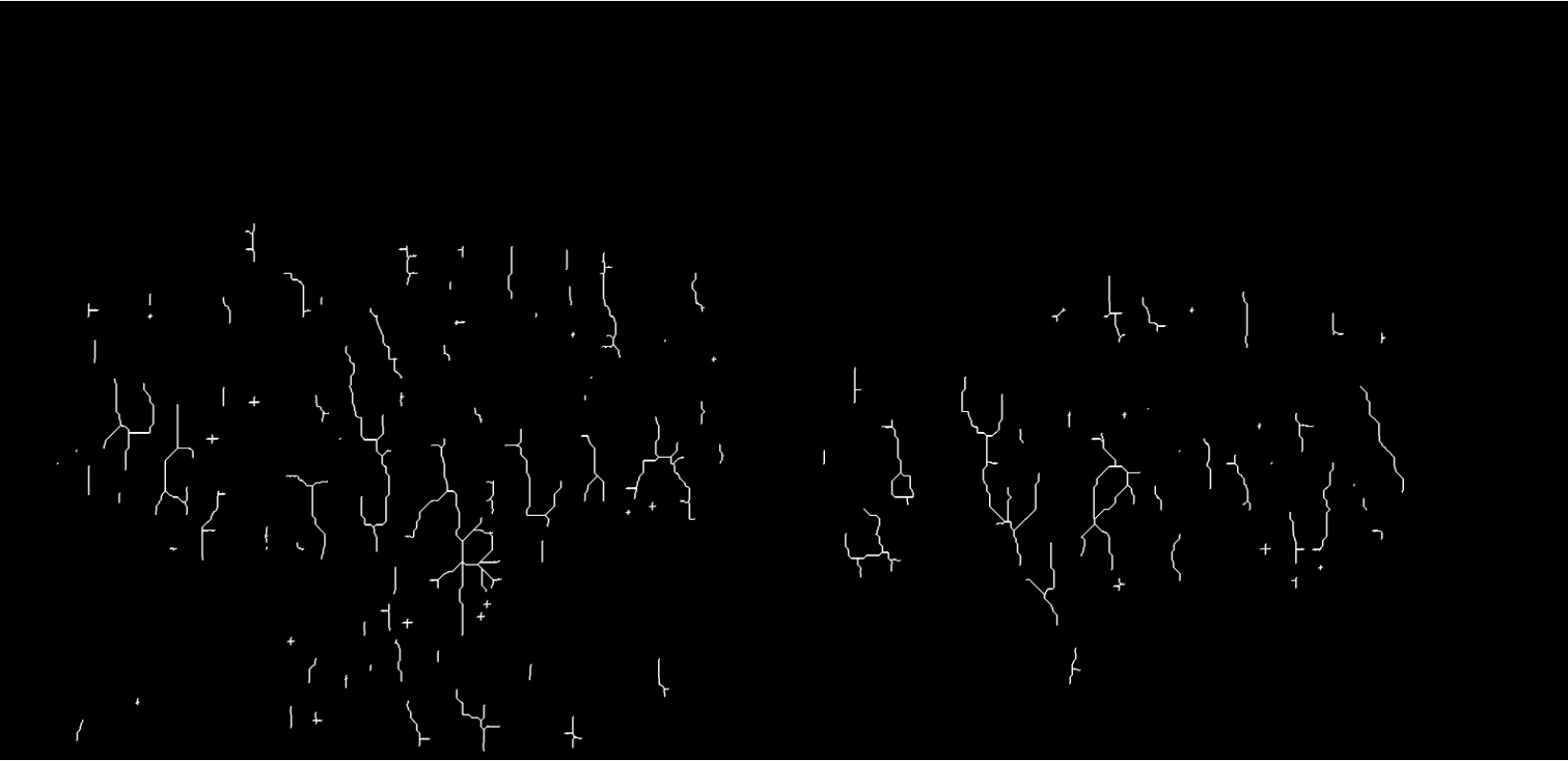}
        \caption{BCD:115.43, Hausdorff: 72.5, Dice: 0.1342}
        \label{fig:bcdvsdice_b}
    \end{subfigure}

    \vspace{1em}

    \begin{subfigure}[b]{\linewidth}
        \centering
        \includegraphics[width=0.8\linewidth]{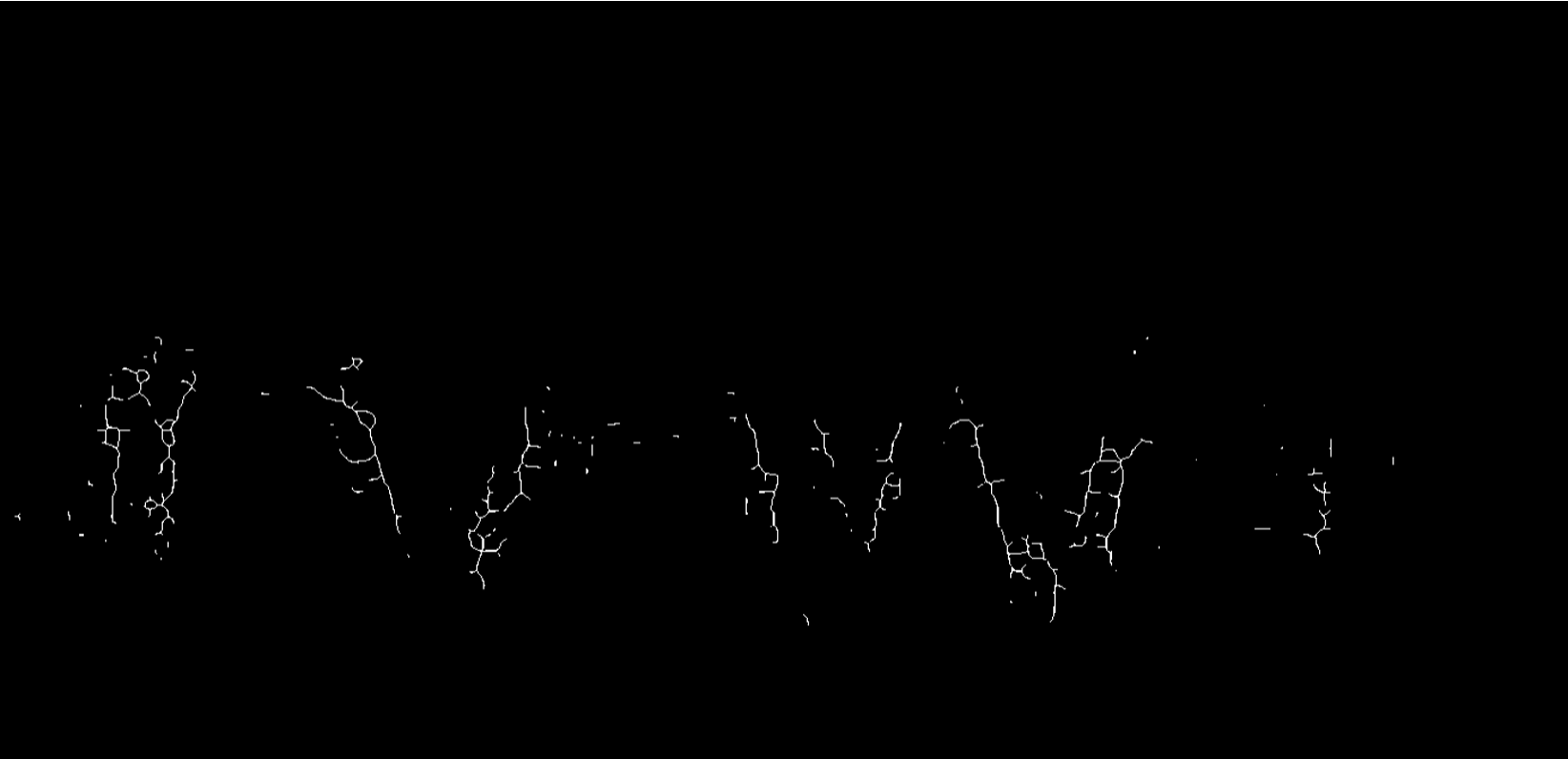}
        \caption{BCD:40.782, Hausdorff: 29.0, Dice: 0.1325}
        \label{fig:bcdvsdice_c}
    \end{subfigure}

    \caption{(a) Ground truth fault annotations. (b) and (c) display predictions from two different models. Although the prediction in (c) appears visually more aligned with the ground truth than (b), it receives a slightly worse Dice score but substantially better BCD and Hausdorff distance}
    \label{fig:bcdvsdice}
\end{figure}

Although pixel-based metrics are more stable, they remain sensitive to slight pixel shifts. In Fig. ~\ref{fig:bcdvsdice}, we show examples corresponding to two different models. The prediction in Fig. \ref{fig:bcdvsdice_c} looks visually closer to the ground truth, but receives a worse Dice score (0.1325 vs. 0.13421) and significantly better BCD and Hausdorff (40.782 vs. 115.436 and 29.0 vs 72.5). Since the faults in both predictions are poorly structured and spatially misaligned, the Dice coefficient (being overlap-based) penalizes them similarly. In contrast, distance-based metrics such as BCD and Hausdorff distances are more sensitive to the spatial coherence of the predicted faults, hence, they are more tolerant to the structure of the faults. These contradictions imply that some metrics may conflict with visual intuition or downstream utility, and point to the need for context-aware metric selection frameworks.

% ...

\section{Analysis Using Fault Characteristic Metrics}
\label{sec:fault_metric_results}

\begin{table}[h!]
\centering
\caption{Summary statistics of the three datasets: CRACKS (A), Thebe (B), and FS (C). Values are reported as mean $\pm$ standard deviation.}
\label{tab:dataset_stats}
\resizebox{\columnwidth}{!}{%
\begin{tabular}{lccccc}
\hline
\textbf{Dataset} & \textbf{Length} & \textbf{Curvature} & \textbf{Sinuosity} & \textbf{Segments} & \textbf{Stepover Density} \\
\hline
CRACKS & 6577.75 $\pm$ 1090.82 & 0.0177 $\pm$ 0.0038 & 95.35 $\pm$ 54.30 & 47.22 $\pm$ 7.61 & 0.978 $\pm$ 0.0046 \\
Thebe & 38457.25 $\pm$ 7676.67 & 0.0261 $\pm$ 0.0058 & 13.46 $\pm$ 12.32 & 10.45 $\pm$ 1.85 & 0.902 $\pm$ 0.0169 \\
FS & 16543.79 $\pm$ 3452.73 & 0.0369 $\pm$ 0.0081 & 5.70 $\pm$ 6.32 & 3.18 $\pm$ 1.10 & 0.623 $\pm$ 0.207 \\
\hline
\end{tabular}%
}
\end{table}

\begin{table*}[h!]
\centering
\caption{Evaluation metrics when models are pretrained on Thebe and tested on Thebe, with or without finetuning on other datasets.}
\label{tab:training_stats}
\resizebox{\textwidth}{!}{%
\begin{tabular}{lcccccccccccc}
\hline
\textbf{Setup} & \textbf{Strike Sim.} & \textbf{Curvature} & \textbf{Curv. RMSE} & \textbf{Curv. Corr} & \textbf{Sinuosity $\Delta$} & \textbf{Sinuosity Ratio} & \textbf{Length $\Delta$} & \textbf{Length Ratio} & \textbf{Segment $\Delta$} & \textbf{Segment Ratio} & \textbf{Stepover $\Delta$} & \textbf{Stepover Ratio} \\
\hline
Thebe Only      & 0.907 $\pm$ 0.081 & -0.001 $\pm$ 0.006 & 0.128 $\pm$ 0.014 & 0.019 $\pm$ 0.054 & 0.246 $\pm$ 0.140 & 0.829 $\pm$ 0.068 & 1715.41 $\pm$ 1044 & 1.649 $\pm$ 0.543 & 51.47 $\pm$ 10.04 & 3.292 $\pm$ 0.455 & 1.779 $\pm$ 6.669 & 0.993 $\pm$ 0.061 \\
Thebe→CRACKS   & 0.756 $\pm$ 0.107 & 0.012 $\pm$ 0.010 & 0.174 $\pm$ 0.015 & -0.010 $\pm$ 0.044 & 0.080 $\pm$ 0.152 & 0.950 $\pm$ 0.090 & -1593.73 $\pm$ 925.87 & 0.560 $\pm$ 0.187 & 67.08 $\pm$ 9.09 & 4.001 $\pm$ 0.490 & 4.770 $\pm$ 0.001 & 0.989 $\pm$ 0.104 \\
Thebe→FS       & 0.658 $\pm$ 0.053 & 0.007 $\pm$ 0.010 & 0.166 $\pm$ 0.017 & -0.003 $\pm$ 0.049 & -0.055 $\pm$ 0.183 & 0.968 $\pm$ 0.114 & 1141.05 $\pm$ 963.78 & 0.705 $\pm$ 0.241 & 90.39 $\pm$ 14.91 & 5.046 $\pm$ 0.788 & 0.0004 $\pm$ 0.0001 & 1.004 $\pm$ 0.024 \\
\hline
\end{tabular}%
}
\end{table*}

When evaluating transfer setups where all models were pretrained on Thebe and subsequently finetuned on different datasets before being tested back on Thebe, we observe a clear correspondence between the statistical characteristics of the finetuning dataset (Table \ref{tab:dataset_stats}) and the error patterns in the predictions (Table \ref{tab:training_stats}). Training and testing solely on Thebe produces the strongest alignment with the ground truth, as indicated by the highest strike similarity (0.907), reflecting that the model effectively captures the long and continuous fault structures characteristic of Thebe. However, this setup also leads to an overestimation of fault length (length ratio of 1.65) and moderate over-segmentation (segment ratio of 3.29), suggesting that the model tends to exaggerate continuity while artificially fragmenting longer structures.

In contrast, when the Thebe-pretrained model is finetuned on CRACKS and evaluated on Thebe, performance degrades in ways consistent with the CRACKS dataset statistics, which emphasize short, jagged, highly segmented faults with frequent crossovers. Strike similarity drops substantially to 0.756, indicating that the orientation of predictions becomes less aligned with Thebe’s long faults. Length is strongly underestimated (length ratio of 0.56), directly mirroring the shorter average fault lengths in CRACKS. Segmentation increases (segment ratio of 4.00), and the stepover delta rises to 4.77, reflecting the fragmentation and crossover-heavy nature of CRACKS. Moreover, spacing becomes strongly negative, showing that the predictions adopt the denser fault placement bias of CRACKS rather than the sparser spacing of Thebe.

A different type of degradation is observed when finetuning on FaultSeg3D, which is dominated by medium-length, smooth faults with very low segmentation and reduced connectivity. Here, strike similarity falls further to 0.658, the lowest among all setups, demonstrating poor alignment with Thebe’s long continuous faults. Length is again underestimated (ratio of 0.70), in line with FaultSeg3D’s shorter structures. Segmentation is highly exaggerated (segment ratio of 5.05), but stepovers almost vanish (stepover delta close to 0), consistent with FaultSeg3D's low stepover density. The negative spacing indicates that the model no longer respects Thebe’s distribution of fault separations. Thus, while finetuning on FaultSeg3D removes the crossover effects seen with CRACKS, it introduces extreme fragmentation of Thebe’s continuous faults and produces disconnected structures.

In summary, the in-domain Thebe-only model best preserves structural alignment but exaggerates continuity and segmentation, whereas finetuning on CRACKS injects a bias toward jagged, fragmented, and crossover-rich structures, and finetuning on FaultSeg3D enforces smooth, disconnected, and overly fragmented representations. Both transfer setups degrade performance relative to the in-domain baseline, but in ways that directly reflect the statistical properties of the finetuning datasets. This demonstrates that the dataset-specific fault geometry strongly governs the inductive bias of models, even when pretrained on the same source.

\noindent Motivated by these observations, we distill them into a self-contained, general ``model pick list'' to guide out-of-the-box use (Table~\ref{tab:model_selection_access_rot}): first, profile the target dataset by labeling its fault--network statistics as High/Medium/Low (H/M/L) relative to Table~\ref{tab:dataset_stats}; then select the row whose triggers match that profile and adopt the corresponding pretrain\,{$\rightarrow$}\,finetune setup; finally, confirm the choice against cross-setup behavior in Table~\ref{tab:training_stats} and apply the suggested lightweight post-prediction fixes. This table is intentionally model-agnostic with respect to architecture and emphasizes geometry-aware selection aligning to fault length, segmentation, sinuosity, curvature, stepovers, and spacing so that practitioners can choose the most appropriate setup for their basin characteristics without additional tuning.

\section{Conclusion}
\label{sec:conclusion}

In this work, we present the first large-scale benchmarking study of seismic fault delineation models under domain shift, spanning more than 200 training configurations across three heterogeneous datasets. Our results show that fine-tuning is generally effective when source and target domains are closely aligned, but becomes brittle under stronger shifts, often leading to catastrophic forgetting. Model capacity also modulates transferability: larger architectures such as Segformer tend to adapt more effectively, while smaller models are more sensitive to mismatch. Domain adaptation methods such as FDA and DANN proved beneficial in highly divergent transfers but sometimes degraded performance in more similar settings, highlighting the risk of negative transfer.

Beyond conventional metrics such as Dice or Hausdorff distance, our geometric and topological analysis demonstrated that models also absorb the structural biases of the datasets they are finetuned on: CRACKS-trained models tended to reproduce short, jagged, crossover-rich structures, while FaultSeg3D-trained models favored smoother but disconnected faults. Even when pretrained on the same source, prediction styles were shaped by the statistical properties of the finetuning dataset. These findings underscore that evaluation should not rely only on pixel-level accuracy but also account for structural plausibility and geological realism.

At the same time, important open questions remain. Future research should focus on designing adaptation methods that remain effective across both mild and severe domain shifts, incorporating geological priors into model architectures to mitigate dataset-specific biases, and developing evaluation protocols that balance quantitative rigor with geological interpretability. Addressing these challenges will be essential to building seismic DL pipelines that are not only accurate but also dependable in real-world interpretation workflows.

 %\appendices
 \onecolumn

%\section{Observations from Benchmark}
%\label{sec:observations}
% \textcolor{orange}{slices example when fine-tuning different , and then fix the data and models to give a }
% \textcolor{orange}{List of observations, e.g., segformer produce discontinous faults while unet produces not necessarily long paragraphs, they can be small sentences of concrete statements.}
% \subsection{Data-based observations} \textcolor{red}{Jorge}
%\textcolor{orange}{talk about this from two perspectives interpreters/ ML perspectives}
% \subsubsection{existence of subspace/metric space along which different data sources lie} 

{\scriptsize
\begin{table*}[h]
\centering
\caption{Key observations}
\label{tab:key_observations}
\resizebox{\textwidth}{!}{%
\begin{tabular}{l p{4.5in}}
\toprule
\textbf{Topic} & \textbf{Key Findings} \\
\midrule
\multirow{3}{1in}{Distributional shift among seismic data} &
The intensity standard deviation of the synthetic \texttt{FaultSeg3D} data is similar to that of \texttt{CRACKS}, while both are very different from \texttt{Thebe}'s. This can be attributed to the discrepancy of seismic features across datasets. \\
\cmidrule(lr){1-2}

\multirow{2}{1.4in}{Relationship between \texttt{CRACKS} and \texttt{FaultSeg3D}} &
Both \texttt{CRACKS} and \texttt{FaultSeg3D} data benefit from pretraining on the other, outperforming models trained from scratch in either. \\
\cmidrule(lr){1-2}

\multirow{2}{1in}{\texttt{Thebe} vs. others} &
\texttt{Thebe} does not benefit significantly from pretraining on other data; training from scratch performs best due to distributional shifts. \\
\cmidrule(lr){1-2}

\multirow{3}{1in}{Model size relationship with pretraining} &
Larger models like \textsf{segformer} and \textsf{unet} (ResNet50) perform well when pretrained on other datasets and finetuned on \texttt{FaultSeg3D}. Smaller models like \textsf{rcf} and \textsf{hed} degrade in performance with pretraining, indicating a lack of transfer capacity.\\
\cmidrule(lr){1-2}

\multirow{2}{1in}{Fault density} &
\texttt{FaultSeg3D} and \texttt{CRACKS} have dense faults; \texttt{Thebe} faults are sparse, affecting model prediction density. \\
\cmidrule(lr){1-2}

\multirow{1}{1.4in}{Joint \texttt{CRACKS}-\texttt{FaultSeg3D}} &
Combining \texttt{CRACKS} and \texttt{FaultSeg3D} data leads to synergistic features and better results. \\
\cmidrule(lr){1-2}

\multirow{2}{1in}{Joint training with \texttt{Thebe}} &
Adding \texttt{Thebe} acts as a regularizer: performance drops on original domains but improves generalization. \\
\cmidrule(lr){1-2}

\multirow{2}{1in}{Domain adaptation} &
FDA and DANN improve large-shift transfers (i.e. \texttt{FaultSeg3D} to \texttt{Thebe}) but degrade performance in aligned domains (i.e. \texttt{FaultSeg3D} to \texttt{CRACKS}). \\
\cmidrule(lr){1-2}

{\textsf{deeplab} behavior} & Produces jagged or stair-like faults. \\
\cmidrule(lr){1-2}

{\textsf{segformer} behavior} & Tends to generate thick, blob-like faults. \\
\cmidrule(lr){1-2}

{\textsf{unet}/\textsf{unet++} behavior} & \textsf{unet} creates thin faults; \textsf{unet++} tends to produce fragmented ones. \\
\cmidrule(lr){1-2}

\multirow{1}{1in}{\textsf{hed}/\textsf{rcf} behavior} &
Less adaptable to \texttt{Thebe} due to fault density mismatch; outputs noisy, distorted shapes. \\
\cmidrule(lr){1-2}

\multirow{2}{1.2in}{Loss–context relationship} &
Dice loss benefits from larger window sizes (captures fault continuity), while BCE is more effective with smaller patches due to class imbalance. \\
\cmidrule(lr){1-2}

\multirow{2}{1.2in}{Metric structural biases} &
Dice penalizes misshaped faults more, while Hausdorff/BCD may still give high scores due to proximity. \\
\cmidrule(lr){1-2}

\multirow{1}{1.2in}{Metric sparsity biases} &
Fewer faults lead to harsher penalty in distance metrics; dense faults often score better. \\
\cmidrule(lr){1-2}

\multirow{2}{1.2in}{Fault characteristic transfer} &
Models inherit structural/geometric biases of finetuning dataset. For instance, \texttt{CRACKS} fragmented and crossover faults, while \texttt{FaultSeg3D} induces smoother and disconnected faults. \\
\bottomrule
\end{tabular}
}
\end{table*}
}

%\input{Tables/96vs256}

% \begin{figure}[htbp]
%   \centering
  
%   % First subfigure (side-by-side)
%   \begin{subfigure}{0.48\linewidth}
%     \centering
%     \includegraphics[width=\linewidth]{Figures/96i.png}
%     \caption{Results of one sample using 96 by 96 window size.}
%     \label{fig:96}
%   \end{subfigure}
%   \hfill
%   % Second subfigure
%   \begin{subfigure}{0.48\linewidth}
%     \centering
%     \includegraphics[width=\linewidth]{Figures/256i.png}
%     \caption{Results of using an adaptive window size.}
%     \label{fig:256}
%   \end{subfigure}
  
%   \caption{Comparison between using a fixed and adaptive window size.}
%   \label{fig:96vs256_s}
% \end{figure}

% Preamble must have:
% \usepackage[figuresleft]{rotating}   % rotates sidewaystable the other way
% \usepackage{booktabs,array,graphicx} % graphicx gives \resizebox

\begin{sidewaystable*}[!ht]
\centering
\caption{Out-of-the-box model pick list based on dataset fault–network statistics (\textbf{Table~\ref{tab:dataset_stats}}) and cross-setup behavior (\textbf{Table~\ref{tab:training_stats}}). Label each metric as High/Medium/Low (H/M/L) relative to \textbf{Table~\ref{tab:dataset_stats}}, then select the row that matches; confirm with \textbf{Table~\ref{tab:training_stats}}.}
\label{tab:model_selection_access_rot}

% ---- ROW/COLUMN TWEAKS ----
\renewcommand{\arraystretch}{1.30}  % <<< WIDEN ROWS (increase for more vertical spacing)
\setlength{\extrarowheight}{1pt}    % <<< Optional extra row height
\setlength{\tabcolsep}{1.8pt}       % <<< THIN COLUMNS (reduce inter-column padding)

% Scale the table width relative to page height in landscape.
% Make columns thinner by lowering the 0.92 factor; raise it if too small.
\resizebox{0.92\textheight}{!}{% <<< MAIN WIDTH KNOB (e.g., 0.88–0.95)
\begin{tabular}{%
  >{\raggedright\arraybackslash}p{0.26\textwidth}%
  >{\raggedright\arraybackslash}p{0.22\textwidth}%
  >{\raggedright\arraybackslash}p{0.12\textwidth}%
  >{\raggedright\arraybackslash}p{0.18\textwidth}%
  >{\raggedright\arraybackslash}p{0.14\textwidth}}
\toprule
\textbf{Dataset profile (vs. Table~\ref{tab:dataset_stats})} &
\textbf{General pick rule (what the model should do)} &
\textbf{Setup Suggestion} &
\textbf{Simple rationale (from Table~\ref{tab:training_stats})} &
\textbf{Quick fixes after prediction} \\
\midrule
\textbf{Tortuous \& highly segmented with frequent stepovers} \newline
\emph{Triggers:} Sinuosity \textbf{H}, Segments \textbf{H}, Stepover density \textbf{H}, Curvature L–M, Length M
&
Pick a model that \emph{keeps tortuosity and stepovers close to the data} (sinuosity ratio $\approx 1$, stepover ratio $\approx 1$), even if total length is slightly low.
&
\texttt{Thebe$\rightarrow$CRACKS}
&
Best on braided/fragmented patterns; keeps sinuosity and stepover frequency closest to ground truth.
&
Merge near-collinear pieces; prune short spurs; bridge small gaps. \\[0.8ex]

\textbf{Long, continuous master faults; sparse stepovers (few segments)} \newline
\emph{Triggers:} Length \textbf{H}, Segments \textbf{L}, Stepover density \textbf{L}, Sinuosity L–M, Curvature M
&
Pick a model that \emph{preserves total length and continuity} (length ratio $\approx 1$) and avoids unnecessary breaks.
&
\texttt{Thebe$\rightarrow$FS}
&
Best length preservation with stable stepovers; avoids over-meandering.
&
Raise connect/NMS thresholds; post-merge nearly collinear segments. \\[0.8ex]

\textbf{Balanced / uncertain regime (no extremes) or new basin} \newline
\emph{Triggers:} All metrics within $\pm 1\sigma$ of Thebe (Table~\ref{tab:dataset_stats}) or mixed signals
&
Start with a \emph{neutral, orientation-faithful} model (high strike similarity) before specializing.
&
\texttt{Thebe Only}
&
Highest strike similarity and stable curvature; safe default when unsure.
&
Enforce minimum segment length; trim dead-ends; lightly straighten detours. \\[0.8ex]

\textbf{High curvature \emph{and} high tortuosity (complex relays)} \newline
\emph{Triggers:} Curvature \textbf{H}, Sinuosity \textbf{H}, Stepover density M, Segments M–H
&
Pick a model that \emph{tracks bends and meanders together}: keep sinuosity near $1$ and avoid flattening sharp turns.
&
Start \texttt{Thebe$\rightarrow$CRACKS}; if bends look underfit, switch \texttt{Thebe$\rightarrow$FS}
&
\texttt{CRACKS} fits fragmented/high-tortuosity cases; \texttt{FS} better preserves local bending.
&
Merge then apply mild spline smoothing; avoid aggressive thinning. \\[0.8ex]

\textbf{Stepover-dominated networks with moderate tortuosity} \newline
\emph{Triggers:} Stepover density \textbf{H}, Sinuosity M, Segments M–H, Length M, Curvature L–M
&
Pick a model that \emph{gets the number of stepovers right} (stepover ratio $\approx 1$) while keeping sinuosity reasonable.
&
\texttt{Thebe$\rightarrow$CRACKS}
&
Matches stepover frequency best; robust to fragmented relay zones.
&
Topology cleanup: collapse tiny relays; enforce minimum relay width; reconnect near-parallel strands. \\
\bottomrule
\end{tabular}%
}% end resizebox

\vspace{4pt}
\footnotesize
\textbf{Abbreviations.} H/M/L: high/medium/low relative to \textbf{Table~\ref{tab:dataset_stats}}. 
Length ratio $= L_{\text{pred}}/L_{\text{gt}}$; 
Sinuosity ratio $= S_{\text{pred}}/S_{\text{gt}}$; 
Stepover ratio $= D_{\text{stepover}}^{\text{pred}}/D_{\text{stepover}}^{\text{gt}}$; 
Strike similarity: cosine similarity of strike histograms (see \textbf{Table~\ref{tab:training_stats}}). 
Use \textbf{Table~\ref{tab:dataset_stats}} to assign H/M/L triggers; confirm with \textbf{Table~\ref{tab:training_stats}} (ratios near $1$, small differences).
\end{sidewaystable*}

% Appendixes, if needed, appear before the acknowledgment.

\newpage
\twocolumn
\section*{Acknowledgment}
This work is supported by ML4Seismic Industry Partners at the Georgia Institute of Technology.  

% The preferred spelling of the word ``acknowledgment'' in American English is 
% without an ``e'' after the ``g.'' Use the singular heading even if you have 
% many acknowledgments. Avoid expressions such as ``One of us (S.B.A.) would 
% like to thank $\ldots$ .'' Instead, write ``F. A. Author thanks $\ldots$ .'' In most 
% cases, sponsor and financial support acknowledgments are placed in the 
% unnumbered footnote on the first page, not here.

\bibliographystyle{IEEEbib} %plainnat
\bibliography{ref}

@article{prabhushankar2024cracks,
  title={CRACKS: Crowdsourcing Resources for Analysis and Categorization of Key Subsurface faults},
  author={Prabhushankar, Mohit and Kokilepersaud, Kiran and Quesada, Jorge and Yarici, Yavuz and Zhou, Chen and Alotaibi, Mohammad and AlRegib, Ghassan and Mustafa, Ahmad and Kumakov, Yusufjon},
  journal={arXiv preprint arXiv:2408.11185},
  year={2024}
}

@article{benkert2024effective,
  title={Effective data selection for seismic interpretation through disagreement},
  author={Benkert, Ryan and Prabhushankar, Mohit and AlRegib, Ghassan},
  journal={IEEE Transactions on Geoscience and Remote Sensing},
  year={2024},
  publisher={IEEE}
}

@article{alaudah2018structure,
  title={Structure label prediction using similarity-based retrieval and weakly supervised label mapping},
  author={Alaudah, Yazeed and Alfarraj, Motaz and AlRegib, Ghassan},
  journal={Geophysics},
  volume={84},
  number={1},
  pages={V67--V79},
  year={2018},
  publisher={Society of Exploration Geophysicists}
}

@inproceedings{shafiq2018towards,
  title={Towards understanding common features between natural and seismic images},
  author={Shafiq, Muhammad A and Prabhushankar, Mohit and Di, Haibin and AlRegib, Ghassan},
  booktitle={SEG International Exposition and Annual Meeting},
  pages={SEG--2018},
  year={2018},
  organization={SEG}
}

@article{LI2023105412,
title = {Fault-Seg-Net: A method for seismic fault segmentation based on multi-scale feature fusion with imbalanced classification},
journal = {Computers and Geotechnics},
volume = {158},
pages = {105412},
year = {2023},
issn = {0266-352X},
doi = {https://doi.org/10.1016/j.compgeo.2023.105412},
url = {https://www.sciencedirect.com/science/article/pii/S0266352X23001696},
author = {Xiao Li and Kewen Li and Zhifeng Xu and Zongchao Huang and Yimin Dou}
}

@inproceedings{di2017seismic,
  title={Seismic-fault detection based on multiattribute support vector machine analysis},
  author={Di, Haibin and Shafiq, Muhammad Amir and AlRegib, Ghassan},
  booktitle={SEG International Exposition and Annual Meeting},
  pages={SEG--2017},
  year={2017},
  organization={SEG}
}

@article{cohen2006detection,
  title={Detection and extraction of fault surfaces in 3D seismic data},
  author={Cohen, Israel and Coult, Nicholas and Vassiliou, Anthony A},
  journal={Geophysics},
  volume={71},
  number={4},
  pages={P21--P27},
  year={2006},
  publisher={Society of Exploration Geophysicists}
}

@article{roberts2001curvature,
  title={Curvature attributes and their application to 3 D interpreted horizons},
  author={Roberts, Andy},
  journal={First break},
  volume={19},
  number={2},
  pages={85--100},
  year={2001}
}

@InProceedings{Quesada_2024_CVPR,
    author    = {Quesada, Jorge and Alotaibi, Mohammad and Prabhushankar, Mohit and Alregib, Ghassan},
    title     = {PointPrompt: A Multi-modal Prompting Dataset for Segment Anything Model},
    booktitle = {Proceedings of the IEEE/CVF Conference on Computer Vision and Pattern Recognition (CVPR) Workshops},
    month     = {June},
    year      = {2024},
    pages     = {1604-1610}
}

@article{mustafa2024visual,
  title={Visual attention guided learning with incomplete labels for seismic fault interpretation},
  author={Mustafa, Ahmad and Rastegar, Reza and Brown, Tim and Nunes, Gregory and DeLilla, Daniel and AlRegib, Ghassan},
  journal={IEEE Transactions on Geoscience and Remote Sensing},
  year={2024},
  publisher={IEEE}
}

@inproceedings{quesada2024benchmarking,
  title={Benchmarking Human and Automated Prompting in the Segment Anything Model},
  author={Quesada, Jorge and Fowler, Zoe and Alotaibi, Mohammad and Prabhushankar, Mohit and AlRegib, Ghassan},
  booktitle={2024 IEEE International Conference on Big Data (BigData)},
  pages={1625--1634},
  year={2024},
  organization={IEEE}
}

@article{chowdhury2025unified,
  title={A unified framework for evaluating robustness of Machine Learning Interpretability for Prospect Risking},
  author={Chowdhury, Prithwijit and Mustafa, Ahmad and Prabhushankar, Mohit and AlRegib, Ghassan},
  journal={Geophysics},
  volume={90},
  number={3},
  pages={1--53},
  year={2025},
  publisher={Society of Exploration Geophysicists}
}

@article{iqbal2023blind,
  title={Blind curvelet-based denoising of seismic surveys in coherent and incoherent noise environments},
  author={Iqbal, Naveed and Deriche, Mohamed and AlRegib, Ghassan and Khan, Sikandar},
  journal={Arabian Journal for Science and Engineering},
  volume={48},
  number={8},
  pages={10925--10935},
  year={2023},
  publisher={Springer}
}

@article{mustafa2023active,
  title={Active learning with deep autoencoders for seismic facies interpretation},
  author={Mustafa, Ahmad and AlRegib, Ghassan},
  journal={Geophysics},
  volume={88},
  number={4},
  pages={IM77--IM86},
  year={2023},
  publisher={Society of Exploration Geophysicists}
}

@article{mustafa2024explainable,
  title={Explainable machine learning for hydrocarbon prospect risking},
  author={Mustafa, Ahmad and Koster, Klaas and AlRegib, Ghassan},
  journal={Geophysics},
  volume={89},
  number={1},
  pages={WA13--WA24},
  year={2024},
  publisher={Society of Exploration Geophysicists}
}

@inproceedings{chowdhury2023counterfactual,
  title={Counterfactual uncertainty for high dimensional tabular dataset},
  author={Chowdhury, Prithwijit and Mustafa, Ahmad and Prabhushankar, Mohit and AlRegib, Ghassan},
  booktitle={SEG International Exposition and Annual Meeting},
  pages={SEG--2023},
  year={2023},
  organization={SEG}
}

@inproceedings{benkert2023samples,
  title={What samples must seismic interpreters label for efficient machine learning?},
  author={Benkert, Ryan and Prabhushankar, Mohit and AlRegib, Ghassan},
  booktitle={Third International Meeting for Applied Geoscience \& Energy},
  pages={1004--1009},
  year={2023},
  organization={Society of Exploration Geophysicists and American Association of Petroleum~…}
}

@inproceedings{zhou2023perceptual,
  title={Perceptual quality-based model training under annotator label uncertainty},
  author={Zhou, Chen and Prabhushankar, Mohit and AlRegib, Ghassan},
  booktitle={SEG International Exposition and Annual Meeting},
  pages={SEG--2023},
  year={2023},
  organization={SEG}
}

@article{wang2016interactive,
  title={Interactive fault extraction in 3-D seismic data using the Hough transform and tracking vectors},
  author={Wang, Zhen and AlRegib, Ghassan},
  journal={IEEE Transactions on Computational Imaging},
  volume={3},
  number={1},
  pages={99--109},
  year={2016},
  publisher={IEEE}
}

@inproceedings{kokilepersaud2022volumetric,
  title={Volumetric supervised contrastive learning for seismic semantic segmentation},
  author={Kokilepersaud, Kiran and Prabhushankar, Mohit and AlRegib, Ghassan},
  booktitle={Second International Meeting for Applied Geoscience \& Energy},
  pages={1699--1703},
  year={2022},
  organization={Society of Exploration Geophysicists and American Association of Petroleum~…}
}

@article{benkert2022example,
  title={Example forgetting: A novel approach to explain and interpret deep neural networks in seismic interpretation},
  author={Benkert, Ryan and Aribido, Oluwaseun Joseph and AlRegib, Ghassan},
  journal={IEEE Transactions on Geoscience and Remote Sensing},
  volume={60},
  pages={1--12},
  year={2022},
  publisher={IEEE}
}

@inproceedings{benkert2022reliable,
  title={Reliable uncertainty estimation for seismic interpretation with prediction switches},
  author={Benkert, Ryan and Prabhushankar, Mohit and AlRegib, Ghassan},
  booktitle={Second International Meeting for Applied Geoscience \& Energy},
  pages={1740--1744},
  year={2022},
  organization={Society of Exploration Geophysicists and American Association of Petroleum~…}
}

@article{shafiq2022novel,
  title={A novel attention model for salient structure detection in seismic volumes},
  author={Shafiq, Muhammad Amir and Long, Zhiling and Di, Haibin and AlRegib, Ghassan},
  journal={arXiv preprint arXiv:2201.06174},
  year={2022}
}

@article{aribido2021self,
  title={Self-supervised delineation of geologic structures using orthogonal latent space projection},
  author={Aribido, Oluwaseun Joseph and AlRegib, Ghassan and Alaudah, Yazeed},
  journal={Geophysics},
  volume={86},
  number={6},
  pages={V497--V508},
  year={2021},
  publisher={Society of Exploration Geophysicists}
}

@inproceedings{benkert2021explaining,
  title={Explaining deep models through forgettable learning dynamics},
  author={Benkert, Ryan and Aribido, Oluwaseun Joseph and AlRegib, Ghassan},
  booktitle={2021 IEEE International Conference on Image Processing (ICIP)},
  pages={3692--3696},
  year={2021},
  organization={IEEE}
}

@inproceedings{benkert2021explainable,
  title={Explainable seismic neural networks using learning statistics},
  author={Benkert, Ryan and Joseph Aribido, Oluwaseun and AlRegib, Ghassan},
  booktitle={First International Meeting for Applied Geoscience \& Energy},
  pages={1425--1429},
  year={2021},
  organization={Society of Exploration Geophysicists}
}

@article{mustafa2021joint,
  title={Joint learning for spatial context-based seismic inversion of multiple data sets for improved generalizability and robustness},
  author={Mustafa, Ahmad and Alfarraj, Motaz and AlRegib, Ghassan},
  journal={Geophysics},
  volume={86},
  number={4},
  pages={O37--O48},
  year={2021},
  publisher={Society of Exploration Geophysicists}
}

@inproceedings{mustafa2021comparative,
  title={A comparative study of transfer learning methodologies and causality for seismic inversion with temporal convolutional networks},
  author={Mustafa, Ahmad and AlRegib, Ghassan},
  booktitle={SEG International Exposition and Annual Meeting},
  pages={D011S067R001},
  year={2021},
  organization={SEG}
}

@inproceedings{aribido2020self,
  title={Self-supervised annotation of seismic images using latent space factorization},
  author={Aribido, Oluwaseun Joseph and AlRegib, Ghassan and Deriche, Mohamed},
  booktitle={2020 IEEE International Conference on Image Processing (ICIP)},
  pages={2421--2425},
  year={2020},
  organization={IEEE}
}

@inproceedings{soliman2020s,
  title={S 6: semi-supervised self-supervised semantic segmentation},
  author={Soliman, Moamen and Lehman, Charles and AlRegib, Ghassan},
  booktitle={2020 IEEE International Conference on Image Processing (ICIP)},
  pages={1861--1865},
  year={2020},
  organization={IEEE}
}

@incollection{mustafa2020joint,
  title={Joint learning for seismic inversion: An acoustic impedance estimation case study},
  author={Mustafa, Ahmad and AlRegib, Ghassan},
  booktitle={SEG Technical Program Expanded Abstracts 2020},
  pages={1686--1690},
  year={2020},
  publisher={Society of Exploration Geophysicists}
}

@inproceedings{mustafa2020spatiotemporal,
  title={Spatiotemporal modeling of seismic images for acoustic impedance estimation},
  author={Mustafa, Ahmad and Alfarraj, Motaz and AlRegib, Ghassan},
  booktitle={SEG International Exposition and Annual Meeting},
  pages={D041S101R005},
  year={2020},
  organization={SEG}
}

@article{di2019developing,
  title={Developing a seismic texture analysis neural network for machine-aided seismic pattern recognition and classification},
  author={Di, Haibin and Gao, Dengliang and AlRegib, Ghassan},
  journal={Geophysical Journal International},
  volume={218},
  number={2},
  pages={1262--1275},
  year={2019},
  publisher={Oxford University Press}
}

@INPROCEEDINGS{576361,
  author={Dubuisson, M.-P. and Jain, A.K.},
  booktitle={Proceedings of 12th International Conference on Pattern Recognition}, 
  title={A modified Hausdorff distance for object matching}, 
  year={1994},
  volume={1},
  number={},
  pages={566-568 vol.1},
  doi={10.1109/ICPR.1994.576361}}

@inproceedings{alaudah2019facies,
  title={Facies classification with weak and strong supervision: A comparative study},
  author={Alaudah, Yazeed and Soliman, Moamen and AlRegib, Ghassan},
  booktitle={SEG International Exposition and Annual Meeting},
  pages={D033S037R004},
  year={2019},
  organization={SEG}
}

@incollection{mustafa2019estimation,
  title={Estimation of acoustic impedance from seismic data using temporal convolutional network},
  author={Mustafa, Ahmad and Alfarraj, Motaz and AlRegib, Ghassan},
  booktitle={SEG technical program expanded abstracts 2019},
  pages={2554--2558},
  year={2019},
  publisher={Society of Exploration Geophysicists}
}

@incollection{alfarraj2019semi,
  title={Semi-supervised learning for acoustic impedance inversion},
  author={Alfarraj, Motaz and AlRegib, Ghassan},
  booktitle={SEG technical program expanded abstracts 2019},
  pages={2298--2302},
  year={2019},
  publisher={Society of Exploration Geophysicists}
}

@inproceedings{alaudah2015curvelet,
  title={A curvelet-based distance measure for seismic images},
  author={Alaudah, Yazeed and AlRegib, Ghassan},
  booktitle={2015 IEEE International Conference on Image Processing (ICIP)},
  pages={4200--4204},
  year={2015},
  organization={IEEE}
}

@article{alfarraj2018multiresolution,
  title={Multiresolution analysis and learning for computational seismic interpretation},
  author={Alfarraj, Motaz and Alaudah, Yazeed and Long, Zhiling and AlRegib, Ghassan},
  journal={The Leading Edge},
  volume={37},
  number={6},
  pages={443--450},
  year={2018},
  publisher={Society of Exploration Geophysicists}
}

@article{long2018comparative,
  title={A comparative study of texture attributes for characterizing subsurface structures in seismic volumes},
  author={Long, Zhiling and Alaudah, Yazeed and Ali Qureshi, Muhammad and Hu, Yuting and Wang, Zhen and Alfarraj, Motaz and AlRegib, Ghassan and Amin, Asjad and Deriche, Mohamed and Al-Dharrab, Suhail and others},
  journal={Interpretation},
  volume={6},
  number={4},
  pages={T1055--T1066},
  year={2018},
  publisher={Society of Exploration Geophysicists and American Association of Petroleum~…}
}

@incollection{long2015characterization,
  title={Characterization of migrated seismic volumes using texture attributes: a comparative study},
  author={Long*, Zhiling and Alaudah, Yazeed and Qureshi, Muhammad Ali and Farraj, Motaz Al and Wang, Zhen and Amin, Asjad and Deriche, Mohamed and AlRegib, Ghassan},
  booktitle={SEG Technical Program Expanded Abstracts 2015},
  pages={1744--1748},
  year={2015},
  publisher={Society of Exploration Geophysicists}
}

@inproceedings{alaudah2018learning,
  title={Learning to label seismic structures with deconvolution networks and weak labels},
  author={Alaudah, Yazeed and Gao, Shan and AlRegib, Ghassan},
  booktitle={SEG international exposition and annual meeting},
  pages={SEG--2018},
  year={2018},
  organization={SEG}
}

@misc{dgb1987netherlands,
  title={The Netherlands Offshore, The North Sea, F3 Block—Complete},
  author={dGB Earth Sciences},
  year={1987},
  publisher={dGB Earth Sciences}
}

@article{ishak2018application,
  title={The application of seismic attributes and wheeler transformations for the geomorphological interpretation of stratigraphic surfaces: a case study of the f3 block, Dutch offshore sector, north sea},
  author={Ishak, Mohammad Afifi and Islam, Md Aminul and Shalaby, Mohamed Ragab and Hasan, Nurul},
  journal={Geosciences},
  volume={8},
  number={3},
  pages={79},
  year={2018},
  publisher={MDPI}
}

@article{safari2023structural,
  title={Structural smoothing on mixed instantaneous phase energy for automatic fault and horizon picking: case study on F3 North Sea},
  author={Safari, Mohammad Reza and Taheri, Kioumars and Hashemi, Hosein and Hadadi, Ali},
  journal={Journal of Petroleum Exploration and Production Technology},
  volume={13},
  number={3},
  pages={775--785},
  year={2023},
  publisher={Springer}
}

@article{an2023current,
  title={Current state and future directions for deep learning based automatic seismic fault interpretation: A systematic review},
  author={An, Yu and Du, Haiwen and Ma, Siteng and Niu, Yingjie and Liu, Dairui and Wang, Jing and Du, Yuhan and Childs, Conrad and Walsh, John and Dong, Ruihai},
  journal={Earth-Science Reviews},
  volume={243},
  pages={104509},
  year={2023},
  publisher={Elsevier}
}

@inproceedings{mustafa2021man,
  title={Man-recon: Manifold learning for reconstruction with deep autoencoder for smart seismic interpretation},
  author={Mustafa, Ahmad and AlRegib, Ghassan},
  booktitle={2021 IEEE International Conference on Image Processing (ICIP)},
  pages={2953--2957},
  year={2021},
  organization={IEEE}
}

@article{AN2021107219,
title = {A gigabyte interpreted seismic dataset for automatic fault recognition},
journal = {Data in Brief},
volume = {37},
pages = {107219},
year = {2021},
issn = {2352-3409},
doi = {https://doi.org/10.1016/j.dib.2021.107219},
url = {https://www.sciencedirect.com/science/article/pii/S2352340921005035},
author = {Yu An and Jiulin Guo and Qing Ye and Conrad Childs and John Walsh and Ruihai Dong}
}

@article{bi2021deep,
  title={Deep relative geologic time: a deep learning method for simultaneously interpreting 3-D seismic horizons and faults},
  author={Bi, Zhengfa and Wu, Xinming and Geng, Zhicheng and Li, Haishan},
  journal={Journal of Geophysical Research: Solid Earth},
  volume={126},
  number={9},
  pages={e2021JB021882},
  year={2021},
  publisher={Wiley Online Library}
}

@article{lin2022automatic,
  title={Automatic geologic fault identification from seismic data using 2.5 D channel attention U-net},
  author={Lin, Lei and Zhong, Zhi and Cai, Zhongxian and Sun, Alexander Y and Li, ChengLong},
  journal={Geophysics},
  volume={87},
  number={4},
  pages={IM111--IM124},
  year={2022},
  publisher={Society of Exploration Geophysicists}
}

@article{pochet2018seismic,
  title={Seismic fault detection using convolutional neural networks trained on synthetic poststacked amplitude maps},
  author={Pochet, Axelle and Diniz, Pedro HB and Lopes, H{\'e}lio and Gattass, Marcelo},
  journal={IEEE Geoscience and Remote Sensing Letters},
  volume={16},
  number={3},
  pages={352--356},
  year={2018},
  publisher={IEEE}
}

@inproceedings{faultsam,
author = {Chen, Ran and Zhang, Zeren and Ma, Jinwen},
year = {2024},
month = {10},
pages = {436-441},
title = {Seismic Fault SAM: Adapting SAM with Lightweight Modules and 2.5D Strategy for Fault Detection},
doi = {10.1109/ICSP62129.2024.10846297}
}

@Article{rs16050922,
AUTHOR = {Zhang, Zeren and Chen, Ran and Ma, Jinwen},
TITLE = {Improving Seismic Fault Recognition with Self-Supervised Pre-Training: A Study of 3D Transformer-Based with Multi-Scale Decoding and Fusion},
JOURNAL = {Remote Sensing},
VOLUME = {16},
YEAR = {2024},
NUMBER = {5},
ARTICLE-NUMBER = {922},
URL = {https://www.mdpi.com/2072-4292/16/5/922},
ISSN = {2072-4292},
DOI = {10.3390/rs16050922}
}

@article{an2023understanding,
  title={Understanding the effect of different prior knowledge on cnn fault interpreter},
  author={An, Yu and Dong, Ruihai},
  journal={IEEE Access},
  volume={11},
  pages={15058--15068},
  year={2023},
  publisher={IEEE}
}

@article{wu2019faultSeg,
    author = {Xinming Wu and Luming Liang and Yunzhi Shi and Sergey Fomel},
    title = {Fault{S}eg3{D}: using synthetic datasets to train an end-to-end convolutional neural network for 3{D} seismic fault segmentation},
    journal = {GEOPHYSICS},
    volume = {84},
    number = {3},
    pages = {IM35-IM45},
    year = {2019},
}

@article{alaudah2019machine,
  title={A machine-learning benchmark for facies classification},
  author={Alaudah, Yazeed and Micha{\l}owicz, Patrycja and Alfarraj, Motaz and AlRegib, Ghassan},
  journal={Interpretation},
  volume={7},
  number={3},
  pages={SE175--SE187},
  year={2019},
  publisher={Society of Exploration Geophysicists and American Association of Petroleum~…}
}

@article{alfarraj2019semisupervised,
  title={Semisupervised sequence modeling for elastic impedance inversion},
  author={Alfarraj, Motaz and AlRegib, Ghassan},
  journal={Interpretation},
  volume={7},
  number={3},
  pages={SE237--SE249},
  year={2019},
  publisher={Society of Exploration Geophysicists and American Association of Petroleum~…}
}

@article{di2019improving,
  title={Improving seismic fault detection by super-attribute-based classification},
  author={Di, Haibin and Shafiq, Mohammod Amir and Wang, Zhen and AlRegib, Ghassan},
  journal={Interpretation},
  volume={7},
  number={3},
  pages={SE251--SE267},
  year={2019},
  publisher={Society of Exploration Geophysicists and American Association of Petroleum~…}
}

@article{di2019semi,
  title={Semi-automatic fault/fracture interpretation based on seismic geometry analysis},
  author={Di, Haibin and AlRegib, Ghassan},
  journal={Geophysical Prospecting},
  volume={67},
  number={5},
  pages={1379--1391},
  year={2019},
  publisher={European Association of Geoscientists \& Engineers}
}

@article{di2019reflector,
  title={Reflector dip estimates based on seismic waveform curvature/flexure analysis},
  author={Di, Haibin and AlRegib, Ghassan},
  journal={Interpretation},
  volume={7},
  number={2},
  pages={SC1--SC9},
  year={2019},
  publisher={Society of Exploration Geophysicists and American Association of Petroleum~…}
}

@inproceedings{di2018seismic,
  title={Seismic fault detection from post-stack amplitude by convolutional neural networks},
  author={Di, Haibin and Wang, Zhen and AlRegib, Ghassan},
  booktitle={80th EAGE Conference and Exhibition 2018},
  volume={2018},
  number={1},
  pages={1--5},
  year={2018},
  organization={European Association of Geoscientists \& Engineers}
}

@article{an2021deep,
  title={Deep convolutional neural network for automatic fault recognition from 3D seismic datasets},
  author={An, Yu and Guo, Jiulin and Ye, Qing and Childs, Conrad and Walsh, John and Dong, Ruihai},
  journal={Computers \& Geosciences},
  volume={153},
  pages={104776},
  year={2021},
  publisher={Elsevier}
}

@inproceedings{an2020overlap,
  title={Overlap training to mitigate inconsistencies caused by image tiling in CNNs},
  author={An, Yu and Ye, Qing and Guo, Jiulin and Dong, Ruihai},
  booktitle={International Conference on Innovative Techniques and Applications of Artificial Intelligence},
  pages={35--48},
  year={2020},
  organization={Springer}
}

@inproceedings{chen2018encoder,
  title={Encoder-decoder with atrous separable convolution for semantic image segmentation},
  author={Chen, Liang-Chieh and Zhu, Yukun and Papandreou, George and Schroff, Florian and Adam, Hartwig},
  booktitle={Proceedings of the European conference on computer vision (ECCV)},
  pages={801--818},
  year={2018}
}

@inproceedings{he2016deep,
  title={Deep residual learning for image recognition},
  author={He, Kaiming and Zhang, Xiangyu and Ren, Shaoqing and Sun, Jian},
  booktitle={Proceedings of the IEEE conference on computer vision and pattern recognition},
  pages={770--778},
  year={2016}
}

@inproceedings{sandler2018mobilenetv2,
  title={MobileNetV2: Inverted residuals and linear bottlenecks},
  author={Sandler, Mark and Howard, Andrew and Zhu, Menglong and Zhmoginov, Andrey and Chen, Liang-Chieh},
  booktitle={Proceedings of the IEEE conference on computer vision and pattern recognition},
  pages={4510--4520},
  year={2018}
}

@inproceedings{xie2015holistically,
  title={Holistically-nested edge detection},
  author={Xie, Saining and Tu, Zhuowen},
  booktitle={Proceedings of the IEEE international conference on computer vision},
  pages={1395--1403},
  year={2015}
}

@inproceedings{liu2017richer,
  title={Richer convolutional features for edge detection},
  author={Liu, Yun and Cheng, Song and Hu, Yunchao and Wang, Yandong and Bai, Xiang and Yuille, Alan L},
  booktitle={Proceedings of the IEEE conference on computer vision and pattern recognition},
  pages={3000--3009},
  year={2017}
}

@inproceedings{ronneberger2015u,
  title={U-net: Convolutional networks for biomedical image segmentation},
  author={Ronneberger, Olaf and Fischer, Philipp and Brox, Thomas},
  booktitle={International Conference on Medical image computing and computer-assisted intervention},
  pages={234--241},
  year={2015},
  organization={Springer}
}

@inproceedings{zhou2018unet++,
  title={Unet++: A nested u-net architecture for medical image segmentation},
  author={Zhou, Zongwei and Siddiquee, Md Mahfuzur Rahman and Tajbakhsh, Nima and Liang, Jianming},
  booktitle={Deep learning in medical image analysis and multimodal learning for clinical decision support},
  pages={3--11},
  year={2018},
  organization={Springer}
}

@inproceedings{xie2021segformer,
  title={SegFormer: Simple and efficient design for semantic segmentation with transformers},
  author={Xie, Enze and Wang, Wenhai and Yu, Zhiding and Anandkumar, Anima and Alvarez, Jose M and Luo, Ping},
  booktitle={Advances in Neural Information Processing Systems},
  volume={34},
  pages={12077--12090},
  year={2021}
}

@inproceedings{sudre2017generalised,
  title={Generalised dice overlap as a deep learning loss function for highly unbalanced segmentations},
  author={Sudre, Carole H and Li, Wenqi and Vercauteren, Tom and Ourselin, Sebastien and Jorge Cardoso, M},
  booktitle={Deep Learning in Medical Image Analysis and Multimodal Learning for Clinical Decision Support: Third International Workshop, DLMIA 2017, and 7th International Workshop, ML-CDS 2017, Held in Conjunction with MICCAI 2017, Qu{\'e}bec City, QC, Canada, September 14, Proceedings 3},
  pages={240--248},
  year={2017},
  organization={Springer}
}

@article{benzion2008collective,
author = {Ben-Zion, Yehuda},
title = {Collective behavior of earthquakes and faults: Continuum-discrete transitions, progressive evolutionary changes, and different dynamic regimes},
journal = {Reviews of Geophysics},
volume = {46},
number = {4},
doi = {https://doi.org/10.1029/2008RG000260},
year = {2008}
}

@incollection{KANAMORI20031205,
title = {72 - Earthquake Prediction: An Overview},
editor = {William H.K. Lee and Hiroo Kanamori and Paul C. Jennings and Carl Kisslinger},
series = {International Geophysics},
publisher = {Academic Press},
volume = {81},
pages = {1205-1216},
year = {2003},
booktitle = {International Handbook of Earthquake and Engineering Seismology, Part B},
issn = {0074-6142},
doi = {https://doi.org/10.1016/S0074-6142(03)80186-9},
url = {https://www.sciencedirect.com/science/article/pii/S0074614203801869},
author = {Hiroo Kanamori}
}

@article{nasim2020seismic,
  title={Seismic Facies Analysis: A Deep Domain Adaptation Approach},
  author={Nasim, M Quamer and Maiti, Tannistha and Srivastava, Ayush and Singh, Tarry and Mei, Jie},
  journal={arXiv preprint arXiv:2011.10510},
  year={2020}
}

@article{alcalde2017impact,
  title={Impact of seismic image quality on fault interpretation uncertainty},
  author={Alcalde, Juan and Bond, Clare E. and Johnson, Gareth and Ellis, Jennifer F. and Butler, Robert W.H.},
  journal={GSA Today},
  volume={27},
  number={2},
  pages={4--10},
  year={2017},
  publisher={Geological Society of America},
  doi={10.1130/GSATG282A.1}
}

@inproceedings{sarajarvi2020robust,
  title={Robust Evaluation of Fault Prediction Results: Machine Learning Using Synthetic Seismic},
  author={Sarajärvi, M. and Hellem Bo, T. and Goledowski, B. and Nickel, M.},
  booktitle={First EAGE Digitalization Conference and Exhibition},
  year={2020},
  publisher={European Association of Geoscientists \& Engineers},
  doi={10.3997/2214-4609.202032015}
}

@article{guillon2020ground,
  title={Ground-truth uncertainty-aware metrics for machine learning applications on seismic image interpretation: Application to faults and horizon extraction},
  author={Guillon, S{\'e}bastien and Joncour, Fr{\'e}d{\'e}ric and Barrallon, Pierre-Emmanuel and Castani{\'e}, Laurent},
  journal={The Leading Edge},
  volume={39},
  number={10},
  pages={734--741},
  year={2020},
  publisher={Society of Exploration Geophysicists},
  doi={10.1190/tle39100734.1}
}

@inproceedings{sun2016deep,
  title={Deep coral: Correlation alignment for deep domain adaptation},
  author={Sun, Baochen and Saenko, Kate},
  booktitle={Computer vision--ECCV 2016 workshops: Amsterdam, the Netherlands, October 8-10 and 15-16, 2016, proceedings, part III 14},
  pages={443--450},
  year={2016},
  organization={Springer}
}

@inproceedings{tsai2018learning,
  title={Learning to adapt structured output space for semantic segmentation},
  author={Tsai, Yi-Hsuan and Hung, Wei-Chih and Schulter, Samuel and Sohn, Kihyuk and Yang, Ming-Hsuan and Chandraker, Manmohan},
  booktitle={Proceedings of the IEEE conference on computer vision and pattern recognition},
  pages={7472--7481},
  year={2018}
}

@article{du2022disentangling,
  title={Disentangling noise patterns from seismic images: Noise reduction and style transfer},
  author={Du, Haiwen and An, Yu and Ye, Qing and Guo, Jiulin and Liu, Lu and Zhu, Dongjie and Childs, Conrad and Walsh, John and Dong, Ruihai},
  journal={IEEE Transactions on Geoscience and Remote Sensing},
  volume={60},
  pages={1--14},
  year={2022},
  publisher={IEEE}
}

@article{kirkpatrick2017overcoming,
  title={Overcoming catastrophic forgetting in neural networks},
  author={Kirkpatrick, James and Pascanu, Razvan and Rabinowitz, Neil and Veness, Joel and Desjardins, Guillaume and Rusu, Andrei A and Milan, Kieran and Quan, John and Ramalho, Tiago and Grabska-Barwinska, Agnieszka and others},
  journal={Proceedings of the national academy of sciences},
  volume={114},
  number={13},
  pages={3521--3526},
  year={2017},
  publisher={National Academy of Sciences}
}

@book{goodfellow2016deep,
  title={Deep learning},
  author={Goodfellow, Ian and Bengio, Yoshua and Courville, Aaron and Bengio, Yoshua},
  volume={1},
  number={2},
  year={2016},
  publisher={MIT press Cambridge}
}

@inproceedings{torralba2011unbiased,
  title={Unbiased look at dataset bias},
  author={Torralba, Antonio and Efros, Alexei A},
  booktitle={CVPR 2011},
  pages={1521--1528},
  year={2011},
  organization={IEEE}
}

@article{ganin2016domain,
  title={Domain-adversarial training of neural networks},
  author={Ganin, Yaroslav and Ustinova, Evgeniya and Ajakan, Hana and Germain, Pascal and Larochelle, Hugo and Laviolette, Fran{\c{c}}ois and March, Mario and Lempitsky, Victor},
  journal={Journal of machine learning research},
  volume={17},
  number={59},
  pages={1--35},
  year={2016}
}

@incollection{panigrahi2020survey,
  title={A survey on transfer learning},
  author={Panigrahi, Santisudha and Nanda, Anuja and Swarnkar, Tripti},
  booktitle={Intelligent and Cloud Computing: Proceedings of ICICC 2019, Volume 1},
  pages={781--789},
  year={2020},
  publisher={Springer}
}

@article{yosinski2014transferable,
  title={How transferable are features in deep neural networks?},
  author={Yosinski, Jason and Clune, Jeff and Bengio, Yoshua and Lipson, Hod},
  journal={Advances in neural information processing systems},
  volume={27},
  year={2014}
}

@article{ercoli2023evidencing,
  title={Evidencing subtle faults in deep seismic reflection profiles: Data pre-conditioning and seismic attribute analysis of the legacy CROP-04 profile},
  author={Ercoli, Maurizio and Carboni, Filippo and Akimbekova, Assel and Carbonell, Ramon Bertran and Barchi, Massimiliano Rinaldo},
  journal={Frontiers in Earth Science},
  volume={11},
  pages={1119554},
  year={2023},
  publisher={Frontiers Media SA}
}

@article{dou2022md,
  title={MD loss: Efficient training of 3-D seismic fault segmentation network under sparse labels by weakening anomaly annotation},
  author={Dou, Yimin and Li, Kewen and Zhu, Jianbing and Li, Timing and Tan, Shaoquan and Huang, Zongchao},
  journal={IEEE Transactions on Geoscience and Remote Sensing},
  volume={60},
  pages={1--14},
  year={2022},
  publisher={IEEE}
}

@article{wang2022structural,
  title={Structural augmentation in seismic data for fault prediction},
  author={Wang, Shenghou and Si, Xu and Cai, Zhongxian and Cui, Yatong},
  journal={Applied Sciences},
  volume={12},
  number={19},
  pages={9796},
  year={2022},
  publisher={MDPI}
}

@article{simonyan2014very,
  title={Very deep convolutional networks for large-scale image recognition},
  author={Simonyan, Karen and Zisserman, Andrew},
  journal={arXiv preprint arXiv:1409.1556},
  year={2014}
}

@inproceedings{long2015fully,
  title={Fully convolutional networks for semantic segmentation},
  author={Long, Jonathan and Shelhamer, Evan and Darrell, Trevor},
  booktitle={Proceedings of the IEEE conference on computer vision and pattern recognition},
  pages={3431--3440},
  year={2015}
}

@article{chen2017deeplab,
  title={Deeplab: Semantic image segmentation with deep convolutional nets, atrous convolution, and fully connected crfs},
  author={Chen, Liang-Chieh and Papandreou, George and Kokkinos, Iasonas and Murphy, Kevin and Yuille, Alan L},
  journal={IEEE transactions on pattern analysis and machine intelligence},
  volume={40},
  number={4},
  pages={834--848},
  year={2017},
  publisher={IEEE}
}

@article{isensee2021nnu,
  title={nnU-Net: a self-configuring method for deep learning-based biomedical image segmentation},
  author={Isensee, Fabian and Jaeger, Paul F and Kohl, Simon AA and Petersen, Jens and Maier-Hein, Klaus H},
  journal={Nature methods},
  volume={18},
  number={2},
  pages={203--211},
  year={2021},
  publisher={Nature Publishing Group}
}

@inproceedings{ioffe2015batch,
  title={Batch normalization: Accelerating deep network training by reducing internal covariate shift},
  author={Ioffe, Sergey and Szegedy, Christian},
  booktitle={International conference on machine learning},
  pages={448--456},
  year={2015},
  organization={pmlr}
}

@article{ulyanov2016instance,
  title={Instance normalization: The missing ingredient for fast stylization},
  author={Ulyanov, Dmitry and Vedaldi, Andrea and Lempitsky, Victor},
  journal={arXiv preprint arXiv:1607.08022},
  year={2016}
}

@article{shorten2019survey,
  title={A survey on image data augmentation for deep learning},
  author={Shorten, Connor and Khoshgoftaar, Taghi M},
  journal={Journal of big data},
  volume={6},
  number={1},
  pages={1--48},
  year={2019},
  publisher={Springer}
}

@article{wang2017effectiveness,
  title={The effectiveness of data augmentation in image classification using deep learning},
  author={Wang, Jason and Perez, Luis and others},
  journal={Convolutional Neural Networks Vis. Recognit},
  volume={11},
  number={2017},
  pages={1--8},
  year={2017}
}

@inproceedings{pohlen2017full,
  title={Full-resolution residual networks for semantic segmentation in street scenes},
  author={Pohlen, Tobias and Hermans, Alexander and Mathias, Markus and Leibe, Bastian},
  booktitle={Proceedings of the IEEE conference on computer vision and pattern recognition},
  pages={4151--4160},
  year={2017}
}

@article{krahenbuhl2011efficient,
  title={Efficient inference in fully connected crfs with gaussian edge potentials},
  author={Kr{\"a}henb{\"u}hl, Philipp and Koltun, Vladlen},
  journal={Advances in neural information processing systems},
  volume={24},
  year={2011}
}

@article{prabhushankar2022olives,
  title={Olives dataset: Ophthalmic labels for investigating visual eye semantics},
  author={Prabhushankar, Mohit and Kokilepersaud, Kiran and Logan, Yash-yee and Trejo Corona, Stephanie and AlRegib, Ghassan and Wykoff, Charles},
  journal={Advances in Neural Information Processing Systems},
  volume={35},
  pages={9201--9216},
  year={2022}
}

@inproceedings{prabhushankar2017generating,
  title={Generating adaptive and robust filter sets using an unsupervised learning framework},
  author={Prabhushankar, Mohit and Temel, Dogancan and AlRegib, Ghassan},
  booktitle={2017 IEEE International Conference on Image Processing (ICIP)},
  pages={3041--3045},
  year={2017},
  organization={IEEE}
}

@inproceedings{kokilepersaud2023focal,
  title={Focal: A cost-aware video dataset for active learning},
  author={Kokilepersaud, Kiran and Logan, Yash-Yee and Benkert, Ryan and Zhou, Chen and Prabhushankar, Mohit and AlRegib, Ghassan and Corona, Enrique and Singh, Kunjan and Parchami, Mostafa},
  booktitle={2023 IEEE International Conference on Big Data (BigData)},
  pages={1269--1278},
  year={2023},
  organization={IEEE}
}

@inproceedings{temel2018cure,
  title={Cure-or: Challenging unreal and real environments for object recognition},
  author={Temel, Dogancan and Lee, Jinsol and AlRegib, Ghassan},
  booktitle={2018 17th IEEE international conference on machine learning and applications (ICMLA)},
  pages={137--144},
  year={2018},
  organization={IEEE}
}

@INPROCEEDINGS{Temel2017_NIPS,
author={D. Temel and G. Kwon and M. Prabhushankar and G. AlRegib},
booktitle={Neural Information Processing Systems (NIPS) Workshop on Machine Learning
for Intelligent Transportation Systems (MLITS)},
title={CURE-TSR: Challenging unreal and real environments for traffic sign recognition},
year={2017},
month={December}, }

@article{prabhushankar2022introspective,
  title={Introspective learning: A two-stage approach for inference in neural networks},
  author={Prabhushankar, Mohit and AlRegib, Ghassan},
  journal={Advances in Neural Information Processing Systems},
  volume={35},
  pages={12126--12140},
  year={2022}
}

@article{temel2019traffic,
  title={Traffic sign detection under challenging conditions: A deeper look into performance variations and spectral characteristics},
  author={Temel, Dogancan and Chen, Min-Hung and AlRegib, Ghassan},
  journal={IEEE Transactions on Intelligent Transportation Systems},
  volume={21},
  number={9},
  pages={3663--3673},
  year={2019},
  publisher={IEEE}
}

@article{cox1958regression,
  title={The regression analysis of binary sequences},
  author={Cox, David R},
  journal={Journal of the Royal Statistical Society Series B: Statistical Methodology},
  volume={20},
  number={2},
  pages={215--232},
  year={1958},
  publisher={Oxford University Press}
}

@inproceedings{milletari2016v,
  title={V-net: Fully convolutional neural networks for volumetric medical image segmentation},
  author={Milletari, Fausto and Navab, Nassir and Ahmadi, Seyed-Ahmad},
  booktitle={2016 fourth international conference on 3D vision (3DV)},
  pages={565--571},
  year={2016},
  organization={Ieee}
}

@article{zu2024resaceunet,
author = {Zu, Shaohuan and Zhao, Penghui and Ke, Chaofan and Junxing, Cao},
title = {ResACEUnet: An Improved Transformer Unet Model for 3D Seismic Fault Detection},
journal = {Journal of Geophysical Research: Machine Learning and Computation},
volume = {1},
number = {3},
pages = {e2024JH000232},
doi = {https://doi.org/10.1029/2024JH000232},
url = {https://agupubs.onlinelibrary.wiley.com/doi/abs/10.1029/2024JH000232},
eprint = {https://agupubs.onlinelibrary.wiley.com/doi/pdf/10.1029/2024JH000232},
note = {e2024JH000232 2024JH000232},
year = {2024}
}

@inproceedings{wazir2022histoseg,
  title={HistoSeg: Quick attention with multi-loss function for multi-structure segmentation in digital histology images},
  author={Wazir, Saad and Fraz, Muhammad Moazam},
  booktitle={2022 12th International Conference on Pattern Recognition Systems (ICPRS)},
  pages={1--7},
  year={2022},
  organization={IEEE}
}

@ARTICLE{10042020,

  author={Wang, Zhiguo and Wang, Qiannan and Yang, Yang and Liu, Naihao and Chen, Yumin and Gao, Jinghuai},

  journal={IEEE Transactions on Geoscience and Remote Sensing}, 

  title={Seismic Facies Segmentation via a Segformer-Based Specific Encoder–Decoder–Hypercolumns Scheme}, 

  year={2023},

  volume={61},

  number={},

  pages={1-11},

  doi={10.1109/TGRS.2023.3244037}}

@inproceedings{deng2009imagenet,
  title={Imagenet: A large-scale hierarchical image database},
  author={Deng, Jia and Dong, Wei and Socher, Richard and Li, Li-Jia and Li, Kai and Fei-Fei, Li},
  booktitle={2009 IEEE conference on computer vision and pattern recognition},
  pages={248--255},
  year={2009},
  organization={Ieee}
}

@inproceedings{chen2019temporal,
  title={Temporal attentive alignment for large-scale video domain adaptation},
  author={Chen, Min-Hung and Kira, Zsolt and AlRegib, Ghassan and Yoo, Jaekwon and Chen, Ruxin and Zheng, Jian},
  booktitle={Proceedings of the IEEE/CVF international conference on computer vision},
  pages={6321--6330},
  year={2019}
}

@inproceedings{qaiser_vision_2025,
	address = {Doha, Qatar,},
	title = {Vision {Transformers} based {Pore} {Type} {Classification} for {Carbonate} {Reservoir} {Characterization}},
	url = {https://www.earthdoc.org/content/papers/10.3997/2214-4609.2025638042},
	doi = {10.3997/2214-4609.2025638042},
	language = {en},
	urldate = {2025-08-08},
	booktitle = {Innovative {Technology} for {Reservoir} {Optimization}},
	publisher = {European Association of Geoscientists \& Engineers},
	author = {Qaiser, Y. and Ansari, I. and Ansari, Y. and Ansari, M.Y. and Sujay, I. and Khan, T. and Rabbani, H. and Laya, J.C. and Seers, T.},
	year = {2025},
	pages = {1--5},
}

@article{yaqoob_fluidnet-lite_2025,
	title = {{FluidNet}-{Lite}: {Lightweight} convolutional neural network for pore-scale modeling of multiphase flow in heterogeneous porous media},
	volume = {200},
	issn = {03091708},
	shorttitle = {{FluidNet}-{Lite}},
	url = {https://linkinghub.elsevier.com/retrieve/pii/S0309170825000661},
	doi = {10.1016/j.advwatres.2025.104952},
	language = {en},
	urldate = {2025-08-08},
	journal = {Advances in Water Resources},
	author = {Yaqoob, Mohammed and Ansari, Mohammed Yusuf and Ishaq, Mohammed and Ashraf, Unais and Pavuluri, Saideep and Rabbani, Arash and Rabbani, Harris Sajjad and Seers, Thomas D.},
	month = jun,
	year = {2025},
	pages = {104952},
}

@article{yaqoob_advancing_2025,
	title = {Advancing paleontology: a survey on deep learning methodologies in fossil image analysis},
	volume = {58},
	issn = {1573-7462},
	shorttitle = {Advancing paleontology},
	url = {https://link.springer.com/10.1007/s10462-024-11080-y},
	doi = {10.1007/s10462-024-11080-y},
	number = {3},
	urldate = {2025-08-08},
	journal = {Artificial Intelligence Review},
	author = {Yaqoob, Mohammed and Ishaq, Mohammed and Ansari, Mohammed Yusuf and Qaiser, Yemna and Hussain, Rehaan and Rabbani, Harris Sajjad and Garwood, Russell J. and Seers, Thomas D.},
	month = jan,
	year = {2025},
	pages = {83},
}

@article{yaqoob_microcrystalnet_2025,
	title = {{MicroCrystalNet}: {An} {Efficient} and {Explainable} {Convolutional} {Neural} {Network} for {Microcrystal} {Classification} {Using} {Scanning} {Electron} {Microscope} {Petrography}},
	volume = {13},
	copyright = {https://creativecommons.org/licenses/by/4.0/legalcode},
	issn = {2169-3536},
	shorttitle = {{MicroCrystalNet}},
	url = {https://ieeexplore.ieee.org/document/10930763/},
	doi = {10.1109/ACCESS.2025.3552626},
	urldate = {2025-08-08},
	journal = {IEEE Access},
	author = {Yaqoob, Mohammed and Yusuf Ansari, Mohammed and Ishaq, Mohammed and Sujay Anand John Jayachandran, Issac and Hashim, Mohammed S. and Daniel Seers, Thomas},
	year = {2025},
	pages = {53865--53884},
}

@article{yaqoob_geocrack_2024,
	title = {{GeoCrack}: {A} {High}-{Resolution} {Dataset} {For} {Segmentation} of {Fracture} {Edges} in {Geological} {Outcrops}},
	volume = {11},
	issn = {2052-4463},
	shorttitle = {{GeoCrack}},
	url = {https://www.nature.com/articles/s41597-024-04107-0},
	doi = {10.1038/s41597-024-04107-0},
	language = {en},
	number = {1},
	urldate = {2025-08-08},
	journal = {Scientific Data},
	author = {Yaqoob, Mohammed and Ishaq, Mohammed and Ansari, Mohammed Yusuf and Konagandla, Venkata Ram Sagar and Tamimi, Tamim Al and Tavani, Stefano and Corradetti, Amerigo and Seers, Thomas Daniel},
	month = dec,
	year = {2024},
	pages = {1318},
}

@article{wenxue_overview_2025,
	title = {An {Overview} {Study} of {Deep} {Learning} in {Geophysics}: {Cross}-{Cutting} {Research} to {Advance} {Geoscience}},
	volume = {13},
	copyright = {https://creativecommons.org/licenses/by-nc-nd/4.0/},
	issn = {2169-3536},
	shorttitle = {An {Overview} {Study} of {Deep} {Learning} in {Geophysics}},
	url = {https://ieeexplore.ieee.org/document/11072451/},
	doi = {10.1109/ACCESS.2025.3586693},
	urldate = {2025-08-08},
	journal = {IEEE Access},
	author = {Wenxue, Zhao and Shikun, Dai and Hongjun, Tian and Dexiang, Zhu and Ying, Zhang and Fan, Jiang},
	year = {2025},
	pages = {124364--124388},
}

@article{isah_review_2025,
	title = {A {Review} of {Data}-{Driven} {Machine} {Learning} {Applications} in {Reservoir} {Petrophysics}},
	issn = {2193-567X, 2191-4281},
	url = {https://link.springer.com/10.1007/s13369-025-10329-0},
	doi = {10.1007/s13369-025-10329-0},
	language = {en},
	urldate = {2025-08-08},
	journal = {Arabian Journal for Science and Engineering},
	author = {Isah, Abubakar and Tariq, Zeeshan and Mustafa, Ayyaz and Mahmoud, Mohamed and Okoroafor, Esuru Rita},
	month = jun,
	year = {2025},
}

@article{zhang2022survey,
  title={A survey on negative transfer},
  author={Zhang, Wen and Deng, Lingfei and Zhang, Lei and Wu, Dongrui},
  journal={IEEE/CAA Journal of Automatica Sinica},
  volume={10},
  number={2},
  pages={305--329},
  year={2022},
  publisher={IEEE}
}

@INPROCEEDINGS{trinidad2023seismic,

  author={Trinidad, Maykol J. Campos and Canchumuni, Smith W. Arauco and Pacheco, Marco A. Cavalcanti},

  booktitle={2023 20th ACS/IEEE International Conference on Computer Systems and Applications (AICCSA)}, 

  title={Seismic Fault Segmentation Using Unsupervised Domain Adaptation}, 

  year={2023},

  volume={},

  number={},

  pages={1-8},

  doi={10.1109/AICCSA59173.2023.10479325}}

@article{mousavi_applications_2024,
	title = {Applications of deep neural networks in exploration seismology: {A} technical survey},
	volume = {89},
	issn = {0016-8033, 1942-2156},
	shorttitle = {Applications of deep neural networks in exploration seismology},
	url = {https://library.seg.org/doi/10.1190/geo2023-0063.1},
	doi = {10.1190/geo2023-0063.1},
	language = {en},
	number = {1},
	urldate = {2025-08-26},
	journal = {GEOPHYSICS},
	author = {Mousavi, S. Mostafa and Beroza, Gregory C. and Mukerji, Tapan and Rasht-Behesht, Majid},
	month = jan,
	year = {2024},
	pages = {WA95--WA115},
}

@article{mousavi_deep-learning_2022,
	title = {Deep-learning seismology},
	volume = {377},
	issn = {0036-8075, 1095-9203},
	url = {https://www.science.org/doi/10.1126/science.abm4470},
	doi = {10.1126/science.abm4470},
	language = {en},
	number = {6607},
	urldate = {2025-08-26},
	journal = {Science},
	author = {Mousavi, S. Mostafa and Beroza, Gregory C.},
	month = aug,
	year = {2022},
	pages = {eabm4470},
}

@article{geirhos2020shortcut,
  title={Shortcut learning in deep neural networks},
  author={Geirhos, Robert and Jacobsen, J{\"o}rn-Henrik and Michaelis, Claudio and Zemel, Richard and Brendel, Wieland and Bethge, Matthias and Wichmann, Felix A},
  journal={Nature Machine Intelligence},
  volume={2},
  number={11},
  pages={665--673},
  year={2020},
  publisher={Nature Publishing Group UK London}
}

@inproceedings{recht2019imagenet,
  title={Do imagenet classifiers generalize to imagenet?},
  author={Recht, Benjamin and Roelofs, Rebecca and Schmidt, Ludwig and Shankar, Vaishaal},
  booktitle={International conference on machine learning},
  pages={5389--5400},
  year={2019},
  organization={PMLR}
}

@book{rutter2018geometry,
  title={Geometry of curves},
  author={Rutter, John W},
  year={2018},
  publisher={Chapman and Hall/CRC}
}

@article{kim2005relationship,
  title={The relationship between displacement and length of faults: a review},
  author={Kim, Young-Seog and Sanderson, David J},
  journal={Earth-Science Reviews},
  volume={68},
  number={3-4},
  pages={317--334},
  year={2005},
  publisher={Elsevier}
}

@article{hanson2004style,
  title={Style and rate of Quaternary deformation of the Hosgri Fault Zone, offshore south-central California},
  author={Hanson, Kathryn L and Lettis, William R and McLaren, Marcia K and Savage, William U and Hall, N Timothy and Keller, MA},
  journal={US Geological Survey Bulletin 1995-BB},
  pages={33},
  year={2004}
}

@article{schwartz1989fault,
  title={Fault segmentation and controls of rupture initiation and termination},
  author={Schwartz, DP and Sibson, RH},
  journal={USGS Open-File Report},
  volume={89315},
  pages={445},
  year={1989},
  publisher={USGS}
}

@article{bailey2013geological,
  title={Geological and geophysical evaluation of the Thebe field, Block XX, offshore Western Australia},
  author={Bailey, Brett B},
  year={2013},
  publisher={University of the Western Cape}
}

@misc{terranubis_f3_demo_2023,
  author    = {{TerraNubis}},
  title     = {Project {F3} Demo 2023 — Data Info},
  year      = {2023},
  url       = {https://terranubis.com/datainfo/F3-Demo-2023},
  urldate   = {2025-10-30},
  note      = {Data page by dGB Earth Sciences}
}

@article{manighetti2021fault,
  title={Fault trace corrugation and segmentation as a measure of fault structural maturity},
  author={Manighetti, I and Mercier, Antoine and De Barros, Louis},
  journal={Geophysical Research Letters},
  volume={48},
  number={20},
  pages={e2021GL095372},
  year={2021},
  publisher={Wiley Online Library}
}

@inproceedings{yang2020fda,
  title={Fda: Fourier domain adaptation for semantic segmentation},
  author={Yang, Yanchao and Soatto, Stefano},
  booktitle={Proceedings of the IEEE/CVF conference on computer vision and pattern recognition},
  pages={4085--4095},
  year={2020}
}

@INPROCEEDINGS{logan2022multi,

  author={Logan, Yash-Yee and Kokilepersaud, Kiran and Kwon, Gukyeong and AlRegib, Ghassan and Wykoff, Charles and Yu, Hannah},

  booktitle={2022 IEEE 19th International Symposium on Biomedical Imaging (ISBI)}, 

  title={Multi-Modal Learning Using Physicians Diagnostics for Optical Coherence Tomography Classification}, 

  year={2022},

  volume={},

  number={},

  pages={1-5},

  doi={10.1109/ISBI52829.2022.9761432}}

@inproceedings{logan2022patient,
  title={Patient aware active learning for fine-grained oct classification},
  author={Logan, Yash-yee and Benkert, Ryan and Mustafa, Ahmad and AlRegib, Ghassan},
  booktitle={2022 IEEE International Conference on Image Processing (ICIP)},
  pages={3908--3912},
  year={2022},
  organization={IEEE}
}

@ARTICLE{kokilepersaud2023clinically,

  author={Kokilepersaud, Kiran and Corona, Stephanie Trejo and Prabhushankar, Mohit and AlRegib, Ghassan and Wykoff, Charles},

  journal={IEEE Journal of Biomedical and Health Informatics}, 

  title={Clinically Labeled Contrastive Learning for OCT Biomarker Classification}, 

  year={2023},

  volume={27},

  number={9},

  pages={4397-4408},

  doi={10.1109/JBHI.2023.3277789}}

@INPROCEEDINGS{kokilepersaud2022gradient,

  author={Kokilepersaud, Kiran and Prabhushankar, Mohit and AlRegib, Ghassan and Trejo Corona, Stephanie and Wykoff, Charles},

  booktitle={2022 IEEE International Conference on Image Processing (ICIP)}, 

  title={Gradient-Based Severity Labeling for Biomarker Classification in Oct}, 

  year={2022},

  volume={},

  number={},

  pages={3416-3420},

  doi={10.1109/ICIP46576.2022.9897215}}

@inproceedings{quesada2022mtneuro,
 author = {Quesada, Jorge and Sathidevi, Lakshmi and Liu, Ran and Ahad, Nauman and Jackson, Joy and Azabou, Mehdi and Xiao, Jingyun and Liding, Christopher and Jin, Matthew and Urzay, Carolina and Gray-Roncal, William and Johnson, Erik and Dyer, Eva},
 booktitle = {Advances in Neural Information Processing Systems},
 editor = {S. Koyejo and S. Mohamed and A. Agarwal and D. Belgrave and K. Cho and A. Oh},
 pages = {5299--5314},
 publisher = {Curran Associates, Inc.},
 title = {MTNeuro:  A Benchmark for Evaluating Representations of Brain Structure Across Multiple Levels of Abstraction},
 url = {https://proceedings.neurips.cc/paper_files/paper/2022/file/22fb65e39d318c4b5b56fbe9cb082e3f-Paper-Datasets_and_Benchmarks.pdf},
 volume = {35},
 year = {2022}
}

@article{ansari_survey_2025,
	title = {A survey of transformers and large language models for {ECG} diagnosis: advances, challenges, and future directions},
	volume = {58},
	issn = {1573-7462},
	shorttitle = {A survey of transformers and large language models for {ECG} diagnosis},
	url = {https://link.springer.com/10.1007/s10462-025-11259-x},
	doi = {10.1007/s10462-025-11259-x},
	language = {en},
	number = {9},
	urldate = {2025-08-30},
	journal = {Artificial Intelligence Review},
	author = {Ansari, Mohammed Yusuf and Yaqoob, Mohammed and Ishaq, Mohammed and Flushing, Eduardo Feo and Mangalote, Iffa Afsa Changaai and Dakua, Sarada Prasad and Aboumarzouk, Omar and Righetti, Raffaella and Qaraqe, Marwa},
	month = jun,
	year = {2025},
	pages = {261},
}

@ARTICLE{iffa2025development,

  author={M, Iffa Afsa C and Ansari, Mohammed Yusuf and Paul, Santu and Halabi, Osama and Alataresh, Ezzedin and Shah, Jassim and Hamze, Afaf and Aboumarzouk, Omar and Al-Ansari, Abdulla and Dakua, Sarada Prasad},

  journal={IEEE Transactions on Biomedical Engineering}, 

  title={Development and Validation of a Class Imbalance-Resilient Cardiac Arrest Prediction Framework Incorporating Multiscale Aggregation, ICA and Explainability}, 

  year={2025},

  volume={72},

  number={5},

  pages={1674-1687},

  doi={10.1109/TBME.2024.3517635}}

@article{guan_domain_2022,
	title = {Domain {Adaptation} for {Medical} {Image} {Analysis}: {A} {Survey}},
	volume = {69},
	copyright = {https://ieeexplore.ieee.org/Xplorehelp/downloads/license-information/IEEE.html},
	issn = {0018-9294, 1558-2531},
	shorttitle = {Domain {Adaptation} for {Medical} {Image} {Analysis}},
	url = {https://ieeexplore.ieee.org/document/9557808/},
	doi = {10.1109/TBME.2021.3117407},
	number = {3},
	urldate = {2025-10-29},
	journal = {IEEE Transactions on Biomedical Engineering},
	author = {Guan, Hao and Liu, Mingxia},
	month = mar,
	year = {2022},
	pages = {1173--1185},
}

\begin{IEEEbiography}[{\includegraphics[width=1in,height=1.25in,clip,keepaspectratio]{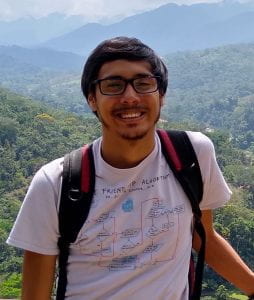}}]{Jorge Quesada} received his B.E. and M.S. degrees from the Pontifical Catholic University of Peru. He joined the Georgia Institute of Technology as a Machine Learning PhD student in the department of Electrical and Computer Engineering in 2021, where he is now part of the Omni Lab for Intelligent Visual Engineering and Science (OLIVES). He is interested in leveraging machine learning and image processing tools to study the mechanisms underlying computer and biological vision.
\end{IEEEbiography}

\begin{IEEEbiography}[{\includegraphics[width=1in,height=1.25in,clip,keepaspectratio]{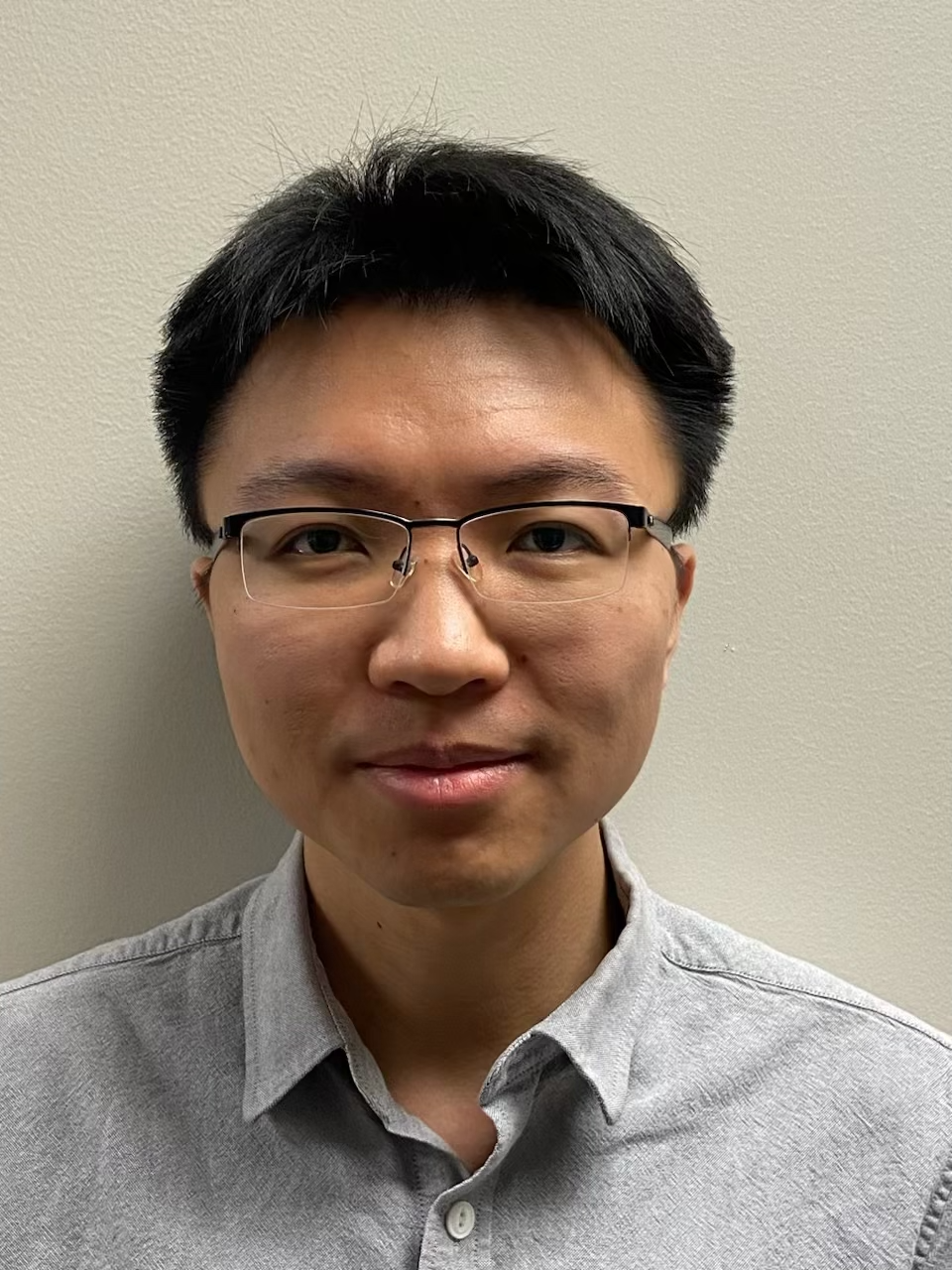}}]{Chen Zhou} is a Ph.D. student in the School of Electrical and Computer Engineering at the Georgia Institute of Technology. He is currently a Graduate Research Assistant in the Omni Lab for Intelligent Visual Engineering and Science (OLIVES). He is working in the fields of machine learning,and image and video processing. His research interests include trajectory prediction and learning from label disagreement for the applications of autonomous vehicle, and seismic interpretation.
\end{IEEEbiography}

\begin{IEEEbiography}[{\includegraphics[width=1in,height=1.25in,clip,keepaspectratio]{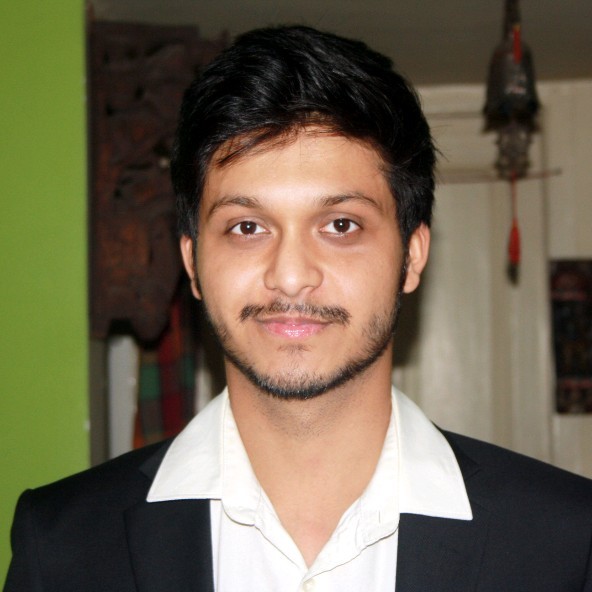}}]{Prithwijit Chowdhury} received his B.Tech. degree from KIIT University, India, in 2020. He joined the Georgia Institute of Technology as an M.S. student in the School of Electrical and Computer Engineering in 2021 and is currently pursuing his Ph.D. as a researcher in The Center for Energy and Geo Processing (CeGP) and as a member of the Omni Lab for Intelligent Visual Engineering and Science (OLIVES). His research interests lie in digital signal and image processing and machine learning with applications to geophysics. He is an IEEE Student Member and a published author, with several works presented at the IMAGE conference and published in the GEOPHYSICS journal.
\end{IEEEbiography}

\begin{IEEEbiography}[{\includegraphics[width=1in,height=1.25in,clip,keepaspectratio]{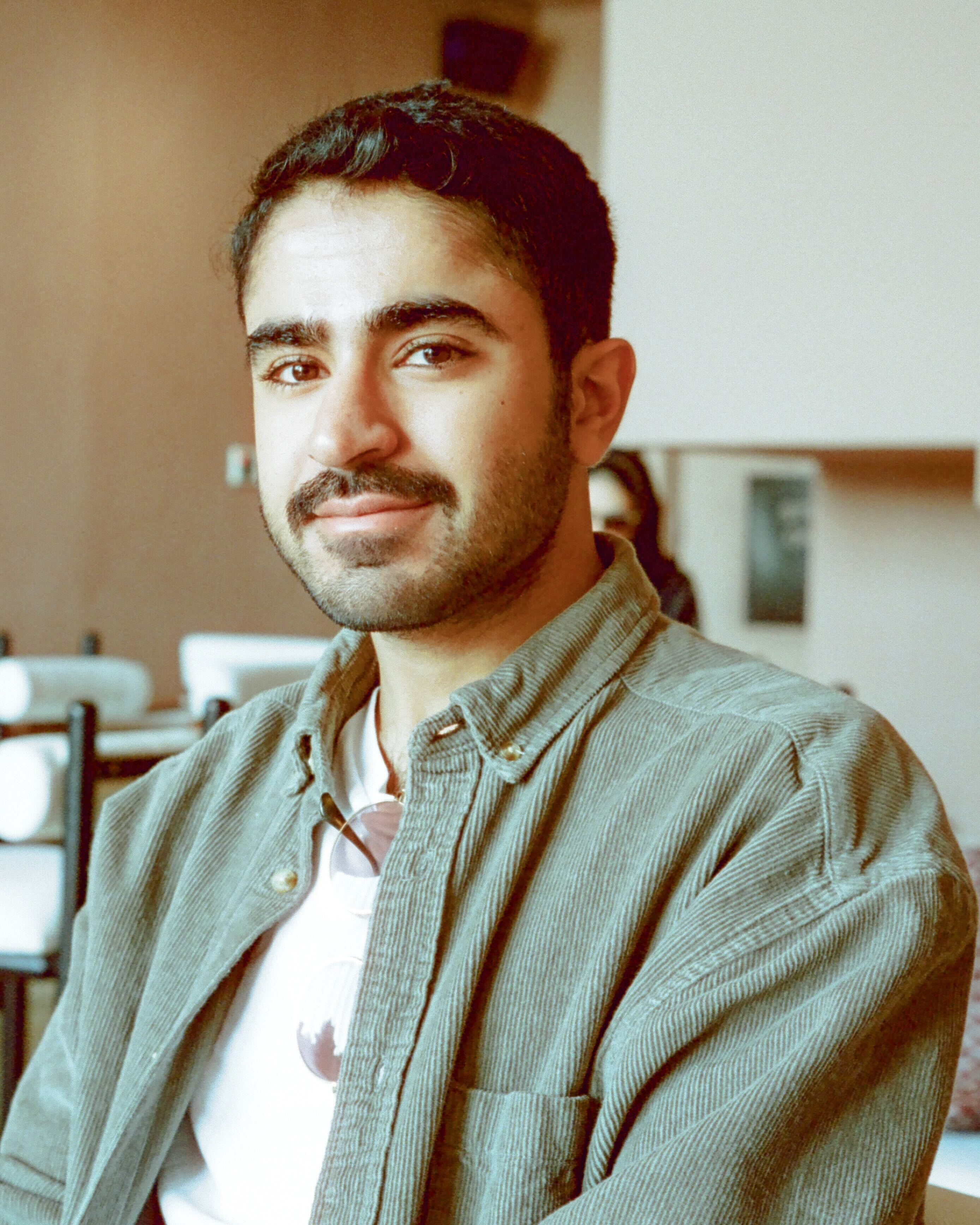}}]{Mohammad Alotaibi} is a Ph.D. student in the School of Electrical and Computer Engineering at the Georgia Institute of Technology. He is a Graduate Research Assistant in the Omni Lab for Intelligent Visual Engineering and Science (OLIVES). His research focuses on machine learning and image processing, with particular interest in domain adaptation for seismic and medical imaging applications.
\end{IEEEbiography}

\begin{IEEEbiography}[{\includegraphics[width=1in,height=1.25in,clip,keepaspectratio]{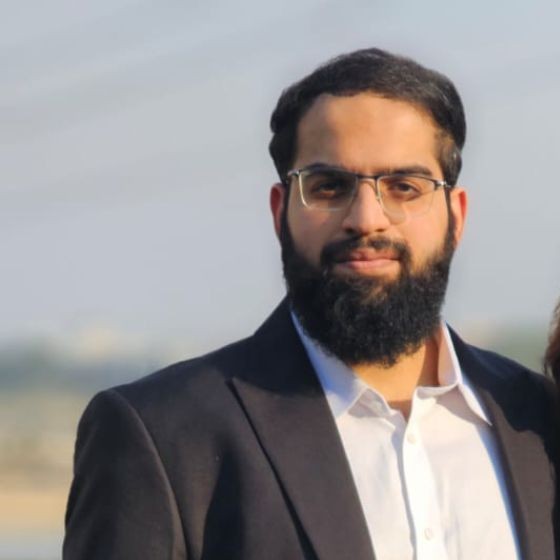}}]{Ahmad Mustafa} is an Assistant Professor at the Department of Computer Science based in the Information Technology University (ITU), Lahore, Pakistan. He is the Director of Computational Imaging, Vision, Intelligence and Learning (CIVIL) lab. His research interests span machine learning, weakly supervised learning, and computational image interpretation in scientific computing domains. He obtained his PhD in Electrical and Computer Engineering at the Georgia Institute of Technology, USA. Prior to joining ITU, he worked at Occidental Petroleum, leading the development of large-scale computational methods to streamline subsurface applications. He was awarded the 2023 Roger P. Webb ECE GRA Excellence Award in recognition of his research contributions by Georgia Tech. He was the recipient of the Google Cloud Research Credits Award in 2025 for his work on deep learning for computational subsurface imaging and understanding. For his teaching services, he was awarded the Outstanding Online Head TA Award, the ECE GTA Excellence Award, and the ECE CREATION Award.  His work has been featured in top-tier peer-reviewed academic journals, conference proceedings, and technical presentations, and is frequently cited by both academic and industry peers.
\end{IEEEbiography}

\begin{IEEEbiography}[{\includegraphics[width=1in,height=1.25in,clip,keepaspectratio]{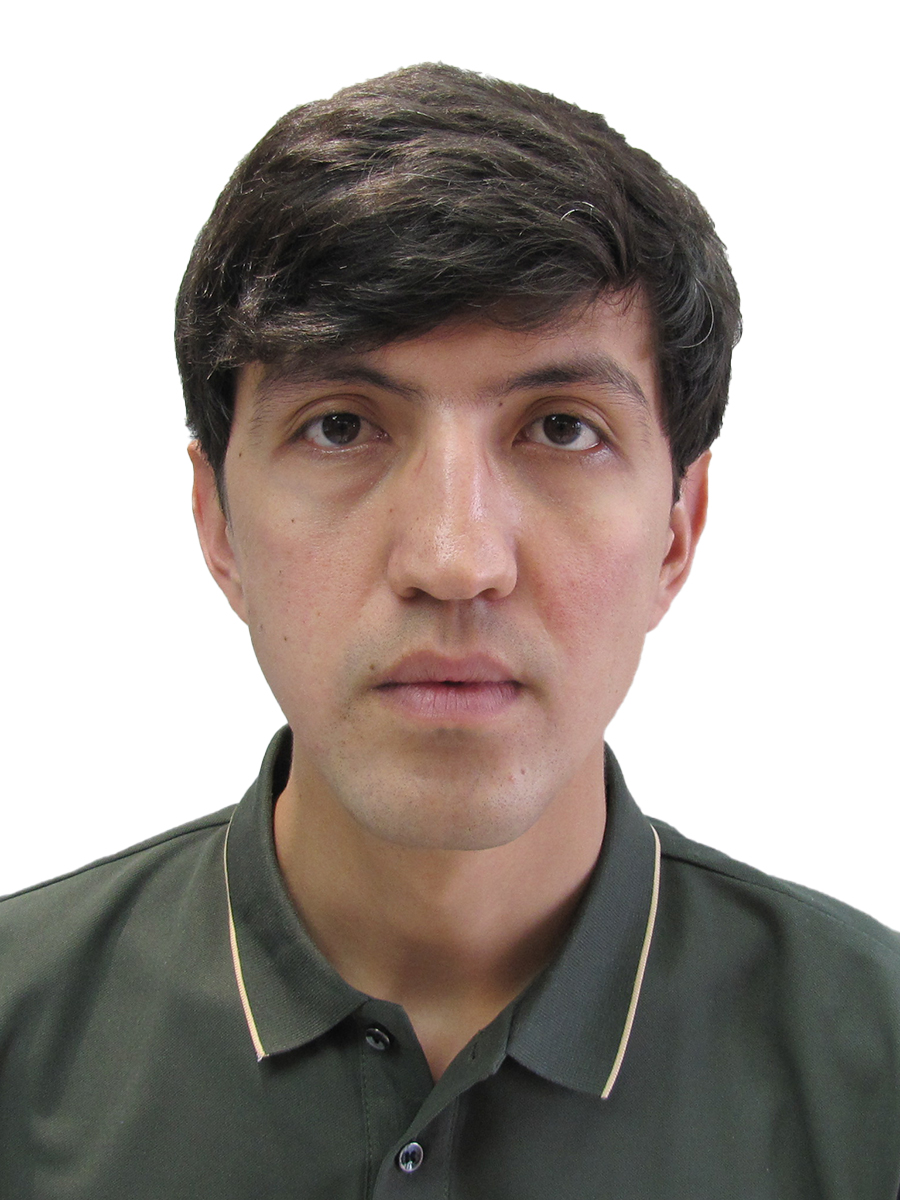}}]{Yusufjon Kumakov} is currently serving as an Assistant Lecturer at Tashkent State Technical University. He earned his Master’s degree in Petroleum Engineering from the Polytechnic University of Turin, Italy. His primary research interests lie in the application of machine learning techniques to geophysics, with a focus on seismic imaging and signal processing.
\end{IEEEbiography}

\begin{IEEEbiography}[{\includegraphics[width=1in,height=1.25in,clip,keepaspectratio]{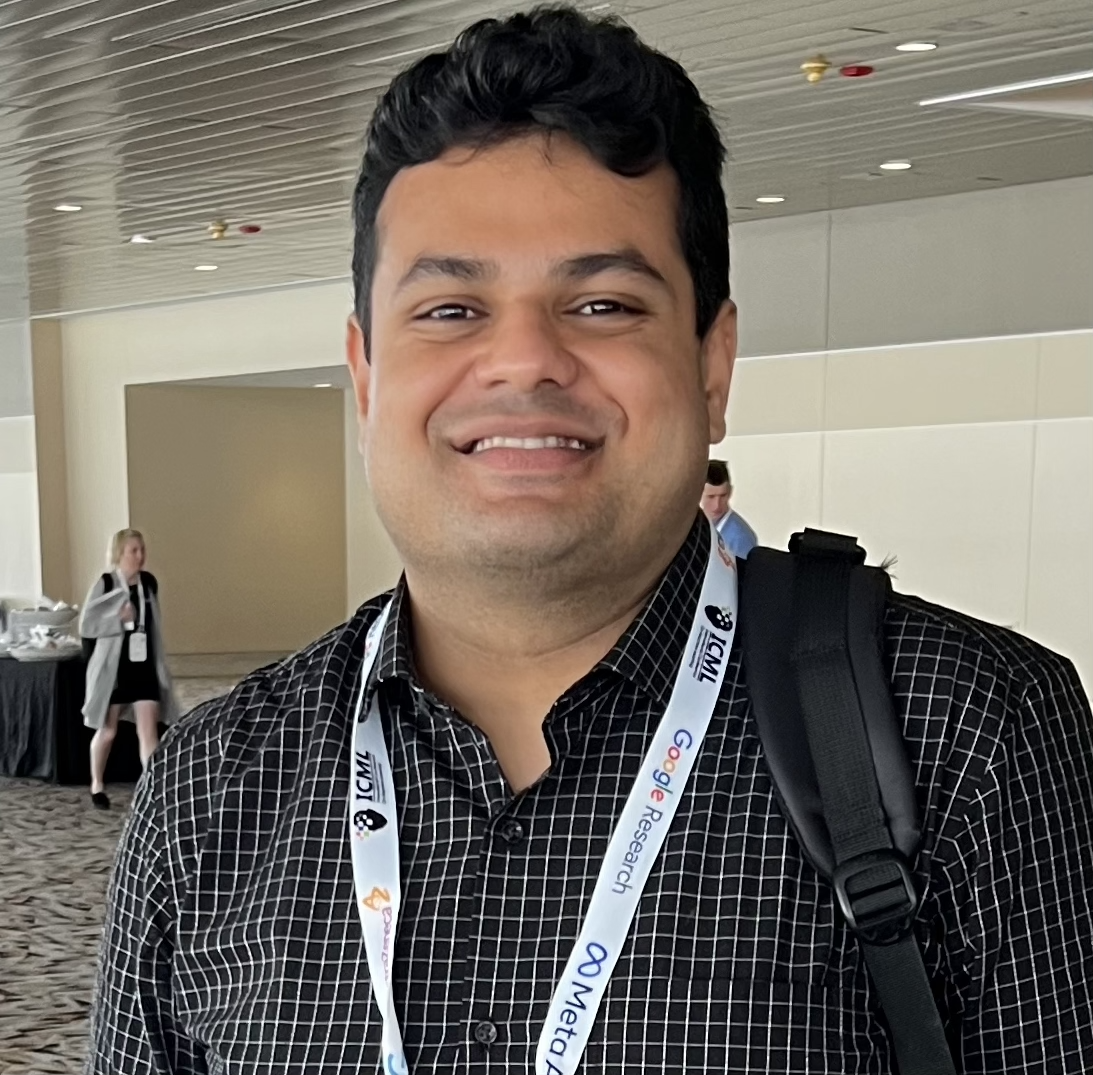}}]{Mohit Prabhushankar} received his Ph.D. degree in electrical engineering from the Georgia Institute of Technology (Georgia Tech), Atlanta, Georgia, 30332, USA, in 2021. He is currently a Postdoctoral Research Fellow in the School of Electrical and Computer Engineering at the Georgia Institute of Technology in the Omni Lab for Intelligent Visual Engineering and Science (OLIVES). He is working in the fields of image processing, machine learning, active learning, healthcare, and robust and explainable AI. He is the recipient of the Best Paper award at ICIP 2019 and Top Viewed Special Session Paper Award at ICIP 2020. He is the recipient of the ECE Outstanding Graduate Teaching Award, the CSIP Research award, and of the Roger P Webb ECE Graduate Research Assistant Excellence award,all in 2022. He has delivered short courses and tutorials at IEEE IV'23, ICIP'23, BigData'23, WACV'24 and AAAI'24.
\end{IEEEbiography}

\begin{IEEEbiography}[{\includegraphics[width=1in,height=1.25in,clip,keepaspectratio]{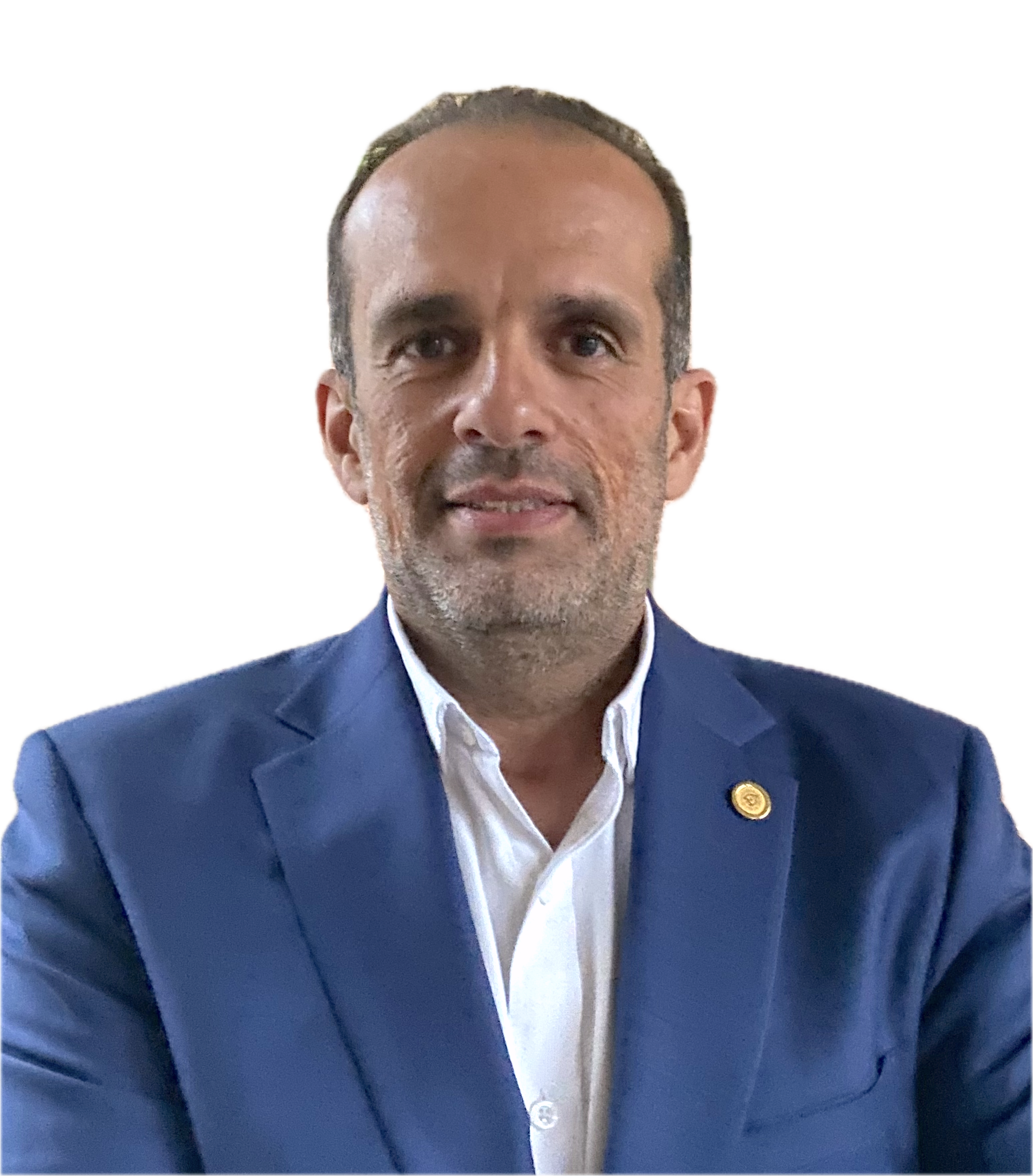}}]{Ghassan AlRegib}  is currently the John and Marilu McCarty Chair Professor in the School of Electrical and Computer Engineeringat the Georgia Institute of Technology. In theOmni Lab for Intelligent Visual Engineering and Science (OLIVES), he and his groupwork on robustand interpretable machine learning algorithms, uncertainty and trust, and human in the loop algorithms. The group has demonstrated their work on a widerange of applications such as Autonomous Systems, Medical Imaging, and Subsurface Imaging. The group isinterested in advancing the fundamentals as well as the deployment of such systems in real-world scenarios. He has been issued several U.S.patents and invention disclosures. He is a Fellow of the IEEE. Prof. AlRegib is active in the IEEE. He served on the editorial board of several transactions and served as the TPC Chair for ICIP 2020, ICIP 2024, and GlobalSIP 2014. He was area editor for the IEEE Signal Processing Magazine. In 2008, he received the ECE Outstanding Junior Faculty Member Award. In 2017, he received the 2017 Denning Faculty Award for Global Engagement.He received the 2024 ECE Distinguished Faculty Achievement Award at Georgia Tech. He and his students received the Best Paper Award in ICIP 2019and the 2023 EURASIP Best Paper Award for Image communication Journal. In addition, one of their papers is the best paper runner-up at BigData 2024. In 2024, he co-founded the AI Makerspace at Georgia Tech, where any student and any community member can access and utilize AI regardless of their background. 
\end{IEEEbiography}

% % \section*{References and Footnotes}

% \subsection{Footnotes}
% Number footnotes separately in superscript numbers.\footnote{It is recommended that footnotes be avoided (except for 
% the unnumbered footnote with the receipt date on the first page). Instead, 
% try to integrate the footnote information into the text.} Place the actual 
% footnote at the bottom of the column in which it is cited; do not put 
% footnotes in the reference list (endnotes). Use letters for table footnotes 
% (see Table \ref{table}).

\EOD
\end{document}